%% file: 0_main.tex
\documentclass[11pt]{article}

\PassOptionsToPackage{numbers}{natbib}
\PassOptionsToPackage{sort}{natbib}

\usepackage{geometry}
\usepackage[utf8]{inputenc}
\usepackage[T1]{fontenc}
\usepackage{fullpage}
\usepackage[numbers]{natbib}

\usepackage{graphicx}
\usepackage{tikz}
\usepackage{color}
\usepackage[accsupp]{axessibility}

\usepackage{subfigure}
\usepackage{booktabs}
\usepackage{amsmath,bm}
\usepackage{amsthm}
\usepackage{enumerate}
\usepackage{algorithm}
\usepackage{algorithmic}
\usepackage{colortbl}
\usepackage{psfrag}
\usepackage{adjustbox}
\usepackage{wrapfig}
\usepackage{xspace}
\usepackage{amssymb}
\usepackage{multirow}
\usepackage{afterpage}
\usepackage{float}

\usepackage{enumitem}

\newenvironment{Itemize}{
    \begin{itemize}[leftmargin=*]
    \setlength{\itemsep}{0pt}
    \setlength{\topsep}{0pt}
    \setlength{\partopsep}{0pt}
    \setlength{\parskip}{0pt}}
{\end{itemize}}
\newenvironment{Enumerate}{
    \begin{enumerate}
    \setlength{\itemsep}{0pt}
    \setlength{\topsep}{0pt}
    \setlength{\partopsep}{0pt}
    \setlength{\parskip}{0pt}}
{\end{enumerate}}
\usepackage[most]{tcolorbox}
\tcbset{
    frame code={}
    center title,
    left=0pt,
    right=0pt,
    top=0pt,
    bottom=0pt,
    colframe=white,
    width=\dimexpr\textwidth\relax,
    enlarge left by=0mm,
    boxsep=5pt,
    arc=0pt,outer arc=0pt,
}

\usepackage{xcolor}
\colorlet{darkgreen}{green!65!black}
\colorlet{darkblue}{blue!75!black}
\colorlet{darkred}{red!80!black}
\definecolor{lightblue}{HTML}{0071bc}
\definecolor{lightgreen}{HTML}{39b54a}
\definecolor{manyshot}{HTML}{6969ff}
\definecolor{medshot}{HTML}{f7c600}
\definecolor{fewshot}{HTML}{ff6969}
\definecolor{mypurple}{HTML}{412F8A}
\definecolor{myorange}{HTML}{fc8e62}

\usepackage[pagebackref=false, breaklinks=true, colorlinks,
            citecolor=citecolor, linkcolor=linkcolor, bookmarks=false]{hyperref}
\definecolor{citecolor}{HTML}{0071BC}
\definecolor{linkcolor}{HTML}{ED1C24}

\global\long\def\real{\mathbb{R}}
\global\long\def\E{\mathbb{E}}

\global\long\def\1{\mathds{1}}
\renewcommand{\tilde}{\widetilde}
\renewcommand{\hat}{\widehat}

\newcommand{\defref}[1]{Definition~\ref{#1}}

\newtheorem{theorem}{Theorem}
\newtheorem{lemma}{Lemma}
\newtheorem{definition}{Definition}

\newcommand{\boda}{\texttt{BoDA}\xspace}
\global\long\def\dist{\mathsf{d}}
\global\long\def\bdist{\tilde{\mathsf{d}}}
\global\long\def\cdist{\hat{\mathsf{d}}}

\renewcommand{\paragraph}[1]{\vspace{2mm}\noindent\textbf{#1}}
\newcommand{\grayrow}{\rowcolor[gray]{.9}}

\title{\bf{On Multi-Domain Long-Tailed Recognition,\\Imbalanced Domain Generalization and Beyond}}

\author{Yuzhe Yang\\ \small MIT \and
Hao Wang\\ \small Rutgers University \and
Dina Katabi\\ \small MIT}

\date{}

\begin{document}

\maketitle

%%%%%%%%%%%%%%%%%%%%%%%%%%%%%%%%%%%%%%%%%%%%%%%%%%%%%%%%%%%%
%%%%%%%%%%%%%%%%%%%%%%%%  Abstract  %%%%%%%%%%%%%%%%%%%%%%%%
%%%%%%%%%%%%%%%%%%%%%%%%%%%%%%%%%%%%%%%%%%%%%%%%%%%%%%%%%%%%
\input{1_abstract}

%%%%%%%%%%%%%%%%%%%%%%%%%%%%%%%%%%%%%%%%%%%%%%%%%%%%%%%%%%%%
%%%%%%%%%%%%%%%%%%%%%%  Introduction  %%%%%%%%%%%%%%%%%%%%%%
%%%%%%%%%%%%%%%%%%%%%%%%%%%%%%%%%%%%%%%%%%%%%%%%%%%%%%%%%%%%
\section{Introduction}
\label{sec:intro}
\input{2_introduction}

%%%%%%%%%%%%%%%%%%%%%%%%%%%%%%%%%%%%%%%%%%%%%%%%%%%%%%%%%%%%
%%%%%%%%%%%%%%%%%%%%%  Related Works  %%%%%%%%%%%%%%%%%%%%%%
%%%%%%%%%%%%%%%%%%%%%%%%%%%%%%%%%%%%%%%%%%%%%%%%%%%%%%%%%%%%
\section{Related Work}
\label{sec:related-work}
\input{3_related}

%%%%%%%%%%%%%%%%%%%%%%%%%%%%%%%%%%%%%%%%%%%%%%%%%%%%%%%%%%%%
%%%%%%%%%%%%%%%%%%%%%%%%  Methods  %%%%%%%%%%%%%%%%%%%%%%%%%
%%%%%%%%%%%%%%%%%%%%%%%%%%%%%%%%%%%%%%%%%%%%%%%%%%%%%%%%%%%%
\section{Domain-Class Transferability Graph}
\label{sec:trans_graph}
\input{4_trans_graph}

\section{What Makes for Good Representations in MDLT?}
\label{sec:rep-learn}
\input{5_rep_learn}

\section{What Makes for Good Classifiers in MDLT?}
\label{sec:cls-learn}
\input{6_cls_learn}

%%%%%%%%%%%%%%%%%%%%%%%%%%%%%%%%%%%%%%%%%%%%%%%%%%%%%%%%%%%%
%%%%%%%%%%%%%%%%%%%%%%  Experiments  %%%%%%%%%%%%%%%%%%%%%%%
%%%%%%%%%%%%%%%%%%%%%%%%%%%%%%%%%%%%%%%%%%%%%%%%%%%%%%%%%%%%
\section{Benchmarking MDLT}
\label{sec:experiment-mdlt}
\input{7_exp_mdlt}

\section{Beyond MDLT: (Imbalanced) Domain Generalization}
\label{sec:experiment-dg}
\input{8_exp_dg}

%%%%%%%%%%%%%%%%%%%%%%%%%%%%%%%%%%%%%%%%%%%%%%%%%%%%%%%%%%%%
%%%%%%%%%%%%%%%%%%%%%%  Conclusion  %%%%%%%%%%%%%%%%%%%%%%%%
%%%%%%%%%%%%%%%%%%%%%%%%%%%%%%%%%%%%%%%%%%%%%%%%%%%%%%%%%%%%
\section{Conclusion}
\label{sec:conclusion}
\input{9_conclusion}

%%%%%%%%%%%%%%%%%%%%%%%%%%%%%%%%%%%%%%%%%%%%%%%%%%%%%%%%%%%%
%%%%%%%%%%%%%%%%%%%  Acknowledgments  %%%%%%%%%%%%%%%%%%%%%%
%%%%%%%%%%%%%%%%%%%%%%%%%%%%%%%%%%%%%%%%%%%%%%%%%%%%%%%%%%%%
\subsection*{Acknowledgments}
This work is supported by the GIST-MIT Research Collaboration grant funded by GIST.
Yuzhe Yang is supported by the MathWorks Fellowship.

%%%%%%%%%%%%%%%%%%%%%%%%%%%%%%%%%%%%%%%%%%%%%%%%%%%%%%%%%%%%
%%%%%%%%%%%%%%%%%%%%%%%  Reference  %%%%%%%%%%%%%%%%%%%%%%%%
%%%%%%%%%%%%%%%%%%%%%%%%%%%%%%%%%%%%%%%%%%%%%%%%%%%%%%%%%%%%
\bibliographystyle{abbrvnat}
\bibliography{mdlt}

%%%%%%%%%%%%%%%%%%%%%%%%%%%%%%%%%%%%%%%%%%%%%%%%%%%%%%%%%%%%
%%%%%%%%%%%%%%%%%%%%%%%  Appendix  %%%%%%%%%%%%%%%%%%%%%%%%%
%%%%%%%%%%%%%%%%%%%%%%%%%%%%%%%%%%%%%%%%%%%%%%%%%%%%%%%%%%%%
\newpage
\appendix
\input{10_appendix}

\end{document}

%% file: 1_abstract.tex
\begin{abstract}
Real-world data often exhibit imbalanced label distributions. Existing studies on data imbalance focus on single-domain settings, i.e., samples are from the same data distribution. However, natural data can originate from distinct domains, where a minority class in one domain could have abundant instances from other domains.
We formalize the task of {Multi-Domain Long-Tailed Recognition} (MDLT), which learns from multi-domain imbalanced data, addresses \emph{label imbalance}, \emph{domain shift}, and \emph{divergent label distributions across domains}, and generalizes to all domain-class pairs.
We first develop the \emph{domain-class transferability graph}, and show that such transferability governs the success of learning in MDLT.
We then propose \boda, a theoretically grounded learning strategy that tracks the upper bound of transferability statistics, and ensures \emph{balanced} alignment and calibration across imbalanced domain-class distributions.
We curate five MDLT benchmarks based on widely-used multi-domain datasets, and compare \boda to twenty algorithms that span different learning strategies.
Extensive and rigorous experiments verify the superior performance of \boda. Further, as a byproduct, \boda establishes new state-of-the-art on Domain Generalization benchmarks, highlighting the importance of addressing data imbalance across domains, which can be crucial for improving generalization to unseen domains.
Code and data are available at: {\url{https://github.com/YyzHarry/multi-domain-imbalance}}.
\end{abstract}

%% file: 2_introduction.tex
Real-world data often exhibit label imbalance -- i.e.,  instead of a uniform label distribution over classes, in reality, data are by their nature imbalanced: a few classes contain a large number of instances, whereas many others have only a few instances \cite{buda2018systematic,cao2019learning,yang2020rethinking}. This phenomenon poses a challenge for deep recognition models, and has motivated several prior solutions~\cite{cao2019learning,ren2020bsoftmax,liu2019large,cui2019class,yang2020rethinking,yang2021delving}. Such prior solutions focus on \emph{single domain} scenarios, i.e., samples are from the same data distribution; they propose techniques for learning from imbalanced training data and generalizing to a balanced test set.

In contrast, this paper formulates the problem of \emph{Multi-Domain Long-Tailed Recognition} (MDLT) as learning from multi-domain imbalanced data, with each domain having its own imbalanced label distribution, and generalizing to a test set that is balanced over all domain-class pairs.  MDLT is a natural extension of the single domain case. It arises in real-world scenarios, where data targeted for one task can originate from different domains. For example, in visual recognition problems, minority classes from ``photo'' images could be complemented with potentially abundant samples from ``sketch'' images. Similarly, in autonomous driving, the minority accident class in ``real'' life could be enriched with accidents generated in ``simulation''. Also, in medical diagnosis, data from distinct populations could enhance each other, where minority samples from one institution could be enriched with instances from others. In the above examples, different data types act as distinct \emph{domains}, and such multi-domain data could be leveraged to tackle the inherent data imbalance within each domain.

\begin{figure}[tb]
\begin{center}
\includegraphics[width=\linewidth]{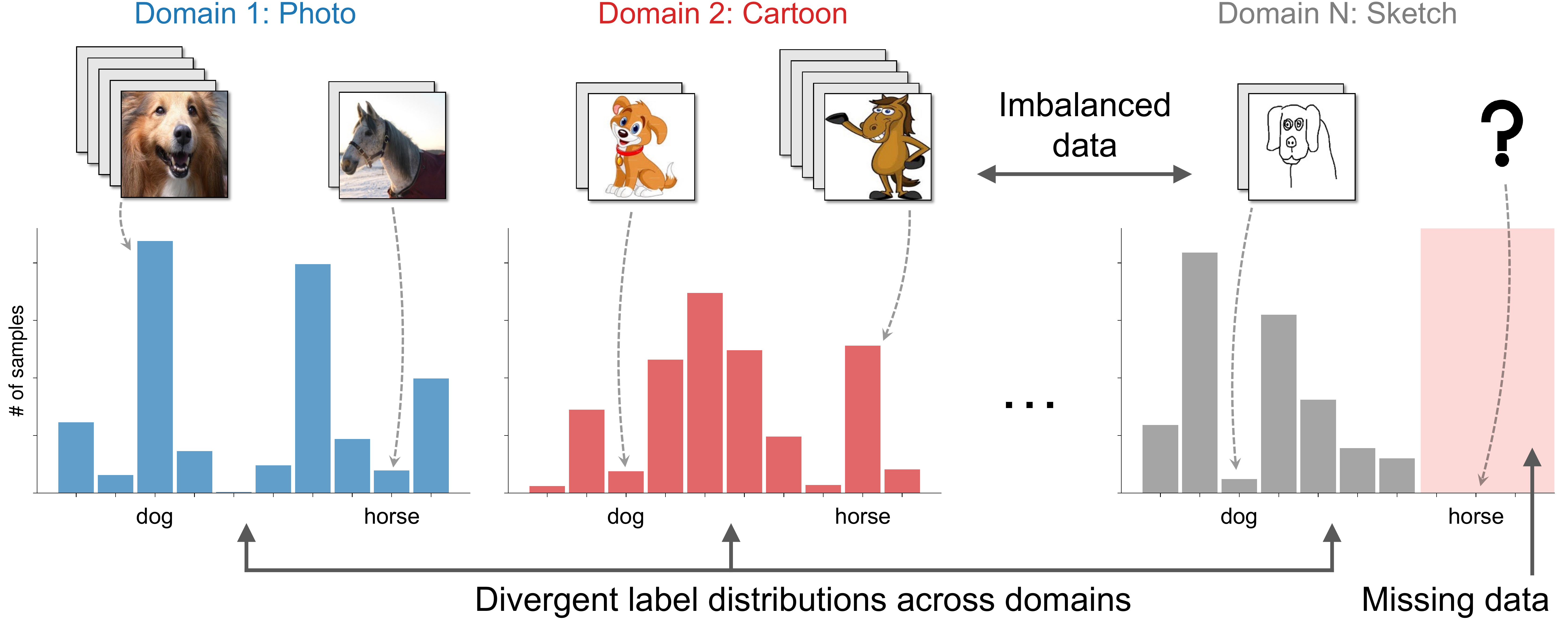}
\end{center}
\vspace{-0.4cm}
\caption{Multi-Domain Long-Tailed Recognition (MDLT) aims to learn from imbalanced data from multiple distinct domains, tackle label imbalance, domain shift, and divergent label distributions across domains, and generalize to the entire set of classes over all domains.}
\label{fig:teaser-intro}
\end{figure}

We note that MDLT has key differences from its single-domain counterpart:
\begin{Itemize}
\item First, the label distribution for each domain is likely different from other domains. For example, in Fig.~\ref{fig:teaser-intro}, both ``Photo'' and ``Cartoon'' domains exhibit imbalanced label distributions; Yet, the ``horse'' class in ``Cartoon'' has many more samples than in ``Photo''. This creates challenges with  \emph{divergent label distributions across domains}, in addition to in-domain data imbalance.
\item Second, multi-domain data inherently involves \emph{domain shift}. Simply treating different domains as a whole and applying traditional data-imbalance methods is unlikely to yield the best results, as the domain gap can be arbitrarily large.
\item Third, MDLT naturally motivates \emph{zero-shot generalization within and across domains} -- i.e., to generalize to both in-domain missing classes (Fig.~\ref{fig:teaser-intro} right part), as well as new domains with no training data, where the latter case is typically denoted as Domain Generalization (DG). 
\end{Itemize}

To deal with the above issues, we first develop the \emph{domain-class transferability graph}, which quantifies the transferability between different domain-class pairs under data imbalance. In this graph, each node refers to a domain-class pair, and each edge refers to the distance between two domain-class pairs in the embedding space. We show that the transferability graph dictates the performance of imbalanced learning across domains. Inspired by this, we design \boda (\underline{B}alanced D\underline{o}main-Class \underline{D}istribution \underline{A}lignment), a new loss function that encourages similarity between features of the same class in different domains, and penalizes similarity between features of different classes within and across domains. \boda does so while accounting for that different classes have very different number of samples, and hence the statistics of their features are intrinsically imbalanced.  Analytically, we prove that minimizing the \boda loss optimizes an upper bound of the \emph{balanced} transferability statistics, corroborating the effectiveness of \boda for learning multi-domain imbalanced data.

For MDLT evaluation, we curate five MDLT benchmarks based on datasets widely used for domain generalization (DG). These datasets naturally exhibit heavy class imbalance within each domain and data shift across domains, highlighting that the MDLT problem is widely present in current benchmarks.
We compare \boda against twenty algorithms that span different learning strategies. Extensive experiments across benchmarks and algorithms verify that \boda consistently outperforms all these baselines on all datasets.

Additionally, we examine how \boda performs in the DG setting. We show that combining \boda with the DG state-of-the-art (SOTA) consistently brings further gains, yielding a new SOTA for DG. These results shed light on how label imbalance can affect out-of-distribution generalization and highlight the importance of integrating label imbalance into practical DG algorithm design.

Our contributions are as follows:
\begin{Itemize}
\item We formulate the MDLT problem as learning from multi-domain imbalanced data and generalizing across all domain-class pairs.
\item We introduce the domain-class transferability graph, a unified model for investigating MDLT. We further show that the transferability statistics induced from such graph are crucial and govern the success of MDLT algorithms.
\item We design \boda, a simple, effective, and interpretable loss function for MDLT. We prove theoretically that minimizing the \boda loss is equivalent to optimizing an upper bound of balanced transferability statistics.
\item Extensive experiments on benchmark datasets verify the superior and consistent performance of \boda. Further, combined with DG algorithms, \boda establishes a new SOTA on DG benchmarks, highlighting the importance of tackling cross-domain data imbalance for domain generalization.
\end{Itemize}

%% file: 3_related.tex
\paragraph{Long-Tailed Recognition.}
The literature is rich with research on long-tailed recognition~\cite{liu2019large,zhang2021survey}. 
Proposed solutions include re-balancing the data by either over-sampling the minority classes or under-sampling the majority classes~\cite{chawla2002smote,he2008adasyn}, re-weighting or adjusting the loss functions \cite{huang2019deep,cui2019class,cao2019learning,dong2019imbalanced}, as well as leveraging relevant learning paradigms such as transfer learning \cite{liu2019large}, metric learning \cite{zhang2017range}, meta-learning \cite{shu2019meta}, two-stage training \cite{kang2020decoupling}, ensemble learning \cite{wang2021ride,zhang2021testagnostic}, and self-supervised learning \cite{yang2020rethinking,li2021targeted}.
Recent studies have also explored imbalanced regression \cite{yang2021delving}.
In contrast to these past works, we extend long-tailed recognition to the multi-domain setting, and introduce new techniques suitable for learning from multi-domain imbalanced data.

\paragraph{Multi-Domain Learning.}
Multi-domain learning (MDL) aims to learn a model of minimal risk from datasets drawn from different underlying distributions \cite{dredze2010multi}, and is a specific case of transfer learning \cite{pan2009survey}. In contrast to domain adaptation (DA) \cite{ben2010theory,pan2009survey}, which aims to minimize the risk over a single ``target'' domain, MDL minimizes the risk over all ``source'' domains, and considers both average and worst risks over all distributions \cite{schoenauer2019multi}.
Past solutions for MDL include designing shared and domain-specific models \cite{dredze2010multi,xiao2016learning}, leveraging multi-task learning \cite{yang2015unified}, and learning domain-invariant features \cite{schoenauer2019multi,sun2016coral,li2018cdann,ganin2016dann}. Our work falls under the MDL framework, but considers the  practical and realistic setting where the label distribution is imbalanced within each domain and across domains.

\paragraph{Domain Generalization.}
Unlike MDL which focuses on in-domain generalization, domain generalization (DG) aims to learn from multiple training domains and generalize to unseen domains \cite{zhou2021survey}.
Previous approaches include learning domain-invariant features \cite{muandet2013icml_DIFL,ganin2016dann,li2018cdann}, learning transferable model parameters using meta-learning \cite{li2018mldg,zhang2020arm}, data augmentation \cite{zhou2021mixstyle,carlucci2019jigsaw_jigen}, and capturing causal relationships \cite{arjovsky2019irm,krueger2020vrex}.
Past work on DG has not investigated label imbalance within a domain and across domains. This paper shows that label imbalance plays a crucial role in DG, and that by combating data imbalance, we substantially boost DG performance on standard benchmarks.

%% file: 4_trans_graph.tex
When learning from MDLT, a natural question arises:
\begin{center}
\emph{How do we model MDLT in the presence of both \textbf{domain shift} and\\ \textbf{class imbalance} within and across domains?}
\end{center}

We argue that in contrast to single-domain imbalanced learning where the basic unit one cares about is a \emph{class} (i.e., minority \emph{vs.} majority classes), in MDLT, the basic unit naturally translates to a \textbf{domain-class pair}.

\paragraph{Problem Setup.}
Given a multi-domain classification task with a discrete label space $\mathcal{C}=\{1,\dots,C\}$ and a domain space $\mathcal{D}=\{1,\dots,D\}$, let $\mathcal{S} = \{ ( \mathbf{x}_i, c_i, d_i )\}_{i=1}^{N}$ be the training set, where $\mathbf{x}_i\in\real^{l}$ denotes the input, $c_i\in\mathcal{C}$ is the class label, and $d_i\in \mathcal{D}$ is the domain label.
We denote as $\mathbf{z} = f(\mathbf{x};\theta)$ the representation of $\mathbf{x}$, where $f:\mathcal{X}\rightarrow\mathcal{Z}$ maps the input into a  representation space $\mathcal{Z}\subseteq \real^{h}$. 
The final prediction $\hat{c}=g(\mathbf{z})$ is given by a classification function $g:\mathcal{Z}\rightarrow\mathcal{C}$.
We denote the set of samples belonging to domain $d$ and class $c$ (i.e., the domain-class pair $(d,c)$) as $\mathcal{S}_{d,c} \subseteq \mathcal{S}$, with $N_{d,c}\triangleq|\mathcal{S}_{d,c}|$ as the number of samples. Similarly, $\mathcal{Z}_{d,c} \subseteq \mathcal{Z}$ denotes the representation set for $(d,c)$. We use $\mathcal{M} = \mathcal{D}\times \mathcal{C} := \{ (d,c): d\in \mathcal{D}, c\in\mathcal{C} \}$ to denote the set of all domain-class pairs.

\begin{tcolorbox}
\begin{definition}[Transferability]
\label{defn:trans}
Given a learned model and a distance function $\dist:\mathbb{R}^{h} \times \mathbb{R}^{h} \rightarrow \mathbb{R}$ in the feature space, the transferability from domain-class pair $(d,c)$ to $(d',c')$ is:
\begin{align*}
\textnormal{trans}\big( (d,c), (d',c') \big)
\triangleq \E_{\mathbf{z}\in\mathcal{Z}_{d,c}} \big[ \dist\left(\mathbf{z}, \boldsymbol{\mu}_{d',c'}\right) \big],
\end{align*}
where $\boldsymbol{\mu}_{d',c'} \triangleq \E_{\mathbf{z}'\in\mathcal{Z}_{d',c'}} [ \mathbf{z}' ]$ is the first order statistics (i.e., mean) of $(d',c')$.
\end{definition}
\end{tcolorbox}

\begin{figure}[tb]
\begin{center}
\includegraphics[width=\linewidth]{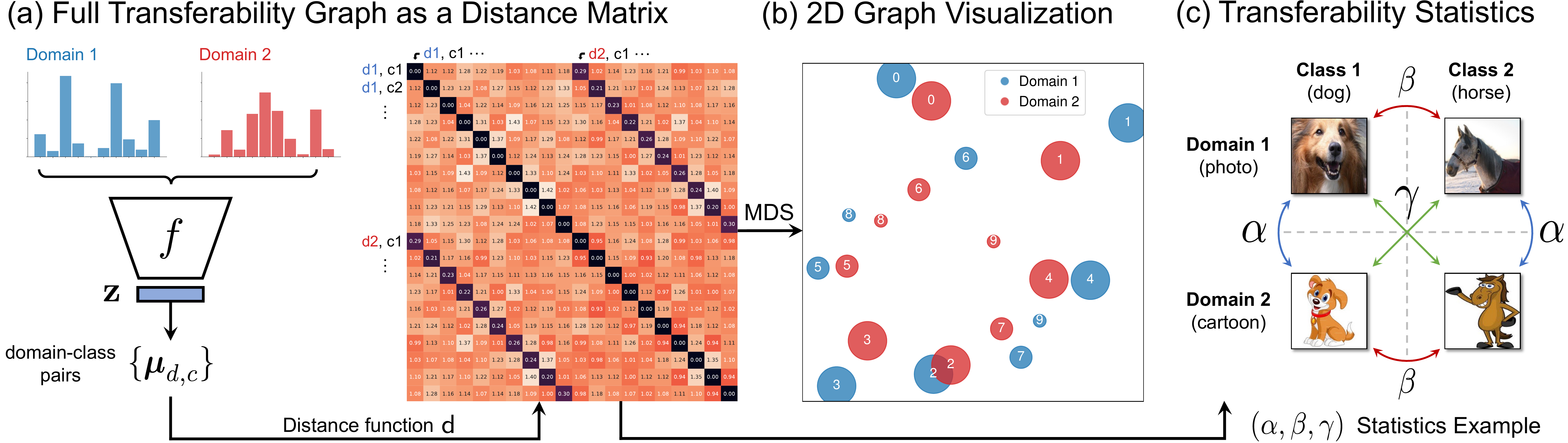}
\end{center}
\vspace{-0.4cm}
\caption{Overall framework of transferability graph. \textbf{(a)} Distribution statistics $\{\boldsymbol{\mu}_{d,c}\}$ is computed for all domain-class pairs, by which we generate a full transferability matrix. \textbf{(b)} MDS is used to project the graph into a 2D space for visualization. \textbf{(c)} We define $(\alpha,\beta,\gamma)$ transferability statistics to further describe the whole transferability graph.}
\label{fig:teaser-trans-graph}
\end{figure}

Intuitively, the transferability between two domain-class pairs is the average distance between their learned representations, characterizing how close they are in the feature space. By default, $\dist$ is chosen as the Euclidean distance, but it can also represent the higher order statistics of $(d,c)$. For example, the Mahalanobis distance \cite{de2000mahalanobis} uses the covariance $\boldsymbol{\Sigma}_{d,c} \triangleq \E_{\mathbf{z}\in\mathcal{Z}_{d,c}} \left[(\mathbf{z} - \boldsymbol{\mu}_{d,c})(\mathbf{z} - \boldsymbol{\mu}_{d,c})^\top \right]$.
In the remainder of the paper, with a slight abuse of the notation, we allow $\boldsymbol{\mu}_{d,c}$ to represent both the first and higher order statistics for $(d,c)$.

\begin{tcolorbox}
\begin{definition}[Transferability Graph]
\label{defn:trans_graph}
The transferability graph for a learned model is defined as $\mathcal{G} = \left( \mathcal{V}, \mathcal{E} \right)$, where the vertices, $\mathcal{V} \subseteq \{\boldsymbol{\mu}_{d,c} \}$, represents the domain-class pairs, and the edges, $\mathcal{E} \subseteq \mathcal{V}\times \mathcal{V}$, are assigned weights equal to $\textnormal{trans}\left( (d,c), (d',c') \right)$.
\end{definition}
\end{tcolorbox}

\paragraph{Transferability Graph Visualization.} It is convenient to visualize the transferability graph of a learned model in a 2D Cartesian space. To do so, we use the average of $\textnormal{trans}\left( (d,c), (d',c') \right)$ and $\textnormal{trans}\left( (d',c'), (d,c) \right)$ as a similarity measure between them. We can then visualize this similarity and the underlying transferability graph using multidimensional scaling (MDS)~\cite{carroll1998multidimensional}.
Figs.~\ref{fig:teaser-trans-graph}a and \ref{fig:teaser-trans-graph}b show this process, where for each $(d,c)$ pair, we estimate its distribution statistics $\{\boldsymbol{\mu}_{d,c}\}$ from the learned model and compute the transferability graph as a distance matrix. We then use MDS to project it into a 2D space, where each dot refers to one $(d,c)$, and the distance represents transferability.

\begin{tcolorbox}
\begin{definition}[$(\alpha,\beta,\gamma)$ Transferability Statistics]
\label{defn:trans_stats}
The transferability graph can be summarized by the following transferability statistics:
\begin{align*}
\textrm{Different domains, same class:} \quad & \alpha = \E_{c} \E_{d} \E_{d'\neq d} \left[ \textnormal{trans} \big( (d,c), (d',c) \big) \right]. \\
\textrm{Same domain, different classes:} \quad & \beta = \E_{d} \E_{c} \E_{c'\neq c} \left[ \textnormal{trans} \big( (d,c), (d,c') \big) \right]. \\
\textrm{Different domains, different classes:} \quad & \gamma = \E_{d} \E_{d'\neq d} \E_{c} \E_{c'\neq c} \left[ \textnormal{trans} \big( (d,c), (d',c') \big) \right].
\end{align*}
\end{definition}
\end{tcolorbox}
As illustrated in Fig.~\ref{fig:teaser-trans-graph}c, $(\alpha,\beta,\gamma)$ captures the similarity between features of the same class across domains and different classes within and across domains.

%% file: 5_rep_learn.tex
\subsection{Divergent Label Distributions Hamper Transferable Features}
MDLT has to deal with differences between the label distributions across domains.  To understand the implications of this issue we start with an example.

\paragraph{Motivating Example.}
We construct \texttt{Digits-MLT}, a two-domain toy MDLT dataset that combines two digit datasets: MNIST-M \cite{ganin2016dann} and SVHN \cite{netzer2011svhn}. The task is 10-class digit classification.
Details of the datasets are in Appendix \ref{sec-appendix:mdlt-dataset-details}.
We manually vary the number of samples for each domain-class pair to simulate different label distributions, and train a plain ResNet-18 \cite{he2016deep} using empirical risk minimization (ERM) for each case. We keep all test sets balanced and identical.

\begin{figure}[tb]
\begin{center}
\includegraphics[width=\linewidth]{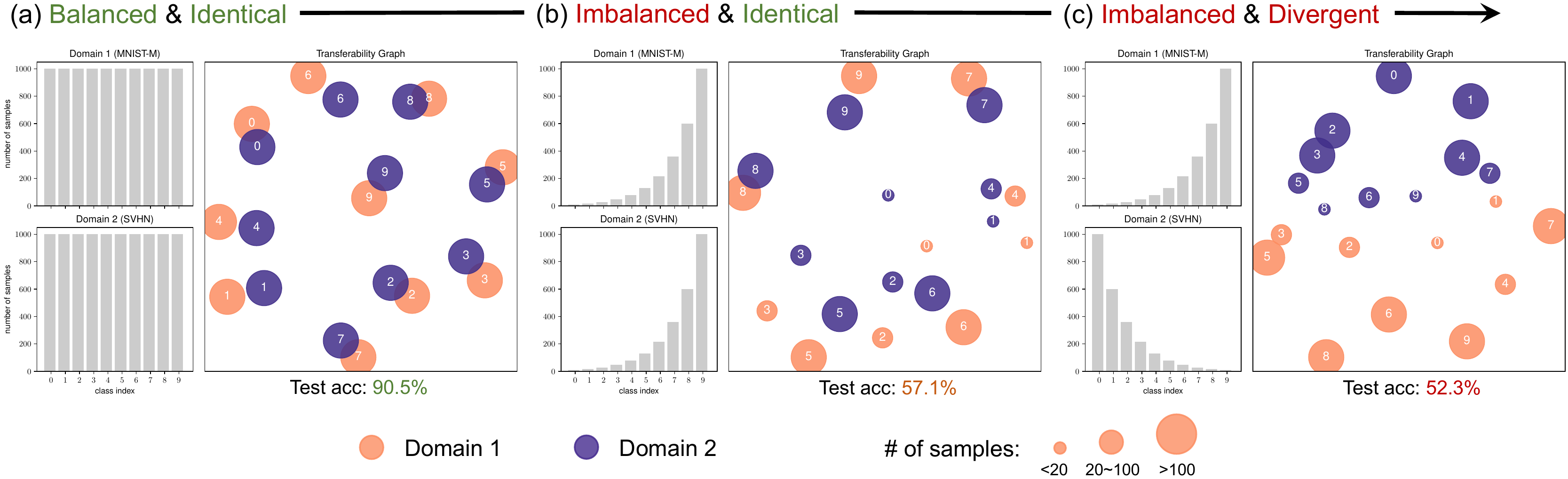}
\end{center}
\vspace{-0.4cm}
\caption{The evolving pattern of transferability graph when varying label proportions of \texttt{Digits-MLT}. \textbf{(a)} Label distributions for two domains are balanced and identical. \textbf{(b)} Label distributions for two domains are imbalanced but identical. \textbf{(c)} Label distributions for two domains are imbalanced and \emph{divergent}.}
\label{fig:motivate-divergence}
\end{figure}

The results in Fig.~\ref{fig:motivate-divergence} reveal interesting observations. When the per-domain label distributions are balanced and \emph{identical} across domains, although a domain gap exists, it does not prohibit the model from learning discriminative features of high accuracy (90.5\%), as shown in Fig.~\ref{fig:motivate-divergence}a.
If the label distributions are imbalanced but \emph{identical}, as in Fig.~\ref{fig:motivate-divergence}b, ERM is still able to align similar classes in the two domains, where majority classes (e.g., class 9) are closer in terms of transferability than minority classes (e.g., class 0). In contrast, when the labels are both imbalanced and \emph{mismatched} across domains, as in Fig.~\ref{fig:motivate-divergence}c, the learned features are no longer transferable, resulting in a clear gap across domains and the worst accuracy. This is because \emph{divergent label distributions} across domains produce an undesirable shortcut; the model can minimize the classification loss simply by separating the two domains.

\paragraph{Transferable Features are Desirable.}
As the results indicate, \emph{transferable} features across $(d,c)$ pairs are needed, especially when imbalance occurs. In particular, the transferability link between the same class across domains should be greater than that between different classes within or across domains. This can be captured via the $(\alpha,\beta,\gamma)$ transferability statistics, as we show next.

\subsection{Transferability Statistics Characterize Generalization}

\paragraph{Motivating Example.}
Again, we use \texttt{Digits-MLT} with varying label distributions. We consider three imbalance types to compose different label configurations: (1) \textbf{Uniform} (i.e., balanced labels), (2) \textbf{Forward-LT}, where the labels exhibit a long tail over class ids, and (3) \textbf{Backward-LT}, where labels are inversely long-tailed with respect to the class ids. For each configuration, we train 20 ERM models with varying hyperparameters. We then calculate the $(\alpha,\beta,\gamma)$ statistics for each model, and plot its classification accuracy against $(\beta+\gamma) - \alpha$.

\begin{figure}[tb]
\begin{center}
\includegraphics[width=\linewidth]{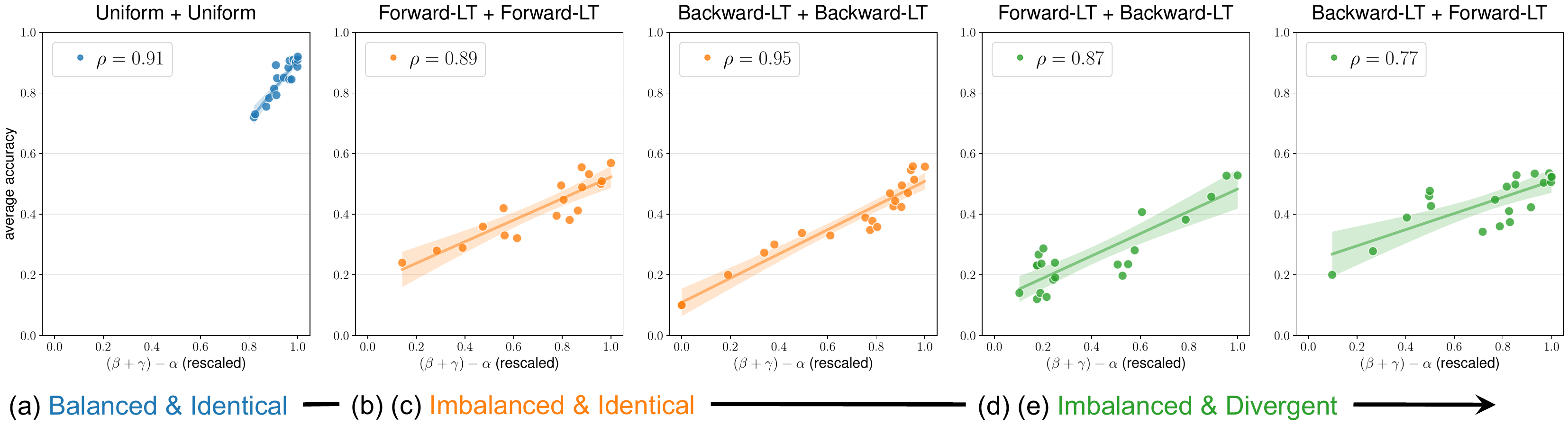}
\end{center}
\vspace{-0.4cm}
\caption{Correspondence between $(\beta+\gamma) - \alpha$ quantity and test accuracy across different label configurations of \texttt{Digits-MLT}. Each plot refers to specific label distributions for two domains (e.g., (a) employs ``Uniform'' for domain 1 and ``Uniform'' for domain 2). Each point corresponds to a model trained with ERM using different hyperparameters.}
\label{fig:motivate-stats}
\end{figure}

Fig.~\ref{fig:motivate-stats} reveals the following findings:
(1) \emph{The $(\alpha,\beta,\gamma)$ statistics characterize a model's performance in MDLT.} In particular, the $(\beta+\gamma) - \alpha$ quantity displays a very strong correlation with test performance across the entire range and every label configuration.
(2) \emph{Data imbalance increases the risk of learning less transferable features.} When the label distributions are similar across domains (Fig.~\ref{fig:motivate-stats}a), the models are robust to varying parameters, clustering in the upper-right region. However, as the labels become imbalanced (Figs.~\ref{fig:motivate-stats}b, \ref{fig:motivate-stats}c) and further divergent (Figs.~\ref{fig:motivate-stats}d, \ref{fig:motivate-stats}e), chances that the model learns non-transferable features (i.e., lower $(\beta+\gamma) - \alpha$) increase, leading to a large drop in performance.
We provide further evidence in Appendix~\ref{subsec-appendix:stats-corr-real-datasets} showing that these observations hold regardless of datasets and training regimes.

\subsection{A Loss that Bounds the Transferability Statistics}

We use the above findings to design a new loss function particularly suitable for MDLT. We will first introduce the loss function then prove that it minimizes an upper bound of the $(\alpha,\beta,\gamma)$ statistics.  We start from a simple loss inspired by the metric learning objective \cite{goldberger2004neighbourhood,sohn2016npairloss}.
We call this loss $\mathcal{L}_{\texttt{DA}}$ since it aims for \underline{D}omain-Class Distribution \underline{A}lignment, i.e., aligning the features of the same class across domains. 
Let $(\mathbf{x}_i,c_i,d_i)$ denote a sample with feature $\mathbf{z}_i$.
Given a set of training samples with feature set $\mathcal{Z}$, we have
\begin{equation}
\label{eqn:initial-da-loss}
\mathcal{L}_{\texttt{DA}}(\mathcal{Z}, \{\boldsymbol{\mu}\}) 
= \sum_{\mathbf{z}_i\in \mathcal{Z}} \frac{-1}{|\mathcal{D}|-1} \sum_{d\in \mathcal{D}\setminus \{d_i\}} \log \frac{\exp{(- \dist(\mathbf{z}_i, \boldsymbol{\mu}_{d,c_i}))}}{\sum_{(d',c') \in \mathcal{M} \setminus \{(d_i, c_i)\}} \exp{(- \dist(\mathbf{z}_i, \boldsymbol{\mu}_{d',c'}))}}.
\end{equation}

Intuitively, $\mathcal{L}_{\texttt{DA}}$ tackles label \emph{divergence}, as $(d,c)$ pairs that share same class would be pulled closer, and vice versa. It is also related to $(\alpha,\beta,\gamma)$ statistics, as the numerator represents \emph{positive} cross-domain pairs ($\alpha$), and the denominator represents \emph{negative} cross-class pairs ($\beta,\gamma$).
A detailed probabilistic interpretation of $\mathcal{L}_{\texttt{DA}}$ is provided in Appendix~\ref{subsec-appendix:prob-derivation-da}.

But, $\mathcal{L}_{\texttt{DA}}$ does not address label \emph{imbalance}. Note that $(\alpha,\beta,\gamma)$ is defined in a \emph{balanced} way, independent of the number of samples of each $(d,c)$. However, given an imbalanced dataset, most samples will come from  majority domain-class pairs, which would dominate $\mathcal{L}_{\texttt{DA}}$ and cause minority pairs to be overlooked.

\paragraph{\underline{B}alanced D\underline{o}main-Class \underline{D}istribution \underline{A}lignment (BoDA).}
To tackle data imbalance across $(d,c)$ pairs, we modify the loss in Eqn.~(\ref{eqn:initial-da-loss}) to the \boda loss:
\begin{equation}
\label{eqn:initial-boda-loss}
\resizebox{\textwidth}{!}{$
\mathcal{L}_{\boda}(\mathcal{Z}, \{\boldsymbol{\mu}\}) 
= \sum\limits_{\mathbf{z}_i\in \mathcal{Z}} \frac{-1}{|\mathcal{D}|-1} \sum\limits_{d\in \mathcal{D}\setminus \{d_i\}} \log \frac{\exp{(- \bdist(\mathbf{z}_i, \boldsymbol{\mu}_{d,c_i}))}}{\sum_{(d',c') \in \mathcal{M} \setminus \{(d_i, c_i)\}} \exp{(- \bdist(\mathbf{z}_i, \boldsymbol{\mu}_{d',c'}))}},~~
\bdist(\mathbf{z}_i, \boldsymbol{\mu}_{d,c}) = \frac{\dist(\mathbf{z}_i, \boldsymbol{\mu}_{d,c})}{N_{d_i,c_i}}.
$}
\end{equation}

\boda scales the original $\dist$ by a factor of $1/N_{d_i,c_i}$, i.e., it counters the effect of imbalanced domain-class pairs by introducing a \emph{balanced} distance measure $\bdist$.

\begin{tcolorbox}
\begin{theorem}[$\mathcal{L}_{\boda}$ as an Upper Bound]
\label{thm:boda-bound}
Given a multi-domain long-tailed dataset $\mathcal{S}$ with domain label space $\mathcal{D}$ and class label space $\mathcal{C}$ satisfying $|\mathcal{D}| > 1$ and $|\mathcal{C}| > 1$, let $\mathcal{Z}$ be the representation set of all training samples, and $(\alpha, \beta,\gamma)$ be the transferability statistics for $\mathcal{S}$ defined in Definition~\ref{defn:trans_stats}. It holds that
\begin{equation}
\mathcal{L}_{\textnormal{\boda}}(\mathcal{Z},\{\boldsymbol{\mu}\}) \geq N \log \left( |\mathcal{D}|-1 + |\mathcal{D}| (|\mathcal{C}|-1) \exp{\left(
\frac{|\mathcal{C}| |\mathcal{D}|}{N} \cdot \alpha - \frac{|\mathcal{C}|}{N} \cdot \beta - \frac{|\mathcal{C}| (|\mathcal{D}|-1)}{N} \cdot \gamma
\right) } \right).
\label{eqn:boda-bound}
\end{equation}
\end{theorem}
\end{tcolorbox}

\noindent
The proof of Theorem~\ref{thm:boda-bound} is in Appendix~\ref{subsec-appendix:proof-thm-1}.
Theorem~\ref{thm:boda-bound} has the following interesting implications: 
(1) \emph{$\mathcal{L}_\textnormal{\boda}$ upper-bounds $(\alpha,\beta,\gamma)$ statistics in a desired form that naturally translates to better performance.}
By minimizing $\mathcal{L}_{\boda}$, we ensure a low $\alpha$ (attract same classes) and high $\beta,\gamma$ (separate different classes), which are essential conditions for generalization in MDLT.
(2) \emph{The constant factors correspond to how much each component contributes to the transferability graph.}
Zooming on the arguments of $\exp(\cdot)$, we observe that the objective is proportional to $\alpha - (\frac{1}{|\mathcal{D}|}\beta + \frac{|\mathcal{D}|-1}{|\mathcal{D}|}\gamma)$. 
According to \defref{defn:trans_stats}, we note that $\alpha$ summarizes data similarity for the same class, while $(\frac{1}{|\mathcal{D}|}\beta + \frac{|\mathcal{D}|-1}{|\mathcal{D}|}\gamma)$ summarizes data similarity across different classes, using the weighted average of $\beta$ and $\gamma$, where their weights are proportional to the number of associated domains (i.e., $1$ for $\beta$, $(|\mathcal{D}|-1)$ for $\gamma$).

\subsection{Calibration for Data Imbalance Leads to Better Transfer}

\boda works by encouraging feature transfer for similar classes across domains, i.e., if $(d,c)$ and $(d',c)$ refer to the same class in different domains, then we want to transfer their features to each other.
But, minority domain-class pairs naturally have worse $\boldsymbol{\mu}_{d,c}$ estimates due to data scarcity, and forcing other pairs to transfer to them hurts learning. Thus, when bringing two domain-class pairs closer in the embedding space, we want the minority $(d,c)$ to transfer to majority ones, not the inverse. The following example further clarifies this point.

\begin{figure}[tb]
\begin{center}
\includegraphics[width=0.99\linewidth]{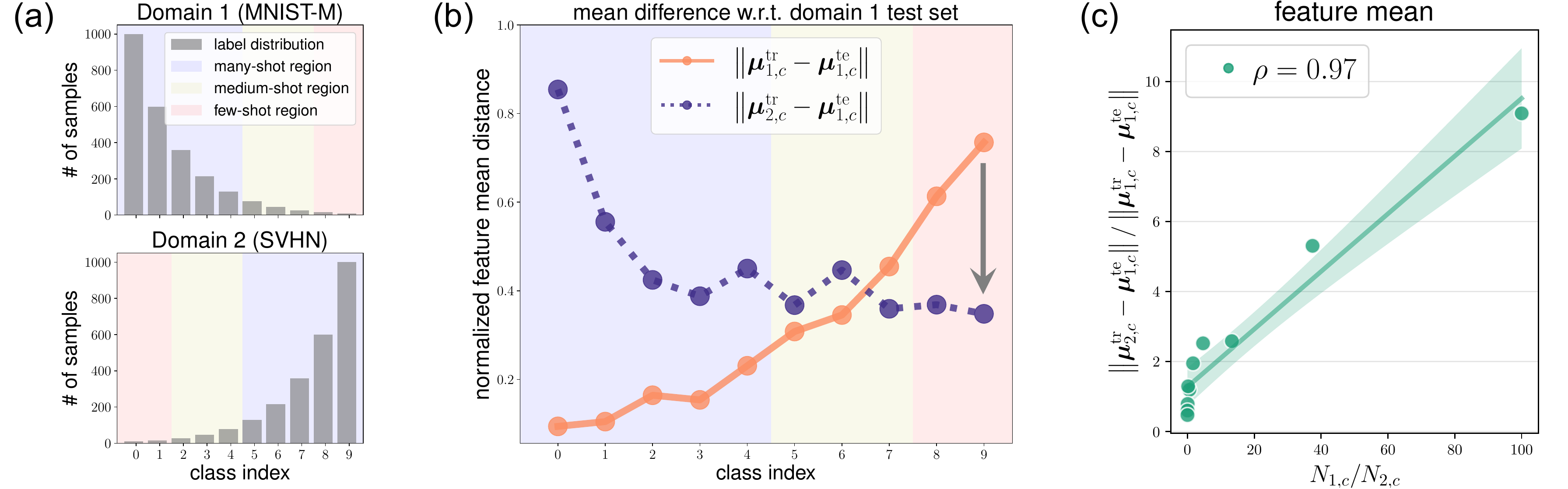}
\end{center}
\vspace{-0.4cm}
\caption{
The need for \emph{calibration}. \textbf{(a)} Per-domain label distribution of \texttt{Digits-MLT}. \textbf{(b)} Distance between training and test data. \textcolor{myorange}{Solid line} plots the distance between training and test data from the same domain-class pairs. \textcolor{mypurple}{Dashed line} plots the distance between test data from a particular domain-class pair and the training data with which it shares the same class but differs in the domain. The \textcolor{manyshot}{blue} and \textcolor{fewshot}{red} background colors refer to majority and minority domain-class pairs, respectively. 
\textbf{(c)} Correspondence between the ratio of the sample size and their feature distances between testing and training across different domain-class pairs.
}
\label{fig:motivate-feat-stats}
\end{figure}

\paragraph{Motivating Example.}
We use \texttt{Digits-MLT} with divergent labels (Fig.~\ref{fig:motivate-feat-stats}). We focus on \emph{feature discrepancy}, i.e., the distance between training and test features for the same class. For each class in domain 1, we compute the distance in the feature space between the means of the training set and test set (\textcolor{myorange}{solid line}). We also compute the distance between the training data of domain 2 and test data of domain 1 (\textcolor{mypurple}{dashed line}), for the same class.

As shown by the solid orange line in Fig.~\ref{fig:motivate-feat-stats}b, for minority domain-class pairs such as class ``8'' and ``9'' in domain 1, the distance in the feature space between training and testing is large. In fact, the test set of these minority domain-class pairs is closer to the training data for ``8'' and ``9'' in domain 2 than in their own domain, as shown by the dashed purple line. This example indicates that a better training would try to transfer the features of minority domain-class pairs to majority pairs with which they share the same class, as shown by the grey arrow in Fig.~\ref{fig:motivate-feat-stats}b. Such transfer will improve generalization to the test set.

\paragraph{\boda with Calibrated Distance.}
The above discussion motivates a modification to \boda to favor transfer to majority domain-class pairs:
\begin{equation}
\label{eqn:final-boda-loss}
\resizebox{\textwidth}{!}{$
\mathcal{\tilde{L}}_{\boda}(\mathcal{Z}, \{\boldsymbol{\mu}\})
= \sum\limits_{\mathbf{z}_i\in \mathcal{Z}} \frac{-1}{|\mathcal{D}|-1} \sum\limits_{d\in \mathcal{D}\setminus \{d_i\}} \log \frac{\exp{\left(- \lambda^{d,c_i}_{d_i,c_i} \bdist(\mathbf{z}_i, \boldsymbol{\mu}_{d,c_i})\right)}}{\sum_{(d',c') \in \mathcal{M} \setminus \{(d_i, c_i)\}} \exp{\left(- \lambda^{d',c'}_{d_i,c_i} \bdist(\mathbf{z}_i, \boldsymbol{\mu}_{d',c'})\right)}},~~
\lambda^{d',c'}_{d,c} = \left(\frac{N_{d',c'}}{N_{d,c}}\right)^{\nu},
$}
\end{equation}
where $\nu$ is a constant that allows for a sublinear relation (default $\nu=1$). $\lambda^{d',c'}_{d,c}$ indicates how much we would like to transfer $(d,c)$ to $(d',c')$, based on their relative sample size. Fig.~\ref{fig:motivate-feat-stats}c verifies that the ratio of the sample size is highly correlated with the ratio of the distance between testing and training. Further, Theorem~\ref{thm:calibrated-boda-bound} in Appendix~\ref{sec-appendix:proof} shows that $\mathcal{\tilde{L}}_{\boda}$ is an upper bound of the calibrated transferability statistics.

\paragraph{Variants of \boda: Matching Higher Order Statistics.}
The distance $\dist$ can be set to the Euclidean distance $\dist(\mathbf{z}, \boldsymbol{\mu}_{d,c}) = \sqrt{(\mathbf{z} - \boldsymbol{\mu}_{d,c})^\top(\mathbf{z} - \boldsymbol{\mu}_{d,c})}$, which captures the first order statistics. To match higher order statistics such as covariance, $\dist(\mathbf{z}, \{ \boldsymbol{\mu}_{d,c}, \boldsymbol{\Sigma}_{d,c} \}) = \sqrt{(\mathbf{z} - \boldsymbol{\mu}_{d,c})^\top \boldsymbol{\Sigma}_{d,c}^{-1} (\mathbf{z} - \boldsymbol{\mu}_{d,c})}$ is used, resembling the Mahalanobis distance~\cite{de2000mahalanobis}. We refer to these variants as $ \mathcal{\tilde{L}}_{\boda}$ and $\mathcal{\tilde{L}}_{\texttt{BoDA-M}}$.

\paragraph{Joint Loss.}
\boda serves as a representation learning scheme for MDLT, which operates over $\mathcal{Z}$. For classification, we train deep networks by combining $\mathcal{\tilde{L}}_\texttt{\boda}$ and the standard cross-entropy (CE) loss in an end-to-end fashion, where CE is applied to the output layer, and \boda is applied to the latent features. We combine the losses as $\mathcal{L}_\texttt{CE} + \omega \mathcal{\tilde{L}}_\texttt{\boda}$, with $\omega$ as a trade-off hyperparameter.

%% file: 6_cls_learn.tex
In the long-tailed recognition literature, an important finding is that decoupling \emph{representation learning} and \emph{classifier learning} leads to better results \cite{kang2020decoupling,zhou2020bbn}.
In particular, instance-balanced sampling is used during the first stage of learning, while class-balanced sampling is used for re-training the classifier (with the representation fixed) in the second stage \cite{kang2020decoupling}.
Motivated by this, we explore whether a similar decoupling benefits MDLT. We use three learning algorithms, ERM \cite{vapnik1999overview}, DANN \cite{li2018cdann}, and CORAL \cite{sun2016coral}.
We train each algorithm with and without the second stage classifier learning, and report the average accuracy over all MDLT datasets (presented later).

\setlength\intextsep{0pt}
\begin{wraptable}[7]{r}{0.4\textwidth}
\setlength{\tabcolsep}{6pt}
\caption{The benefits of decoupling the classifier.}
\vspace{-19pt}
\label{table:rep_cls_table}
\small
\begin{center}
\adjustbox{max width=0.4\textwidth}{
\begin{tabular}{lcc}
\toprule[1.5pt]
\textbf{Algorithm} & \textbf{w/o decouple} & \textbf{w/ decouple} \\ \midrule
ERM~\cite{vapnik1999overview} & 77.6 \scriptsize$\pm0.2$ & \textbf{79.2} \scriptsize$\pm0.3$ \\
DANN~\cite{ganin2016dann} & 77.7 \scriptsize$\pm0.6$ & \textbf{79.0} \scriptsize$\pm0.1$ \\
CORAL~\cite{sun2016coral} & 78.0 \scriptsize$\pm0.1$ & \textbf{79.6} \scriptsize$\pm0.2$ \\
\bottomrule[1.5pt]
\end{tabular}}
\end{center}
\end{wraptable}

As Table~\ref{table:rep_cls_table} shows, similar to what has been observed in the single domain case \cite{kang2020decoupling,zhou2020bbn}, regardless of algorithm, decoupling the classifier learning consistently improves performance.
Since \boda can support both coupled and decoupled classifier learning, we use $\boda_{r}$ to refer to models that couple representation and classifier learning, and $\boda_{r,c}$ for models that decouple representation from classifier learning. In the classifier learning stage, we simply use class-balanced sampling.

%% file: 7_exp_mdlt.tex
\begin{figure}[tb]
\centering
\includegraphics[width=\textwidth]{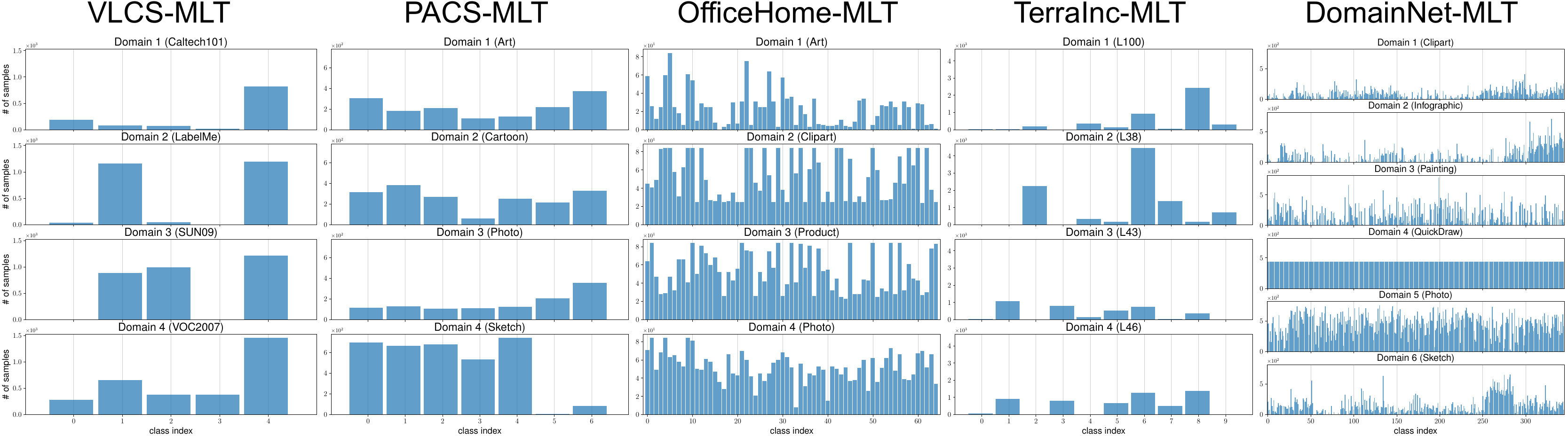}
\vspace{-0.6cm}
\caption{Overview of training set label distribution for five MDLT datasets. We set up MDLT benchmarks from datasets traditionally used for DG, and make validation/test sets balanced across all domain-class pairs. More details are provided in Appendix~\ref{sec-appendix:mdlt-dataset-details}.}
\label{fig:dataset-info}
\end{figure}

\paragraph{Datasets.}
We curate  five multi-domain datasets typically used in DG and adapt them for MDLT evaluation. To do so, for each dataset, we create two balanced datasets one for validation and the other for testing, and leave the rest for training. The size of the validation and test data sets is roughly 5\% and 10\% of original data, respectively. Table~\ref{appendix:table:dataset-details} in Appendix~\ref{sec-appendix:mdlt-dataset-details} provides the statistics of each MDLT dataset.
Fig.~\ref{fig:dataset-info} shows the label distributions across domains in the five datasets.

\begin{Enumerate}
    \item \texttt{VLCS-MLT}. We construct \texttt{VLCS-MLT} using the \texttt{VLCS} dataset \cite{fang2013vlcs}, which is an object recognition dataset with 10,729 images from 4 domains and 5 classes.
    \item \texttt{PACS-MLT}. \texttt{PACS-MLT} is constructed from the \texttt{PACS} dataset \cite{li2017pacs}, an object recognition dataset with 9,991 images from 4 domains and 7 classes.
    \item \texttt{OfficeHome-MLT}. We set up \texttt{OfficeHome-MLT} using the \texttt{OfficeHome} dataset \cite{venkateswara2017officehome} which contains 15,588 images from 4 domains and 65 classes.
    \item \texttt{TerraInc-MLT}. \texttt{TerraInc-MLT} is created from \texttt{TerraIncognita}  \cite{beery2018recognition}, a species classification dataset including 24,788 images from 4 domains and 10 classes.
    \item \texttt{DomainNet-MLT}. We construct \texttt{DomainNet-MLT} using \texttt{DomainNet}  \cite{peng2019domainnet}, a large-scale multi-domain dataset for object recognition. It contains 586,575 images from 345 classes and 6 domains.
\end{Enumerate}

\paragraph{Network Architectures.}
For experiments on the synthetic \texttt{Digits-MLT} dataset, we use a simple CNN architecture as in \cite{gulrajani2020domainbed}. For the MDLT datasets, we follow \cite{gulrajani2020domainbed}, and use ResNet-50 \cite{he2016deep} for all algorithms.

\paragraph{Competing Algorithms.}
We compare \boda to a large number of algorithms that span different learning strategies and categories, including 
(1) \emph{vanilla:} \textbf{ERM} \cite{vapnik1999overview}, 
(2) \emph{distributionally robust optimization:} \textbf{GroupDRO} \cite{sagawa2020groupdro}, 
(3) \emph{data augmentation:} \textbf{Mixup}~\cite{xu2020interdomain_mixup_aaai}, \textbf{SagNet} \cite{nam2019sagnet}, 
(4) \emph{meta-learning:} \textbf{MLDG} \cite{li2018mldg}, 
(5) \emph{domain-invariant feature learning:} \textbf{IRM} \cite{arjovsky2019irm}, \textbf{DANN} \cite{ganin2016dann}, \textbf{CDANN} \cite{li2018cdann}, \textbf{CORAL} \cite{sun2016coral}, \textbf{MMD} \cite{li2018mmd}, 
(6) \emph{transfer learning:} \textbf{MTL} \cite{blanchard2021mtl_marginal_transfer_learning}, 
(7) \emph{multi-task learning:} \textbf{Fish} \cite{shi2021fish}, and 
(8) \emph{imbalanced learning:} \textbf{Focal} \cite{lin2017focal}, \textbf{CBLoss} \cite{cui2019class}, \textbf{LDAM} \cite{cao2019learning}, \textbf{BSoftmax} \cite{ren2020bsoftmax}, \textbf{SSP} \cite{yang2020rethinking}, \textbf{CRT} \cite{kang2020decoupling}.
We provide detailed descriptions in Appendix~\ref{subsec-appendix:all-algo-details}.

\paragraph{Implementation and Evaluation Metrics.}
For a fair evaluation, following \cite{gulrajani2020domainbed}, for each algorithm we conduct a random search of 20 trials over a joint distribution of all hyperparameters (see Appendix \ref{subsec-appendix:hp-details} for details). We then use the validation set to select the best hyperparameters for each algorithm, fix them and rerun the experiments under three different random seeds to report the final average accuracy with standard deviation. Such process ensures the comparison is best-versus-best, and the hyperparameters are optimized for all algorithms.
In addition to the average accuracy across domains, we also report the worst accuracy over domains, and further divide all domain-class pairs into \emph{many-shot} (pairs with over 100 training samples), \emph{medium-shot} (pairs with 20$\sim$100 training samples), \emph{few-shot} (pairs with under 20 training samples), and \emph{zero-shot} (pairs with no training data), and report the results for these subsets.

\begin{table}[t]
\setlength{\tabcolsep}{2.5pt}
\parbox{.5\linewidth}{
\caption{Results on \texttt{VLCS-MLT}.}
\vspace{-7pt}
\label{table:vlcs-mlt}
\small
\begin{center}
\adjustbox{max width=0.493\textwidth}{
\begin{tabular}{lcccccc}
\toprule[1.5pt]
& \multicolumn{2}{c}{\textbf{Accuracy (by domain)}} & \multicolumn{4}{c}{\textbf{Accuracy (by shot)}} \\
\cmidrule(lr){2-3} \cmidrule(lr){4-7}
\textbf{Algorithm} & \textbf{Average} & \textbf{Worst} & \textbf{Many} & \textbf{Medium} & \textbf{Few} & \textbf{Zero} \\
\midrule
ERM~\cite{vapnik1999overview} & 76.3 \scriptsize$\pm0.4$ & 53.6 \scriptsize$\pm1.1$ & 84.6 \scriptsize$\pm0.5$ & 76.6 \scriptsize$\pm0.4$ & $-$ & 32.9 \scriptsize$\pm0.4$ \\
IRM~\cite{arjovsky2019irm} & 76.5 \scriptsize$\pm0.2$ & 52.3 \scriptsize$\pm0.7$ & 85.3 \scriptsize$\pm0.6$ & 75.5 \scriptsize$\pm1.0$ & $-$ & 33.5 \scriptsize$\pm1.0$ \\
GroupDRO~\cite{sagawa2020groupdro} & 76.7 \scriptsize$\pm0.4$ & 54.1 \scriptsize$\pm1.3$ & 85.3 \scriptsize$\pm0.9$ & 76.2 \scriptsize$\pm1.0$ & $-$ & 34.5 \scriptsize$\pm2.0$ \\
Mixup~\cite{xu2020interdomain_mixup_aaai} & 75.9 \scriptsize$\pm0.1$ & 52.7 \scriptsize$\pm1.3$ & 84.4 \scriptsize$\pm0.2$ & 77.1 \scriptsize$\pm0.6$ & $-$ & 29.2 \scriptsize$\pm1.4$ \\
MLDG~\cite{li2018mldg} & 76.9 \scriptsize$\pm0.2$ & 53.6 \scriptsize$\pm0.5$ & 84.9 \scriptsize$\pm0.3$ & 77.5 \scriptsize$\pm1.0$ & $-$ & 34.4 \scriptsize$\pm0.9$ \\
CORAL~\cite{sun2016coral} & 75.9 \scriptsize$\pm0.5$ & 51.6 \scriptsize$\pm0.7$ & 84.3 \scriptsize$\pm0.6$ & 75.5 \scriptsize$\pm0.5$ & $-$ & 34.5 \scriptsize$\pm0.8$ \\
\end{tabular}}
\end{center}
}
\hfill
\parbox{.51\linewidth}{
\caption{Results on \texttt{PACS-MLT}.}
\vspace{-7pt}
\label{table:pacs-mlt}
\small
\begin{center}
\adjustbox{max width=0.505\textwidth}{
\begin{tabular}{lcccccc}
\toprule[1.5pt]
& \multicolumn{2}{c}{\textbf{Accuracy (by domain)}} & \multicolumn{4}{c}{\textbf{Accuracy (by shot)}} \\
\cmidrule(lr){2-3} \cmidrule(lr){4-7}
\textbf{Algorithm} & \textbf{Average} & \textbf{Worst} & \textbf{Many} & \textbf{Medium} & \textbf{Few} & \textbf{Zero} \\
\midrule
ERM~\cite{vapnik1999overview} & 97.1 \scriptsize$\pm0.1$ & 95.8 \scriptsize$\pm0.2$ & 97.1 \scriptsize$\pm0.0$ & 97.0 \scriptsize$\pm0.0$ & 98.0 \scriptsize$\pm0.9$ & $-$ \\
IRM~\cite{arjovsky2019irm} & 96.7 \scriptsize$\pm0.2$ & 95.2 \scriptsize$\pm0.4$ & 96.8 \scriptsize$\pm0.2$ & 96.7 \scriptsize$\pm0.7$ & 94.7 \scriptsize$\pm1.4$ & $-$ \\
GroupDRO~\cite{sagawa2020groupdro} & 97.0 \scriptsize$\pm0.1$ & 95.3 \scriptsize$\pm0.4$ & 97.3 \scriptsize$\pm0.1$ & 95.3 \scriptsize$\pm1.2$ & 94.7 \scriptsize$\pm3.6$ & $-$ \\
Mixup~\cite{xu2020interdomain_mixup_aaai} & 96.7 \scriptsize$\pm0.2$ & 95.1 \scriptsize$\pm0.2$ & 97.0 \scriptsize$\pm0.1$ & 96.7 \scriptsize$\pm0.3$ & 91.3 \scriptsize$\pm2.7$ & $-$ \\
MLDG~\cite{li2018mldg} & 96.6 \scriptsize$\pm0.1$ & 94.1 \scriptsize$\pm0.3$ & 96.8 \scriptsize$\pm0.1$ & 96.3 \scriptsize$\pm0.7$ & 92.7 \scriptsize$\pm0.5$ & $-$ \\
CORAL~\cite{sun2016coral} & 96.6 \scriptsize$\pm0.5$ & 94.3 \scriptsize$\pm0.7$ & 96.6 \scriptsize$\pm0.5$ & 97.0 \scriptsize$\pm0.8$ & 94.7 \scriptsize$\pm0.5$ & $-$ \\
\end{tabular}}
\end{center}
}
\vspace{-0.4cm}
\end{table}

\subsection{Main Results}

We report the main results in this section for all MDLT datasets. The complete results and all additional experiments are provided in Appendix \ref{sec-appendix:complete-results-mdlt} and \ref{sec-appendix:additional-study}.

\begin{table}[!t]
\setlength{\tabcolsep}{2.5pt}
\parbox{.503\linewidth}{
\caption{Results on \texttt{OfficeHome-MLT}.}
\vspace{-7pt}
\label{table:officehome-mlt}
\small
\begin{center}
\adjustbox{max width=0.5\textwidth}{
\begin{tabular}{lcccccc}
\toprule[1.5pt]
& \multicolumn{2}{c}{\textbf{Accuracy (by domain)}} & \multicolumn{4}{c}{\textbf{Accuracy (by shot)}} \\
\cmidrule(lr){2-3} \cmidrule(lr){4-7}
\textbf{Algorithm} & \textbf{Average} & \textbf{Worst} & \textbf{Many} & \textbf{Medium} & \textbf{Few} & \textbf{Zero} \\
\midrule
ERM~\cite{vapnik1999overview} & 80.7 \scriptsize$\pm0.0$ & 71.3 \scriptsize$\pm0.1$ & 87.8 \scriptsize$\pm0.2$ & 81.0 \scriptsize$\pm0.2$ & 63.1 \scriptsize$\pm0.1$ & 63.3 \scriptsize$\pm7.2$ \\
IRM~\cite{arjovsky2019irm} & 80.6 \scriptsize$\pm0.4$ & 70.7 \scriptsize$\pm0.2$ & 87.6 \scriptsize$\pm0.4$ & 81.5 \scriptsize$\pm0.4$ & 61.1 \scriptsize$\pm0.9$ & 56.7 \scriptsize$\pm1.4$ \\
GroupDRO~\cite{sagawa2020groupdro} & 80.1 \scriptsize$\pm0.3$ & 68.7 \scriptsize$\pm0.9$ & 88.1 \scriptsize$\pm0.2$ & 80.8 \scriptsize$\pm0.4$ & 59.8 \scriptsize$\pm1.2$ & 51.7 \scriptsize$\pm3.6$ \\
Mixup~\cite{xu2020interdomain_mixup_aaai} & 81.2 \scriptsize$\pm0.2$ & 72.3 \scriptsize$\pm0.6$ & 87.9 \scriptsize$\pm0.4$ & 81.8 \scriptsize$\pm0.1$ & 64.1 \scriptsize$\pm0.4$ & 60.0 \scriptsize$\pm4.1$ \\
MLDG~\cite{li2018mldg} & 80.4 \scriptsize$\pm0.2$ & 70.2 \scriptsize$\pm0.6$ & 87.1 \scriptsize$\pm0.1$ & 81.3 \scriptsize$\pm0.3$ & 61.3 \scriptsize$\pm1.0$ & 61.7 \scriptsize$\pm1.4$ \\
CORAL~\cite{sun2016coral} & 81.9 \scriptsize$\pm0.1$ & \textbf{72.7} \scriptsize$\pm0.6$ & 87.9 \scriptsize$\pm0.1$ & 83.0 \scriptsize$\pm0.1$ & 63.5 \scriptsize$\pm0.7$ & 65.0 \scriptsize$\pm2.4$ \\
\end{tabular}}
\end{center}
}
\hfill
\parbox{.503\linewidth}{
\caption{Results on \texttt{TerraInc-MLT}.}
\vspace{-7pt}
\label{table:terrainc-mlt}
\small
\begin{center}
\adjustbox{max width=0.5\textwidth}{
\begin{tabular}{lcccccc}
\toprule[1.5pt]
& \multicolumn{2}{c}{\textbf{Accuracy (by domain)}} & \multicolumn{4}{c}{\textbf{Accuracy (by shot)}} \\
\cmidrule(lr){2-3} \cmidrule(lr){4-7}
\textbf{Algorithm} & \textbf{Average} & \textbf{Worst} & \textbf{Many} & \textbf{Medium} & \textbf{Few} & \textbf{Zero} \\
\midrule
ERM~\cite{vapnik1999overview} & 75.3 \scriptsize$\pm0.3$ & 67.4 \scriptsize$\pm0.3$ & 85.6 \scriptsize$\pm0.8$ & 69.6 \scriptsize$\pm3.2$ & 66.1 \scriptsize$\pm2.4$ & 14.4 \scriptsize$\pm2.8$ \\
IRM~\cite{arjovsky2019irm} & 73.3 \scriptsize$\pm0.7$ & 64.3 \scriptsize$\pm1.3$ & 83.5 \scriptsize$\pm0.6$ & 70.0 \scriptsize$\pm1.8$ & 58.3 \scriptsize$\pm3.4$ & 20.1 \scriptsize$\pm1.4$ \\
GroupDRO~\cite{sagawa2020groupdro} & 72.0 \scriptsize$\pm0.4$ & 66.6 \scriptsize$\pm0.2$ & 84.7 \scriptsize$\pm1.1$ & 64.6 \scriptsize$\pm4.7$ & 38.9 \scriptsize$\pm1.2$ & 13.5 \scriptsize$\pm1.1$ \\
Mixup~\cite{xu2020interdomain_mixup_aaai} & 71.1 \scriptsize$\pm0.7$ & 60.4 \scriptsize$\pm1.1$ & 83.2 \scriptsize$\pm0.7$ & 60.0 \scriptsize$\pm0.6$ & 56.1 \scriptsize$\pm3.0$ & 12.2 \scriptsize$\pm2.1$ \\
MLDG~\cite{li2018mldg} & 76.6 \scriptsize$\pm0.2$ & 66.9 \scriptsize$\pm0.5$ & 86.1 \scriptsize$\pm0.6$ & 73.8 \scriptsize$\pm3.9$ & 70.6 \scriptsize$\pm3.7$ & 18.8 \scriptsize$\pm2.4$ \\
CORAL~\cite{sun2016coral} & 76.4 \scriptsize$\pm0.5$ & 67.8 \scriptsize$\pm0.9$ & 86.3 \scriptsize$\pm0.3$ & 77.5 \scriptsize$\pm3.1$ & 66.1 \scriptsize$\pm2.0$ & 11.0 \scriptsize$\pm1.4$ \\
\end{tabular}}
\end{center}
}
\vspace{-0.5cm}
\end{table}

\begin{table}[!t]
\setlength{\tabcolsep}{2.5pt}
\parbox{.47\linewidth}{
\caption{Results on \texttt{DomainNet-MLT}.}
\vspace{-7pt}
\label{table:domainnet-mlt}
\small
\begin{center}
\adjustbox{max width=0.462\textwidth}{
\begin{tabular}{lcccccc}
\toprule[1.5pt]
& \multicolumn{2}{c}{\textbf{Accuracy (by domain)}} & \multicolumn{4}{c}{\textbf{Accuracy (by shot)}} \\
\cmidrule(lr){2-3} \cmidrule(lr){4-7}
\textbf{Algorithm} & \textbf{Average} & \textbf{Worst} & \textbf{Many} & \textbf{Medium} & \textbf{Few} & \textbf{Zero} \\
\midrule
ERM~\cite{vapnik1999overview} & 58.6 \scriptsize$\pm0.2$ & 29.4 \scriptsize$\pm0.3$ & 66.0 \scriptsize$\pm0.1$ & 56.1 \scriptsize$\pm0.1$ & 35.9 \scriptsize$\pm0.5$ & 27.6 \scriptsize$\pm0.3$ \\
IRM~\cite{arjovsky2019irm} & 57.1 \scriptsize$\pm0.1$ & 27.6 \scriptsize$\pm0.1$ & 64.7 \scriptsize$\pm0.1$ & 54.3 \scriptsize$\pm0.3$ & 33.5 \scriptsize$\pm0.3$ & 25.8 \scriptsize$\pm0.3$ \\
GroupDRO~\cite{sagawa2020groupdro} & 53.6 \scriptsize$\pm0.1$ & 25.9 \scriptsize$\pm0.2$ & 61.8 \scriptsize$\pm0.1$ & 49.1 \scriptsize$\pm0.3$ & 30.7 \scriptsize$\pm0.7$ & 22.0 \scriptsize$\pm0.1$ \\
Mixup~\cite{xu2020interdomain_mixup_aaai} & 57.6 \scriptsize$\pm0.1$ & 28.7 \scriptsize$\pm0.0$ & 64.9 \scriptsize$\pm0.2$ & 54.5 \scriptsize$\pm0.1$ & 35.6 \scriptsize$\pm0.2$ & 27.3 \scriptsize$\pm0.3$ \\
MLDG~\cite{li2018mldg} & 58.5 \scriptsize$\pm0.0$ & 28.7 \scriptsize$\pm0.1$ & 66.0 \scriptsize$\pm0.1$ & 55.7 \scriptsize$\pm0.1$ & 35.3 \scriptsize$\pm0.2$ & 26.9 \scriptsize$\pm0.3$ \\
CORAL~\cite{sun2016coral} & 59.4 \scriptsize$\pm0.1$ & 30.1 \scriptsize$\pm0.4$ & 66.4 \scriptsize$\pm0.1$ & 57.1 \scriptsize$\pm0.0$ & 37.7 \scriptsize$\pm0.6$ & 29.9 \scriptsize$\pm0.2$ \\
\end{tabular}}
\end{center}
}
\hfill
\parbox{.55\linewidth}{
\caption{Results over all MDLT benchmarks.}
\vspace{-7pt}
\label{table:avg-all-mdlt}
\small
\begin{center}
\adjustbox{max width=0.536\textwidth}{
\begin{tabular}{lccccc|c}
\toprule[1.5pt]
\textbf{Algorithm} & \textbf{\texttt{VLCS-MLT}} & \textbf{\texttt{PACS-MLT}} & \textbf{\texttt{OfficeHome-MLT}} & \textbf{\texttt{TerraInc-MLT}} & \textbf{\texttt{DomainNet-MLT}} & \textbf{Avg} \\
\midrule
ERM~\cite{vapnik1999overview} & 76.3 \scriptsize$\pm0.4$ & 97.1 \scriptsize$\pm0.1$ & 80.7 \scriptsize$\pm0.0$ & 75.3 \scriptsize$\pm0.3$ & 58.6 \scriptsize$\pm0.2$ & 77.6 \\
IRM~\cite{arjovsky2019irm} & 76.5 \scriptsize$\pm0.2$ & 96.7 \scriptsize$\pm0.2$ & 80.6 \scriptsize$\pm0.4$ & 73.3 \scriptsize$\pm0.7$ & 57.1 \scriptsize$\pm0.1$ & 76.8 \\
GroupDRO~\cite{sagawa2020groupdro} & 76.7 \scriptsize$\pm0.4$ & 97.0 \scriptsize$\pm0.1$ & 80.1 \scriptsize$\pm0.3$ & 72.0 \scriptsize$\pm0.4$ & 53.6 \scriptsize$\pm0.1$ & 75.9 \\
Mixup~\cite{xu2020interdomain_mixup_aaai} & 75.9 \scriptsize$\pm0.1$ & 96.7 \scriptsize$\pm0.2$ & 81.2 \scriptsize$\pm0.2$ & 71.1 \scriptsize$\pm0.7$ & 57.6 \scriptsize$\pm0.1$ & 76.5 \\
MLDG~\cite{li2018mldg} & 76.9 \scriptsize$\pm0.2$ & 96.6 \scriptsize$\pm0.1$ & 80.4 \scriptsize$\pm0.2$ & 76.6 \scriptsize$\pm0.2$ & 58.5 \scriptsize$\pm0.0$ & 77.8 \\
CORAL~\cite{sun2016coral} & 75.9 \scriptsize$\pm0.5$ & 96.6 \scriptsize$\pm0.5$ & 81.9 \scriptsize$\pm0.1$ & 76.4 \scriptsize$\pm0.5$ & 59.4 \scriptsize$\pm0.1$ & 78.0 \\
MMD~\cite{li2018mmd} & 76.3 \scriptsize$\pm0.6$ & 96.9 \scriptsize$\pm0.1$ & 78.4 \scriptsize$\pm0.4$ & 73.3 \scriptsize$\pm0.4$ & 56.7 \scriptsize$\pm0.0$ & 76.3 \\
\end{tabular}}
\end{center}
}
\vspace{-0.5cm}
\end{table}

\paragraph{Benchmark Results on MDLT Datasets.}
The performance of all methods on \texttt{VLCS-MLT}, \texttt{PACS-MLT}, \texttt{OfficeHome-MLT}, \texttt{TerraInc-MLT} and \texttt{DomainNet-MLT} are in Table \ref{table:vlcs-mlt}, \ref{table:pacs-mlt}, \ref{table:officehome-mlt}, \ref{table:terrainc-mlt} and \ref{table:domainnet-mlt}, respectively.
We highlight rows in gray for \boda and its variants, and bolden the best result in each column.
First, as all tables indicate, \boda consistently achieves the best average accuracy across all datasets. It also achieves the best worst-case accuracy most of the time.
Moreover, on certain datasets (e.g., \texttt{OfficeHome-MLT}), MDL methods perform better (e.g., CORAL), while on others (e.g., \texttt{TerraInc-MLT}), imbalanced methods achieve higher gains (e.g., CRT); Nevertheless, regardless of dataset, \boda outperforms all methods, highlighting its effectiveness for the MDLT task.
Finally, compared to ERM, \boda slightly improves the average and many-shot performance, while substantially boosting the performance for the medium-shot, few-shot, and zero-shot pairs.
Table \ref{table:avg-all-mdlt} summarizes the averaged accuracy across all datasets, where \boda brings large overall improvements of $\sim3\%$.

\begin{figure}[tb]
\centering
\includegraphics[width=\textwidth]{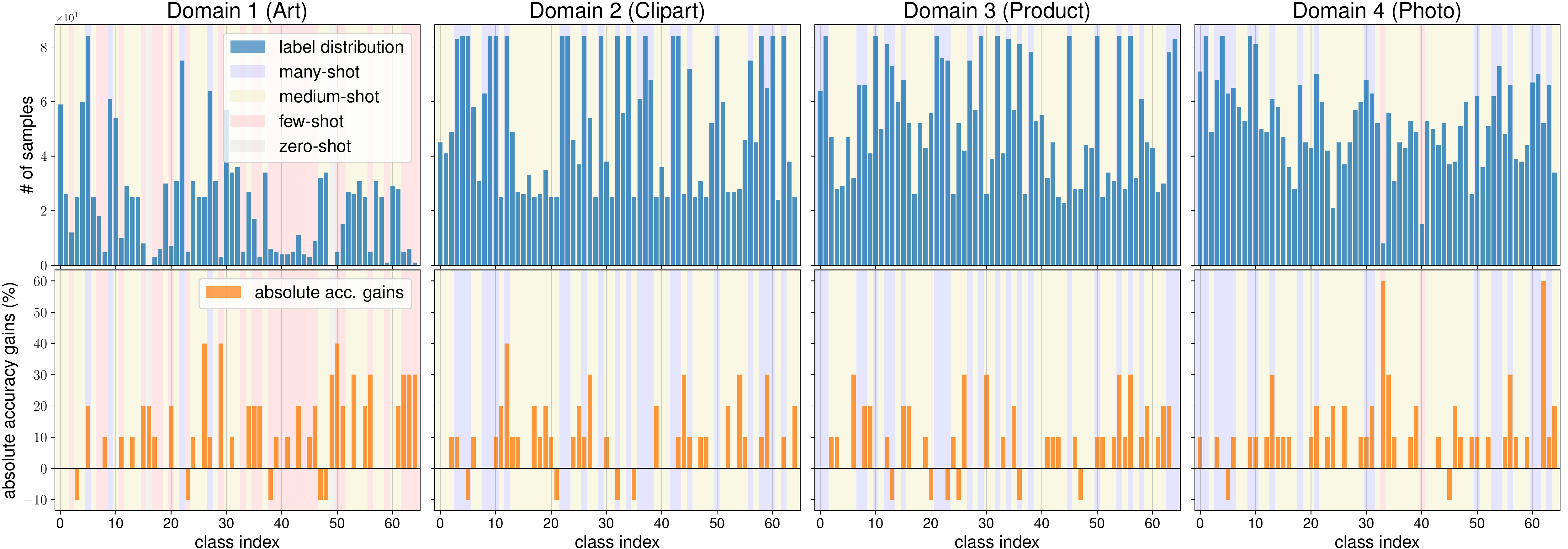}
\vspace{-0.6cm}
\caption{The absolute accuracy improvements of \boda \emph{vs.} ERM over all domain-class pairs on \texttt{OfficeHome-MLT}. \boda establishes large improvements w.r.t. all regions, especially for the few-shot and zero-shot ones. Results for other datasets are in Appendix~\ref{subsec-appendix:abs-gains}.}
\label{fig:gains-officehome}
\end{figure}

\paragraph{A Closer Look at Accuracy Gains.}
We further explore how \boda performs across \emph{all} domain-class pairs. Fig.~\ref{fig:gains-officehome} shows the absolute accuracy gains of \boda over ERM on \texttt{OfficeHome-MLT}, where \boda consistently improves the performance over all domains. The improvements are especially large for domain ``Art'', where most of the classes lie in the \emph{few-shot} region. For certain classes, \boda can improve up to 50\% accuracy, indicating its effectiveness on tackling MDLT.

\paragraph{Ablation Studies on \boda Components (Appendix \ref{subsec-appendix:ablation}).}
We study the effects of (1) adding balanced distance (i.e., \boda \emph{vs.} vanilla \texttt{DA}), and (2) different choices of distance calibration coefficient $\lambda^{d',c'}_{d,c}$ in \boda.
We observe that \boda improves over \texttt{DA} by a large margin ($2.3\%$ on average over all MDLT datasets), highlighting the importance of using \emph{balanced} distance.
Interestingly, as for $\lambda^{d',c'}_{d,c}$, we find that \boda is pretty robust to different choices within a given range, and obtain similar gains ($1.9\%$ to $2.9\%$ over ERM).

\subsection{Understanding the Behavior of \boda on MDLT}

To better understand how the design of \boda contributes to its ability to outperform other algorithms, we go back to the \texttt{Digits-MLT} dataset, but this time we run \boda as opposed to ERM.

\begin{figure}[tb]
\begin{center}
\includegraphics[width=0.98\linewidth]{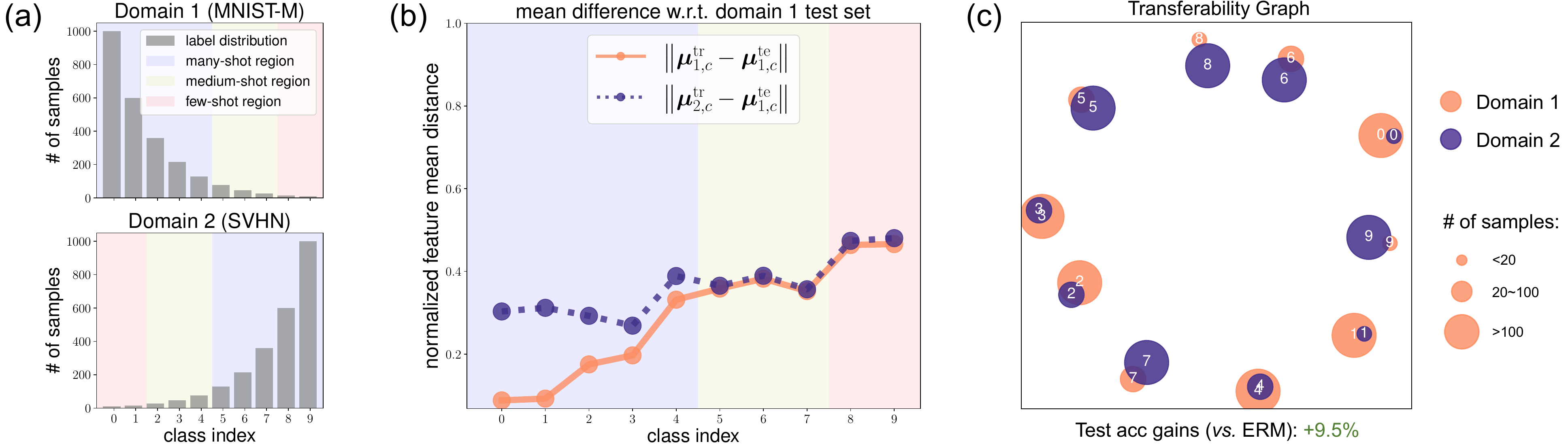}
\end{center}
\vspace{-0.5cm}
\caption{\boda analysis. \textbf{(a)} Label distribution setup. \textbf{(b)} Distance of feature mean between train and test data. \boda enables better learned tail $(d,c)$ with smaller feature discrepancy. \textbf{(c)} \boda learns features that are more aligned across domains even in the presence of divergent labels, and significantly improves upon ERM by $9.5\%$.
}
\label{fig:ours-qualitative}
\end{figure}

\paragraph{Better Learned Representations for Minority Data.}
Similar to Fig.~\ref{fig:motivate-feat-stats}, we plot in Fig.~\ref{fig:ours-qualitative}b the feature mean distance between training and test data for \boda on \texttt{Digits-MLT}. The plot shows that \boda learns better representations with smaller feature discrepancy, especially for minority classes.

\paragraph{Improved Transferability against Severe Imbalance.}
Fig.~\ref{fig:ours-qualitative}c plots the transferability graph induced by \boda. It shows that even in the presence of severe and divergent label imbalance (Fig.~\ref{fig:ours-qualitative}a), \boda still learns transferable features. Further, \boda learns a \emph{balanced} feature space that separates different classes away. The better learned features translate to better accuracy ($9.5\%$ absolute accuracy gains \emph{vs}. ERM in Fig.~\ref{fig:motivate-divergence}c).
We provide more related results in Appendix \ref{subsec-appendix:diverse-skewed-labels} and \ref{subsec-appendix:feat-stats-discrepancy}.

\setlength\intextsep{-8pt}
\begin{wraptable}[4]{r}{0.27\textwidth}
\setlength{\tabcolsep}{5pt}
\caption{\boda bound.}
\vspace{-19pt}
\label{table:exp-bound}
\small
\begin{center}
\adjustbox{max width=0.27\textwidth}{
\begin{tabular}{lc}
\toprule[1.5pt]
            & $\mathcal{L}_\boda$      \\ \midrule
Empirical   & 2.92947 \scriptsize$\pm7.3\texttt{e-3}$ \\
Theoretical & 2.92513 \scriptsize$\pm7.8\texttt{e-3}$ \\
\bottomrule[1.5pt]
\end{tabular}}
\end{center}
\vskip -0.35in
\end{wraptable}

\paragraph{Tightness of the Bound.}
We study whether the \boda bound derived in Theorem \ref{thm:boda-bound} is tight.
We train a ResNet-18 on \texttt{Digits-MLT} for 5,000 steps to ensure convergence. We compute the loss over all samples, and combine the results over 3 random seeds. Table~\ref{table:exp-bound} confirms the bound is empirically tight.

%% file: 8_exp_dg.tex
\begin{table}[t]
\setlength{\tabcolsep}{10pt}
\caption{\boda strengthens performance on Domain Generalization (DG) benchmarks. Full tables including detailed results for each DG dataset are provided in Appendix \ref{sec-appendix:complete-results-dg}.}
\vspace{-6pt}
\label{table:avg-all-dg}
\small
\begin{center}
\adjustbox{max width=0.98\textwidth}{
\begin{tabular}{lccccc|c}
\toprule[1.5pt]
\textbf{Algorithm} & \texttt{VLCS} & \texttt{PACS} & \texttt{OfficeHome} & \texttt{TerraInc} & \texttt{DomainNet} & \textbf{Avg} \\
\midrule
ERM & 77.5 \scriptsize$\pm0.4$ & 85.5 \scriptsize$\pm0.2$ & 66.5 \scriptsize$\pm0.3$ & 46.1 \scriptsize$\pm1.8$ & 40.9 \scriptsize$\pm0.1$ & 63.3 \\
Current SOTA~\cite{sun2016coral} & \textbf{78.8} \scriptsize$\pm0.6$ & 86.2 \scriptsize$\pm0.3$ & 68.7 \scriptsize$\pm0.3$ & 47.6 \scriptsize$\pm1.0$ & 41.5 \scriptsize$\pm0.1$ & 64.5 \\
\grayrow
BoDA$_{r,c}$ & 78.5 \scriptsize$\pm0.3$ & \textbf{86.9} \scriptsize$\pm0.4$ & \textbf{69.3} \scriptsize$\pm0.1$ & \textbf{50.2} \scriptsize$\pm0.4$ & \textbf{42.7} \scriptsize$\pm0.1$ & \textbf{65.5} \\
\midrule \midrule
BoDA$_{r,c}$ \texttt{+} Current SOTA~\cite{sun2016coral} & 79.1 \scriptsize$\pm0.1$ & 87.9 \scriptsize$\pm0.5$ & 69.9 \scriptsize$\pm0.2$ & 50.7 \scriptsize$\pm0.6$ & 43.5 \scriptsize$\pm0.3$ & 66.2 \\
\midrule
BoDA \emph{vs.} ERM & \textcolor{darkgreen}{\texttt{+}\textbf{1.6}} & \textcolor{darkgreen}{\texttt{+}\textbf{2.4}} & \textcolor{darkgreen}{\texttt{+}\textbf{3.4}} & \textcolor{darkgreen}{\texttt{+}\textbf{4.6}} & \textcolor{darkgreen}{\texttt{+}\textbf{2.6}} & \textcolor{darkgreen}{\texttt{+}\textbf{2.9}} \\
\bottomrule[1.5pt]
\end{tabular}
}
\end{center}
\end{table}

Domain Generalization (DG) refers to learning from multiple domains and generalizing to unseen domains. Since naturally the learning domains differ in their label distributions and may even have class imbalance within each domain, we investigate whether tackling cross-domain data imbalance can further strengthen the performance for DG. Note that all datasets we adapted for MDLT are standard benchmarks for DG, which confirms that data imbalance is an intrinsic problem in DG, but has been overlooked by past works. 

We study whether \boda can improve performance for DG.
To test \boda, we follow standard DG evaluation protocol \cite{gulrajani2020domainbed}, and compare to the current SOTA~\cite{sun2016coral}.
Table~\ref{table:avg-all-dg} reveals the following findings:
First, \boda alone can improve upon the current SOTA on four out of the five datasets, and achieves notable average performance gains.
Moreover, combined with the current SOTA, \boda further boosts the result by a notable margin across all datasets, suggesting that label imbalance is orthogonal to existing DG-specific algorithms.
Finally, similar to  MDLT, the gains depend on how severe the imbalance is within a dataset -- e.g., \texttt{TerraInc} exhibits the most severe label imbalance across domains, on which \boda achieves the highest gains.
Detailed results for each DG dataset are provided in Appendix~\ref{sec-appendix:complete-results-dg}.
These intriguing results shed light on how label imbalance can affect out-of-distribution generalization, and highlight the importance of integrating label imbalance for practical DG algorithm design.

%% file: 9_conclusion.tex
We formalize the MDLT task as learning from multi-domain imbalanced data, and generalizing to all domain-class pairs.
We introduce the domain-class transferability graph, and propose \boda, a theoretically grounded loss that tackles MDLT.
Extensive results on five curated real-world MDLT benchmarks verify its superiority.
Furthermore, incorporating \boda into DG algorithms establishes a new SOTA on DG benchmarks.
Our work opens up new avenues for realistic multi-domain learning and generalization in the presence of data imbalance.

%% file: 10_appendix.tex
\section{Theoretical Analysis and Complete Proofs}
\label{sec-appendix:proof}

In this section, we explain the details of Theorem \ref{thm:boda-bound} in the main paper, and also formally describe Theorem \ref{thm:calibrated-boda-bound}. We start with giving additional definitions and providing a useful lemma and its proof, which invoked through the proof of the theorems. We then formally prove the arguments in Theorem \ref{thm:boda-bound} and \ref{thm:calibrated-boda-bound}.

\subsection{Additional Definition, Lemma, and Theorem}
\label{subsec-appendix:additional-defn-lemma-thm}

\begin{tcolorbox}
\begin{definition}[$(\tilde{\alpha},\tilde{\beta},\tilde{\gamma})$ Calibrated Transferability Statistics]
\label{appendix:defn:cal_trans_stats}
The transferability graph can be further described by the following three components:
\begin{align*}
\tilde{\alpha} & = \E_{c} \E_{d} \E_{d'\neq d} \left[ \lambda^{d',c}_{d,c} \cdot \textnormal{trans} \big( (d,c), (d',c) \big) \right], \\
\tilde{\beta} & = \E_{d} \E_{c} \E_{c'\neq c} \left[ \lambda^{d,c'}_{d,c} \cdot \textnormal{trans} \big( (d,c), (d,c') \big) \right], \\
\tilde{\gamma} & = \E_{d} \E_{d'\neq d} \E_{c} \E_{c'\neq c} \left[ \lambda^{d',c'}_{d,c} \cdot \textnormal{trans} \big( (d,c), (d',c') \big) \right],
\end{align*}
where $\lambda^{d',c'}_{d,c} = \left(\frac{N_{d',c'}}{N_{d,c}}\right)^{\nu}$ denotes the distance calibration coefficient.
\end{definition}
\end{tcolorbox}

\vspace{0.1cm}
\begin{tcolorbox}
\begin{lemma}
\label{appendix:lemma:log_func_convex}
Let $\eta, \pi > 0$ and $\varphi:\real\rightarrow\real$, $\varphi(x)=\log(\eta + \pi\exp(x))$. Given a finite sequence $x_1,x_2,\dots,x_M\in\real$, it holds that 
\begin{equation*}
\frac{1}{M}\sum_{i=1}^{M}\varphi(x_i) \geq \varphi\left( \frac{1}{M}\sum_{i=1}^{M} x_i\right).
\end{equation*}
\end{lemma}
\end{tcolorbox}

\begin{proof}
Note that $\varphi$ is smooth and thus twice differentiable for all $x\in \real$. We obtain the second derivative of $\varphi$ as
\begin{equation*}
\varphi''(x)=\frac{\eta\pi\exp(x)}{(\eta+\pi\exp(x))^2} > 0, \quad \forall x\in \real.
\end{equation*}
Therefore, $\varphi$ is convex. Thus, by Jensen's inequality, we obtain that
$\frac{1}{M}\sum_{i=1}^{M}\varphi(x_i) \geq \varphi\left( \frac{1}{M}\sum_{i=1}^{M} x_i\right)$,
which completes the proof.
\end{proof}

\vspace{0.1cm}
\begin{tcolorbox}
\begin{theorem}[$\mathcal{\tilde{L}}_{\boda}$ as an Upper Bound]
\label{thm:calibrated-boda-bound}
Given a multi-domain long-tailed dataset $\mathcal{S}$ with domain label space $\mathcal{D}$ and class label space $\mathcal{C}$ satisfying $|\mathcal{D}| > 1$ and $|\mathcal{C}| > 1$, let $\mathcal{Z}$ be the representation set of all training samples. It holds that
\begin{equation}
\mathcal{\tilde{L}}_{\textnormal{\boda}}(\mathcal{Z},\{\boldsymbol{\mu}\}) \geq N \log \left( |\mathcal{D}|-1 + |\mathcal{D}| (|\mathcal{C}|-1) \exp{\left(
\frac{|\mathcal{C}| |\mathcal{D}|}{N} \cdot \tilde{\alpha} - \frac{|\mathcal{C}|}{N} \cdot \tilde{\beta} - \frac{|\mathcal{C}| (|\mathcal{D}|-1)}{N} \cdot \tilde{\gamma}
\right) } \right),
\label{eqn:calibrated-boda-bound}
\end{equation}
where $(\tilde{\alpha},\tilde{\beta},\tilde{\gamma})$ are the calibrated transferability statistics for $\mathcal{S}$ defined in \defref{appendix:defn:cal_trans_stats}.
\end{theorem}
\end{tcolorbox}

\subsection{Proof of Theorem \ref{thm:boda-bound}}
\label{subsec-appendix:proof-thm-1}

Recall that $\mathcal{M} = \mathcal{D}\times \mathcal{C} := \{ (d,c): d\in \mathcal{D}, c\in\mathcal{C} \}$ is the set of all domain-class pairs. $\mathcal{L}_{\boda}$ is given by
\begin{align*}
\mathcal{L}_{\boda}(\mathcal{Z}, \{\boldsymbol{\mu}\}) & = \sum_{\mathbf{z}_i\in \mathcal{Z}} \frac{-1}{\left|\mathcal{D}\right|-1} \sum_{d\in \mathcal{D}\setminus \{d_i\}} \log \frac{\exp{(- \bdist(\mathbf{z}_i, \boldsymbol{\mu}_{d,c_i}))}}{\sum_{(d',c') \in \mathcal{M} \setminus \{(d_i, c_i)\}} \exp{(- \bdist(\mathbf{z}_i, \boldsymbol{\mu}_{d',c'}))}} \\
& = \sum_{\mathbf{z}_i\in \mathcal{Z}} \ell_{\boda}(\mathbf{z}_i, \{\boldsymbol{\mu}\}),
\end{align*}
where $\ell_{\boda}(\mathbf{z}_i, \{\boldsymbol{\mu}\})$ is the \emph{sample-wise} BoDA loss. We rewrite $\ell_{\boda}$ in the following format
\begin{align}
\ell_{\boda}(\mathbf{z}_i, \{\boldsymbol{\mu}\})
& = - \frac{1}{\left|\mathcal{D}\right|-1} \sum_{d\in \mathcal{D}\setminus \{d_i\}} \log \frac{\exp{(- \bdist(\mathbf{z}_i, \boldsymbol{\mu}_{d,c_i}))}}{\sum_{(d',c') \in \mathcal{M} \setminus \{(d_i, c_i)\}} \exp{(- \bdist(\mathbf{z}_i, \boldsymbol{\mu}_{d',c'}))}} \nonumber \\
& = \log \left( \frac{\sum_{(d',c') \in \mathcal{M} \setminus \{(d_i, c_i)\}} \exp{(- \bdist(\mathbf{z}_i, \boldsymbol{\mu}_{d',c'}))}}{\prod_{d\in \mathcal{D}\setminus \{d_i\}}\exp{(- \bdist(\mathbf{z}_i, \boldsymbol{\mu}_{d,c_i}))}^{\frac{1}{\left|\mathcal{D}\right|-1}}} \right) \nonumber \\
& = \log \left( \frac{\sum_{(d',c') \in \mathcal{M} \setminus \{(d_i, c_i)\}} \exp{(- \bdist(\mathbf{z}_i, \boldsymbol{\mu}_{d',c'}))}}{\exp{\left( -\frac{1}{\left|\mathcal{D}\right|-1} \sum_{d\in \mathcal{D}\setminus \{d_i\}} \bdist(\mathbf{z}_i, \boldsymbol{\mu}_{d,c_i}) \right)}} \right).
\label{appendix:thm1:eqn:transformed-sample-boda-loss}
\end{align}
We will first focus on the term in the numerator of Eqn.~(\ref{appendix:thm1:eqn:transformed-sample-boda-loss}). We can rewrite the sum into two terms
\begin{align*}
& \sum_{(d',c') \in \mathcal{M} \setminus \{(d_i, c_i)\}} \exp{(- \bdist(\mathbf{z}_i, \boldsymbol{\mu}_{d',c'}))} \\
& = \underbrace{\sum_{d'\in\mathcal{D}\setminus\{d_i\}} \sum_{c'\in \{c_i\}} \exp{(- \bdist(\mathbf{z}_i, \boldsymbol{\mu}_{d',c'}))}}_{T_1} + 
\underbrace{\sum_{d'\in\mathcal{D}} \sum_{c'\in \mathcal{C}\setminus \{c_i \}} \exp{(- \bdist(\mathbf{z}_i, \boldsymbol{\mu}_{d',c'}))}}_{T_2}.
\end{align*}
Since the exponential function $\exp(\cdot)$ is convex, we apply Jensen's inequality on both $T_1$ and $T_2$:
\begin{align*}
T_1 \ & \geq\ (|\mathcal{D}|-1) \exp{\left( - \frac{1}{|\mathcal{D}|-1} \sum_{d'\in\mathcal{D}\setminus\{d_i\}} \sum_{c'\in \{c_i\}} \bdist(\mathbf{z}_i, \boldsymbol{\mu}_{d',c'})\right)} \\
& =\ (|\mathcal{D}|-1) \exp{\left( - \frac{1}{|\mathcal{D}|-1} \sum_{d'\in\mathcal{D}\setminus\{d_i\}} \bdist(\mathbf{z}_i, \boldsymbol{\mu}_{d',c_i})\right)}, \\
T_2 \ & \geq\ |\mathcal{D}| (|\mathcal{C}|-1) \exp{\left( - \frac{1}{|\mathcal{D}| (|\mathcal{C}|-1)} \sum_{d'\in\mathcal{D}} \sum_{c'\in \mathcal{C}\setminus \{c_i \}} \bdist(\mathbf{z}_i, \boldsymbol{\mu}_{d',c'}) \right)}.
\end{align*}
Thus, by using $\exp(x)/\exp(y)=\exp(x - y)$ and rearranging terms, we bound $\ell_{\boda}$ by
\begin{equation*}
\resizebox{\textwidth}{!}{$
\begin{aligned}
& \ell_{\boda}(\mathbf{z}_i, \{\boldsymbol{\mu}\}) \\
& \geq \log \Bigg( |\mathcal{D}|-1 + |\mathcal{D}| (|\mathcal{C}|-1) \exp{\Bigg( 
\underbrace{
\frac{1}{|\mathcal{D}|-1} \sum_{d'\in\mathcal{D}\setminus\{d_i\}} \bdist(\mathbf{z}_i, \boldsymbol{\mu}_{d',c_i}) - \frac{1}{|\mathcal{D}| (|\mathcal{C}|-1)} \sum_{d'\in\mathcal{D}} \sum_{c'\in \mathcal{C}\setminus \{c_i \}} \bdist(\mathbf{z}_i, \boldsymbol{\mu}_{d',c'}) 
}_{T(\mathbf{z}_i, \{\boldsymbol{\mu}\})}
\Bigg) } \Bigg).
\end{aligned}
$}
\end{equation*}
Leveraging Lemma~\ref{appendix:lemma:log_func_convex}, by setting $\eta=|\mathcal{D}|-1$, $\pi=|\mathcal{D}| (|\mathcal{C}|-1)$, and $x_i=T(\mathbf{z}_i, \{\boldsymbol{\mu}\})$, we further bound $\mathcal{L}_{\boda}(\mathcal{Z}, \{\boldsymbol{\mu}\})$ by
\begin{align}
\mathcal{L}_{\boda}(\mathcal{Z}, \{\boldsymbol{\mu}\}) 
& = \sum_{\mathbf{z}_i\in \mathcal{Z}} \ell_{\boda}(\mathbf{z}_i, \{\boldsymbol{\mu}\}) \nonumber \\
& \geq \sum_{\mathbf{z}_i\in \mathcal{Z}} \log \left( |\mathcal{D}|-1 + |\mathcal{D}| (|\mathcal{C}|-1) \exp{\left( T(\mathbf{z}_i, \{\boldsymbol{\mu}\})\right) } \right) \nonumber \\
& \geq {|\mathcal{Z}|} \log \left( |\mathcal{D}|-1 + |\mathcal{D}| (|\mathcal{C}|-1) \exp{\left( \frac{1}{|\mathcal{Z}|} \sum_{\mathbf{z}_i\in \mathcal{Z}} T(\mathbf{z}_i, \{\boldsymbol{\mu}\})\right) } \right). \label{appendix:thm1:eqn:boda-bound-1st}
\end{align}
Note that the argument of the $\exp(\cdot)$ in Eqn.~(\ref{appendix:thm1:eqn:boda-bound-1st}) can be expanded and further rearranged as
\begin{align}
\frac{1}{|\mathcal{Z}|} \sum_{\mathbf{z}_i\in \mathcal{Z}} T(\mathbf{z}_i, \{\boldsymbol{\mu}\}) =\ & \frac{1}{|\mathcal{Z}|} \sum_{\mathbf{z}_i\in \mathcal{Z}}
\frac{1}{|\mathcal{D}|-1} \sum_{d'\in\mathcal{D}\setminus\{d_i\}} \bdist(\mathbf{z}_i, \boldsymbol{\mu}_{d',c_i}) - \nonumber \\
& \frac{1}{|\mathcal{Z}|} \sum_{\mathbf{z}_i\in \mathcal{Z}} \frac{1}{|\mathcal{D}| (|\mathcal{C}|-1)}  \sum_{d'\in\mathcal{D}} \sum_{c'\in \mathcal{C}\setminus \{c_i \}} \bdist(\mathbf{z}_i, \boldsymbol{\mu}_{d',c'}) \nonumber \\
=\ & \underbrace{\frac{1}{|\mathcal{Z}|} \frac{1}{|\mathcal{D}|-1} \sum_{\mathbf{z}_i\in \mathcal{Z}}
\sum_{d'\in\mathcal{D}\setminus\{d_i\}} \bdist(\mathbf{z}_i, \boldsymbol{\mu}_{d',c_i})}_{T_\alpha} 
- \nonumber \\
& \underbrace{\frac{1}{|\mathcal{Z}|} \frac{1}{|\mathcal{D}| (|\mathcal{C}|-1)} \sum_{\mathbf{z}_i\in \mathcal{Z}}  \sum_{c'\in \mathcal{C}\setminus \{c_i \}} \bdist(\mathbf{z}_i, \boldsymbol{\mu}_{d_i,c'})}_{T_\beta} - \nonumber \\
& \underbrace{\frac{1}{|\mathcal{Z}|} \frac{1}{|\mathcal{D}| (|\mathcal{C}|-1)} \sum_{\mathbf{z}_i\in \mathcal{Z}}  \sum_{d'\in\mathcal{D}\setminus\{d_i\}} \sum_{c'\in \mathcal{C}\setminus \{c_i \}} \bdist(\mathbf{z}_i, \boldsymbol{\mu}_{d',c'})}_{T_\gamma}. \label{appendix:thm1:eqn:divide-a-b-g-terms}
\end{align}
Recall that each $\mathbf{z}_i\in \mathcal{Z}$ belongs to a domain-class pair $(d_i, c_i)$, and $\mathcal{Z}_{d,c}$ denotes the representation set of $\mathcal{S}_{d,c}$ with size $N_{d,c}$. For simplicity, we remove the subscript $i$ in the following derivation. We can further rewrite $T_\alpha, T_\beta, T_\gamma$ as
\begin{align}
T_\alpha &\ =\ \frac{1}{|\mathcal{Z}|} \frac{1}{|\mathcal{D}|-1} \sum_{c\in\mathcal{C}} \sum_{d\in\mathcal{D}} \sum_{d'\in\mathcal{D}\setminus\{d\}} \sum_{\mathbf{z}\in \mathcal{Z}_{d,c}} \bdist(\mathbf{z}, \boldsymbol{\mu}_{d',c}) \nonumber \\
&\ =\ \frac{1}{|\mathcal{Z}|} \frac{1}{|\mathcal{D}|-1} |\mathcal{C}| |\mathcal{D}| (|\mathcal{D}| - 1) \E_{c} \E_{d} \E_{d'\neq d} \E_{\mathbf{z}\in\mathcal{Z}_{d,c}} \big[ \underbrace{N_{d,c}\cdot \bdist(\mathbf{z}, \boldsymbol{\mu}_{d',c})}_{\dist(\mathbf{z}, \boldsymbol{\mu}_{d',c})} \big] \nonumber \\
&\ =\ \frac{|\mathcal{C}| |\mathcal{D}|}{|\mathcal{Z}|} \underbrace{\E_{c} \E_{d} \E_{d'\neq d} \E_{\mathbf{z}\in\mathcal{Z}_{d,c}} \big[ \dist(\mathbf{z}, \boldsymbol{\mu}_{d',c}) \big]}_{\alpha}, \label{appendix:thm1:eqn:term-alpha}
\\
T_\beta &\ =\ \frac{1}{|\mathcal{Z}|} \frac{1}{|\mathcal{D}| (|\mathcal{C}|-1)} \sum_{c\in\mathcal{C}} \sum_{d\in\mathcal{D}} \sum_{c'\in \mathcal{C}\setminus \{c \}} \sum_{\mathbf{z}\in \mathcal{Z}_{d,c}} \bdist(\mathbf{z}, \boldsymbol{\mu}_{d,c'}) \nonumber \\
&\ =\ \frac{1}{|\mathcal{Z}|} \frac{1}{|\mathcal{D}| (|\mathcal{C}|-1)} |\mathcal{C}| |\mathcal{D}| (|\mathcal{C}| - 1) \E_{d} \E_{c} \E_{c'\neq c} \E_{\mathbf{z}\in\mathcal{Z}_{d,c}} \big[ \underbrace{N_{d,c}\cdot \bdist(\mathbf{z}, \boldsymbol{\mu}_{d,c'})}_{\dist(\mathbf{z}, \boldsymbol{\mu}_{d,c'})} \big] \nonumber \\
&\ =\ \frac{|\mathcal{C}|}{|\mathcal{Z}|} \underbrace{\E_{d} \E_{c} \E_{c'\neq c} \E_{\mathbf{z}\in\mathcal{Z}_{d,c}} \big[ \dist(\mathbf{z}, \boldsymbol{\mu}_{d,c'}) \big]}_{\beta}, \label{appendix:thm1:eqn:term-beta}
\\
T_\gamma &\ =\ \frac{1}{|\mathcal{Z}|} \frac{1}{|\mathcal{D}| (|\mathcal{C}|-1)} \sum_{c\in\mathcal{C}} \sum_{d\in\mathcal{D}} \sum_{d'\in\mathcal{D}\setminus\{d\}} \sum_{c'\in \mathcal{C}\setminus \{c \}} \sum_{\mathbf{z}\in \mathcal{Z}_{d,c}} \bdist(\mathbf{z}, \boldsymbol{\mu}_{d',c'}) \nonumber \\
&\ =\ \frac{1}{|\mathcal{Z}|} \frac{|\mathcal{C}| |\mathcal{D}| (|\mathcal{D}| - 1) (|\mathcal{C}| - 1)}{|\mathcal{D}| (|\mathcal{C}|-1)} \E_{d} \E_{d'\neq d} \E_{c} \E_{c'\neq c} \E_{\mathbf{z}\in\mathcal{Z}_{d,c}} \big[ \underbrace{N_{d,c}\cdot \bdist(\mathbf{z}, \boldsymbol{\mu}_{d',c'})}_{\dist(\mathbf{z}, \boldsymbol{\mu}_{d',c'})} \big] \nonumber \\
&\ =\ \frac{|\mathcal{C}| (|\mathcal{D}|-1)}{|\mathcal{Z}|} \underbrace{\E_{d} \E_{d'\neq d} \E_{c} \E_{c'\neq c} \E_{\mathbf{z}\in\mathcal{Z}_{d,c}} \big[ \dist(\mathbf{z}, \boldsymbol{\mu}_{d',c'}) \big]}_{\gamma}, \label{appendix:thm1:eqn:term-gamma}
\end{align}
where $(\alpha,\beta,\gamma)$ are the transferability statistics for $\mathcal{S}$ as in Definition~\ref{defn:trans_stats}. Finally, replace $|\mathcal{Z}|=N$ and combine Eqn.~(\ref{appendix:thm1:eqn:boda-bound-1st}), (\ref{appendix:thm1:eqn:divide-a-b-g-terms}), (\ref{appendix:thm1:eqn:term-alpha}), (\ref{appendix:thm1:eqn:term-beta}), and (\ref{appendix:thm1:eqn:term-gamma}), we have
\begin{equation*}
\mathcal{L}_{\boda}(\mathcal{Z}, \{\boldsymbol{\mu}\}) \geq N \log \left( |\mathcal{D}|-1 + |\mathcal{D}| (|\mathcal{C}|-1) \exp{\left(
\frac{|\mathcal{C}| |\mathcal{D}|}{N} \cdot \alpha - \frac{|\mathcal{C}|}{N} \cdot \beta - \frac{|\mathcal{C}| (|\mathcal{D}|-1)}{N} \cdot \gamma
\right) } \right).
\end{equation*}
This completes the proof.

\subsection{Proof of Theorem \ref{thm:calibrated-boda-bound}}
\label{subsec-appendix:proof-thm-2}

We first define a notion of \emph{calibrated distance} $\cdist$. Let $\mathbf{z}\in\mathcal{Z}_{d,c}$, we have
\begin{equation*}
\cdist(\mathbf{z}, \boldsymbol{\mu}_{d',c'}) \triangleq \lambda^{d',c'}_{d,c} \cdot \bdist(\mathbf{z}, \boldsymbol{\mu}_{d',c'}) = \left(\frac{N_{d',c'}}{N_{d,c}}\right)^{\nu} \cdot \bdist(\mathbf{z}, \boldsymbol{\mu}_{d',c'}).
\end{equation*}
From Theorem~\ref{thm:boda-bound}, by substituting $\bdist$ with $\cdist$, it holds that
\begin{equation}
\begin{aligned}
\mathcal{\tilde{L}}_{\textnormal{\boda}}(\mathcal{Z},\{\boldsymbol{\mu}\}) 
& = \mathcal{L}_{\textnormal{\boda}}(\mathcal{Z},\{\boldsymbol{\mu}\}) \Big{\vert}_{\bdist \to \cdist} \\
& \geq N \log \left( |\mathcal{D}|-1 + |\mathcal{D}| (|\mathcal{C}|-1) \exp{\left(
T'_\alpha - T'_\beta - T'_\gamma
\right) } \right),
\label{appendix:thm2:eqn:boda-bound-1st}
\end{aligned}
\end{equation}
where $T'_\alpha$, $T'_\beta$, and $T'_\gamma$ can be expressed as
\begin{align}
T'_\alpha
&\ =\ \frac{|\mathcal{C}| |\mathcal{D}|}{N} \E_{c} \E_{d} \E_{d'\neq d} \E_{\mathbf{z}\in\mathcal{Z}_{d,c}} \big[ N_{d,c}\cdot \cdist(\mathbf{z}, \boldsymbol{\mu}_{d',c}) \big] \nonumber \\
&\ =\ \frac{|\mathcal{C}| |\mathcal{D}|}{N} \E_{c} \E_{d} \E_{d'\neq d} \E_{\mathbf{z}\in\mathcal{Z}_{d,c}} \big[ \lambda^{d',c}_{d,c} \cdot \underbrace{N_{d,c}\cdot \bdist(\mathbf{z}, \boldsymbol{\mu}_{d',c})}_{\dist(\mathbf{z}, \boldsymbol{\mu}_{d',c})} \big] \nonumber \\
&\ =\ \frac{|\mathcal{C}| |\mathcal{D}|}{N} \underbrace{\E_{c} \E_{d} \E_{d'\neq d} \left[ \lambda^{d',c}_{d,c} \cdot \E_{\mathbf{z}\in\mathcal{Z}_{d,c}} \big[ \dist(\mathbf{z}, \boldsymbol{\mu}_{d',c}) \big] \right]}_{\tilde{\alpha}}, \label{appendix:thm2:eqn:term-alpha}
\\
T'_\beta
&\ =\ \frac{|\mathcal{C}|}{N} \E_{d} \E_{c} \E_{c'\neq c} \E_{\mathbf{z}\in\mathcal{Z}_{d,c}} \big[ N_{d,c}\cdot \cdist(\mathbf{z}, \boldsymbol{\mu}_{d,c'}) \big] \nonumber \\
&\ =\ \frac{|\mathcal{C}|}{N} \E_{d} \E_{c} \E_{c'\neq c} \E_{\mathbf{z}\in\mathcal{Z}_{d,c}} \big[ \lambda^{d,c'}_{d,c} \cdot \underbrace{N_{d,c}\cdot \bdist(\mathbf{z}, \boldsymbol{\mu}_{d,c'})}_{\dist(\mathbf{z}, \boldsymbol{\mu}_{d,c'})} \big] \nonumber \\
&\ =\ \frac{|\mathcal{C}|}{N} \underbrace{\E_{d} \E_{c} \E_{c'\neq c} \left[ \lambda^{d,c'}_{d,c} \cdot \E_{\mathbf{z}\in\mathcal{Z}_{d,c}} \big[ \dist(\mathbf{z}, \boldsymbol{\mu}_{d,c'}) \big] \right]}_{\tilde{\beta}}, \label{appendix:thm2:eqn:term-beta}
\\
T'_\gamma
&\ =\ \frac{|\mathcal{C}| (|\mathcal{D}|-1)}{N} \E_{d} \E_{d'\neq d} \E_{c} \E_{c'\neq c} \E_{\mathbf{z}\in\mathcal{Z}_{d,c}} \big[ N_{d,c}\cdot \cdist(\mathbf{z}, \boldsymbol{\mu}_{d',c'}) \big] \nonumber \\
&\ =\ \frac{|\mathcal{C}| (|\mathcal{D}|-1)}{N} \E_{d} \E_{d'\neq d} \E_{c} \E_{c'\neq c} \E_{\mathbf{z}\in\mathcal{Z}_{d,c}} \big[ \lambda^{d',c'}_{d,c} \cdot \underbrace{N_{d,c}\cdot \bdist(\mathbf{z}, \boldsymbol{\mu}_{d',c'})}_{\dist(\mathbf{z}, \boldsymbol{\mu}_{d',c'})} \big] \nonumber \\
&\ =\ \frac{|\mathcal{C}| (|\mathcal{D}|-1)}{N} \underbrace{\E_{d} \E_{d'\neq d} \E_{c} \E_{c'\neq c} \left[ \lambda^{d',c'}_{d,c} \cdot \E_{\mathbf{z}\in\mathcal{Z}_{d,c}} \big[ \dist(\mathbf{z}, \boldsymbol{\mu}_{d',c'}) \big] \right]}_{\tilde{\gamma}}, \label{appendix:thm2:eqn:term-gamma}
\end{align}
where $(\tilde{\alpha},\tilde{\beta},\tilde{\gamma})$ are formally defined in Definition~\ref{appendix:defn:cal_trans_stats}. Combine Eqn.~(\ref{appendix:thm2:eqn:boda-bound-1st}), (\ref{appendix:thm2:eqn:term-alpha}), (\ref{appendix:thm2:eqn:term-beta}), and (\ref{appendix:thm2:eqn:term-gamma}), we have
\begin{equation*}
\mathcal{\tilde{L}}_{\textnormal{\boda}}(\mathcal{Z},\{\boldsymbol{\mu}\}) \geq N \log \left( |\mathcal{D}|-1 + |\mathcal{D}| (|\mathcal{C}|-1) \exp{\left(
\frac{|\mathcal{C}| |\mathcal{D}|}{N} \cdot \tilde{\alpha} - \frac{|\mathcal{C}|}{N} \cdot \tilde{\beta} - \frac{|\mathcal{C}| (|\mathcal{D}|-1)}{N} \cdot \tilde{\gamma}
\right) } \right),
\end{equation*}
which completes the proof.

\section{Additional Discussions, Properties, and Interpretations}
\label{sec-appendix:discuss-property}

\subsection{Unified Interpretation for Single- and Multi-Domain Imbalance}
\label{subsec-appendix:consistent-view-imbalance}

In the main paper we show that, in the multi-domain setting, label imbalance implicitly brings \emph{label divergence} across domains, which brings additional challenges and potentially harms MDLT performance. Here we provide a unified viewpoint from the \emph{label divergence} perspective to explain single- and multi-domain data imbalance.

To elaborate, in single domain imbalanced learning, we essentially cope with the divergence between the imbalanced training label distribution and the uniform test label distribution:
\begin{align*}
\textnormal{div}(p(y)\ \| \ \mathcal{U}),
\end{align*}
where $\textnormal{div}(\cdot \| \cdot)$ indicates certain divergence measure.
In contrast, when extending to the multi-domain scenario, given $|\mathcal{D}|$ domains with (different) imbalanced label distributions, the target divergence becomes
\begin{align*}
\underbrace{\sum_{d} \textnormal{div}\big(p_d(y) \ \| \ \mathcal{U}\big)}_{\text{imbalanced training}} + \
\textnormal{const}\cdot \underbrace{\sum_{d\neq d'} \textnormal{div}\big(p_d(y)\ \| \ p_{d'}(y)\big)}_{\text{divergence across domains}},
\end{align*}
where one not only needs to tackle the imbalanced training data for each domain $d\in \mathcal{D}$ in order to generalize to the balanced test set, but also takes into consideration the \emph{label divergence} across domains.

Such interpretation echoes our \boda objective: We design the \texttt{DA} loss for cross-domain distribution alignment to tackle the latter term, and further adapt it to \boda via balanced distance to address the former term.

\subsection{A Probabilistic Perspective of $\mathcal{L}_{\texttt{DA}}$ Derivation}
\label{subsec-appendix:prob-derivation-da}

Recall $\mathcal{M} = \mathcal{D}\times \mathcal{C}$ the set of all $(d,c)$ pairs.
Let $(\mathbf{x}_i,c_i,d_i)$ denote a sample with feature $\mathbf{z}_i$.
Following the metric learning setting \cite{goldberger2004neighbourhood}, we model the likelihood of $\boldsymbol{\mu}_{d,c}$ given $\mathbf{z}_i$ to decay exponentially with respect to their distance in the representation space. Such modeling can be viewed as performing a random walk with transition probability inversely related to distance \cite{globerson2004euclidean}.
For domain-class pairs that share the same class label but different domain labels with $\mathbf{x}_i$ (i.e., $(d,c_i), d\neq d_i$), the normalized likelihood of $\boldsymbol{\mu}_{d,c_i}$ given $\mathbf{z}_i$ can be written as
\begin{equation*}
\mathbb{P}((d,c_i) | \mathbf{z}_i) = \frac{\exp{(- \dist(\mathbf{z}_i, \boldsymbol{\mu}_{d,c_i}))}}{
\sum_{(d',c') \in \mathcal{M} \setminus \{(d_i, c_i)\}} \exp{(- \dist(\mathbf{z}_i, \boldsymbol{\mu}_{d',c'}))}},
\end{equation*}
where the denominator is a sum over all domain-class pairs except $(d_i,c_i)$.
As motivated, we want to concentrate all $\mathbf{z}_i$ from the same class across different domains (i.e., smaller $\alpha$), while separating $\mathbf{z}_i$ from different classes within and across domains (i.e., larger $\beta,\gamma$). Therefore, the positive domain-class pairs with $\mathbf{x}_i$ are those share the same class labels but different domain labels. As a result, we define the per-sample loss as the average negative log-likelihood over all positive domain-class pairs:
\begin{equation*}
\ell_{\texttt{DA}}(\mathbf{z}_i,\{\boldsymbol{\mu}\}) = - \frac{1}{|\mathcal{D}|-1} \sum_{d\in \mathcal{D}\setminus \{d_i\}} \log \frac{\exp{(- \dist(\mathbf{z}_i, \boldsymbol{\mu}_{d,c_i}))}}{\sum_{(d',c') \in \mathcal{M} \setminus \{(d_i, c_i)\}} \exp{(- \dist(\mathbf{z}_i, \boldsymbol{\mu}_{d',c'}))}}.
\end{equation*}
Given a set of all training samples with representation set as $\mathcal{Z}$, the total loss can then be derived as
\begin{equation*}
\mathcal{L}_{\texttt{DA}}(\mathcal{Z}, \{\boldsymbol{\mu}\})
= \sum_{\mathbf{z}_i\in \mathcal{Z}} \frac{-1}{|\mathcal{D}|-1} \sum_{d\in \mathcal{D}\setminus \{d_i\}} \log \frac{\exp{(- \dist(\mathbf{z}_i, \boldsymbol{\mu}_{d,c_i}))}}{\sum_{(d',c') \in \mathcal{M} \setminus \{(d_i, c_i)\}} \exp{(- \dist(\mathbf{z}_i, \boldsymbol{\mu}_{d',c'}))}}.
\end{equation*}

\subsection{Intrinsic Hardness-Aware Property of \boda}
\label{subsec-appendix:gradient-analysis}

Below, we demonstrate an additional property of \boda: the intrinsic \emph{hardness-aware} property.
Specifically, we analyze the gradients of \boda loss with respect to positive $(d,c)$ pairs and different negative $(d,c)$ pairs. We observe that the gradient contributions from \emph{hard} positives/negatives are larger than that from the \emph{easy} ones, indicating that \boda automatically concentrates on the \emph{hard} $(d,c)$ pairs, where penalties are given according to their hardness.

Recall that the sample-wise calibrated \boda loss $\tilde{\ell}_{\boda}$ can be written as
\begin{align}
& \tilde{\ell}_{\boda}(\mathbf{z}_i, \{\boldsymbol{\mu}\}) \nonumber \\
& = - \frac{1}{\left|\mathcal{D}\right|-1} \sum_{d\in \mathcal{D}\setminus \{d_i\}} \log \frac{\exp{\left(- \lambda^{d,c_i}_{d_i,c_i} \bdist(\mathbf{z}_i, \boldsymbol{\mu}_{d,c_i})\right)}}{\sum_{(d',c') \in \mathcal{M} \setminus \{(d_i, c_i)\}} \exp{\left(- \lambda^{d',c'}_{d_i,c_i} \bdist(\mathbf{z}_i, \boldsymbol{\mu}_{d',c'})\right)}} \nonumber \\
& = - \frac{1}{\left|\mathcal{D}\right|-1} \sum_{d\in \mathcal{D}\setminus \{d_i\}} \log \frac{\exp{\left(- \frac{\lambda^{d,c_i}_{d_i,c_i}}{N_{d_i,c_i}} \dist(\mathbf{z}_i, \boldsymbol{\mu}_{d,c_i})\right)}}{\sum_{(d',c') \in \mathcal{M} \setminus \{(d_i, c_i)\}} \exp{\left(- \frac{\lambda^{d',c'}_{d_i,c_i}}{N_{d_i,c_i}} \dist(\mathbf{z}_i, \boldsymbol{\mu}_{d',c'})\right)}},
\label{appendix:eqn:grad-analysis-init-boda}
\end{align}
where $\mathbf{z}_i\in \mathcal{Z}_{d_i,c_i}$. For convenience, we further define the probability of $\mathbf{z}_i$ being recognized as belonging to $\boldsymbol{\mu}_{d,c}$ as
\begin{equation*}
P^{i}_{d,c} \triangleq \frac{\exp{\left(- \frac{\lambda^{d,c}_{d_i,c_i}}{N_{d_i,c_i}} \dist(\mathbf{z}_i, \boldsymbol{\mu}_{d,c})\right)}}{\sum_{(d',c') \in \mathcal{M} \setminus \{(d_i, c_i)\}} \exp{\left(- \frac{\lambda^{d',c'}_{d_i,c_i}}{N_{d_i,c_i}} \dist(\mathbf{z}_i, \boldsymbol{\mu}_{d',c'})\right)}}, \quad (d,c) \in \mathcal{M} \setminus \{(d_i, c_i)\}.
\end{equation*}
Note that the essential goal of Eqn.~(\ref{appendix:eqn:grad-analysis-init-boda}) is to align (minimize) \emph{positive} distances $\dist(\mathbf{z}_i, \boldsymbol{\mu}_{d,c_i})$ and to separate (maximize) \emph{negative} distances $\dist(\mathbf{z}_i, \boldsymbol{\mu}_{d',c'})$. Therefore, we analyze the gradients with respect to positive distance and different negative distances to explore the properties of $\tilde{\ell}_{\boda}$. Specifically, we have

\begin{equation*}
\resizebox{\textwidth}{!}{$
\begin{aligned}
& \frac{\partial \tilde{\ell}_{\boda}(\mathbf{z}_i, \{\boldsymbol{\mu}\})}{\partial \dist(\mathbf{z}_i, \boldsymbol{\mu}_{d,c_i})} \nonumber \\
& = \frac{-1}{\left|\mathcal{D}\right|-1} \sum_{d\in \mathcal{D}\setminus \{d_i\}} \frac{\partial}{\partial \dist(\mathbf{z}_i, \boldsymbol{\mu}_{d,c_i})} \left\{
- \frac{\lambda^{d,c_i}_{d_i,c_i}}{N_{d_i,c_i}} \dist(\mathbf{z}_i, \boldsymbol{\mu}_{d,c_i}) - \log \sum_{(d',c') \in \mathcal{M} \setminus \{(d_i, c_i)\}} \exp{\left(- \frac{\lambda^{d',c'}_{d_i,c_i}}{N_{d_i,c_i}} \dist(\mathbf{z}_i, \boldsymbol{\mu}_{d',c'})\right)} \right\} \nonumber \\
& = \frac{1}{\left|\mathcal{D}\right|-1} \sum_{d\in \mathcal{D}\setminus \{d_i\}} \frac{\lambda^{d,c_i}_{d_i,c_i}}{N_{d_i,c_i}} \left( 1 - \frac{\exp{\left(- \frac{\lambda^{d,c_i}_{d_i,c_i}}{N_{d_i,c_i}} \dist(\mathbf{z}_i, \boldsymbol{\mu}_{d,c_i})\right)}}{\sum_{(d',c') \in \mathcal{M} \setminus \{(d_i, c_i)\}} \exp{\left(- \frac{\lambda^{d',c'}_{d_i,c_i}}{N_{d_i,c_i}} \dist(\mathbf{z}_i, \boldsymbol{\mu}_{d',c'})\right)}} \right) \nonumber \\
& = \frac{1}{\left|\mathcal{D}\right|-1} \sum_{d\in \mathcal{D}\setminus \{d_i\}} \frac{N^\nu_{d,c_i}}{N^{(1+\nu)}_{d_i,c_i}} \left( 1 - P^{i}_{d,c_i} \right) \nonumber \\
& \propto \sum_{d\in \mathcal{D}\setminus \{d_i\}} N^\nu_{d,c_i} \left( 1 - P^{i}_{d,c_i} \right),
\end{aligned}
$}
\end{equation*}
\begin{equation*}
\resizebox{\textwidth}{!}{$
\begin{aligned}
& \frac{\partial \tilde{\ell}_{\boda}(\mathbf{z}_i, \{\boldsymbol{\mu}\})}{\partial \dist(\mathbf{z}_i, \boldsymbol{\mu}_{d',c'})} \nonumber \\
& = \frac{-1}{\left|\mathcal{D}\right|-1} \sum_{d\in \mathcal{D}\setminus \{d_i\}} \frac{\partial}{\partial \dist(\mathbf{z}_i, \boldsymbol{\mu}_{d',c'})} \left\{
- \frac{\lambda^{d,c_i}_{d_i,c_i}}{N_{d_i,c_i}} \dist(\mathbf{z}_i, \boldsymbol{\mu}_{d,c_i}) - \log \sum_{(d',c') \in \mathcal{M} \setminus \{(d_i, c_i)\}} \exp{\left(- \frac{\lambda^{d',c'}_{d_i,c_i}}{N_{d_i,c_i}} \dist(\mathbf{z}_i, \boldsymbol{\mu}_{d',c'})\right)} \right\} \nonumber \\
& = - \frac{1}{\left|\mathcal{D}\right|-1} \sum_{d\in \mathcal{D}\setminus \{d_i\}} \frac{\lambda^{d',c'}_{d_i,c_i}}{N_{d_i,c_i}} \frac{\exp{\left(- \frac{\lambda^{d',c'}_{d_i,c_i}}{N_{d_i,c_i}} \dist(\mathbf{z}_i, \boldsymbol{\mu}_{d,c_i})\right)}}{\sum_{(d',c') \in \mathcal{M} \setminus \{(d_i, c_i)\}} \exp{\left(- \frac{\lambda^{d',c'}_{d_i,c_i}}{N_{d_i,c_i}} \dist(\mathbf{z}_i, \boldsymbol{\mu}_{d',c'})\right)}} \nonumber \\
& = - \frac{1}{\left|\mathcal{D}\right|-1} \sum_{d\in \mathcal{D}\setminus \{d_i\}} \frac{N^\nu_{d',c'}}{N^{(1+\nu)}_{d_i,c_i}}  P^{i}_{d',c'} \nonumber \\
& \propto - N^\nu_{d',c'} P^{i}_{d',c'}.
\end{aligned}
$}
\end{equation*}
Combine the above results, we have
\begin{align}
\textrm{positive:}\quad & \frac{\partial \tilde{\ell}_{\boda}(\mathbf{z}_i, \{\boldsymbol{\mu}\})}{\partial \dist(\mathbf{z}_i, \boldsymbol{\mu}_{d,c_i})} \propto \sum_{d\in \mathcal{D}\setminus \{d_i\}} N^\nu_{d,c_i} \left( 1 - P^{i}_{d,c_i} \right), \label{appendix:eqn:grad-pos} \\
\textrm{negative:}\quad & \frac{\partial \tilde{\ell}_{\boda}(\mathbf{z}_i, \{\boldsymbol{\mu}\})}{\partial \dist(\mathbf{z}_i, \boldsymbol{\mu}_{d',c'})} \propto - N^\nu_{d',c'} P^{i}_{d',c'}. \label{appendix:eqn:grad-neg}
\end{align}

\paragraph{Interpretation.}
Eqn.~(\ref{appendix:eqn:grad-pos}) and (\ref{appendix:eqn:grad-neg}) illustrate several interesting and important properties of \boda:
\begin{Enumerate}
    \item \emph{Intrinsic hard positive and negative mining}.
    For positive pairs, we observe that the gradient magnitudes are proportional to $(1 - P^{i}_{d,c_i})$, where for an easy $(d,c_i)$ pair, $P^{i}_{d,c_i}\approx 1$ and $(1 - P^{i}_{d,c_i})\approx 0$, and for a hard $(d,c_i)$ pair, $P^{i}_{d,c_i}\approx 0$ and $(1 - P^{i}_{d,c_i})\approx 1$, indicating that the gradient contributions from \emph{hard} positives are larger than \emph{easy} ones. Similarly, for negative pairs, the gradient magnitudes are proportional to $P^{i}_{d',c'}$, where an easy $(d',c')$ pair has $P^{i}_{d',c'}\approx 0$ and a hard $(d,c_i)$ pair induces $P^{i}_{d',c'}\approx 1$, showing that the gradient contribution is large for hard negatives and small for easy negatives. Therefore, \boda is a hardness-aware loss with intrinsic hard positive/negative mining property.
    \vspace{0.5em}
    \item \emph{Scaling gradients according to the number of samples of each $(d,c)$}.
    Furthermore, as we have shown in Fig.~\ref{fig:motivate-feat-stats}, when data are imbalanced across different $(d,c)$ pairs, minority pairs with smaller number of samples would induce worse $\boldsymbol{\mu}_{d,c}$ estimates. We further observe that the gradients for both positive and negative pairs are proportional to their number of samples (i.e., $N^\nu_{d,c_i}$ and $N^\nu_{d',c'}$). This suggests that \boda automatically adjusts the gradient scale for each $(d,c)$ according to how accurate the estimation of $\boldsymbol{\mu}_{d,c}$ is. The appealing property highlights that \boda also implicitly calibrates the gradient scale, emphasizing gradients from majority pairs (which are more reliable) while suppressing gradients from minority pairs (which are less reliable). Such behavior is essential for better statistics transfer as we demonstrated in the main paper.
\end{Enumerate}

\section{Pseudo Code for \boda}
\label{sec-appendix:pseudo-code}

We provide the pseudo code of \boda in Algorithm~\ref{alg:boda}.

\begin{algorithm}[!th]
   \caption{Balanced Domain-Class Distribution Alignment (\boda)}
   \label{alg:boda}
\begin{algorithmic}
   \STATE {\bfseries Input:} Training set $\mathcal{D}=\{ ( \mathbf{x}_i, c_i, d_i )\}_{i=1}^{N}$, all domain-class pairs $\mathcal{M}=\{(d,c)\}$, encoder $f$, classifier $g$, total training epochs $E$, calibration parameter $\nu$, loss weight $\omega$, momentum $\alpha$
   \FORALL{$(d,c)\in \mathcal{M}$}
   \STATE Initialize the feature statistics $\{{\boldsymbol{\mu}}_{d,c}^{(0)}, \boldsymbol{{\Sigma}}_{d,c}^{(0)}\}$
   \ENDFOR
   \FOR{$e = 0$ \textbf{to} $E$}
   \REPEAT
   \STATE Sample a mini-batch $\{ (\mathbf{x}_i, c_i, d_i) \}_{i=1}^m$ from $\mathcal{D}$
   \FOR{$i=1$ \textbf{to} $m$ (in parallel)}
   \STATE $\mathbf{z}_i = f(\mathbf{x}_i)$
   \STATE $\hat{c}_i = g(\mathbf{z}_i)$
   \ENDFOR
   \STATE Calculate $\mathcal{\tilde{L}}_\texttt{\boda}$ using $\{\mathbf{z}_i\}$ based on Eqn.~(\ref{eqn:final-boda-loss})
   \STATE Calculate $\mathcal{L}_{\texttt{CE}}$ using $\frac{1}{m} \sum_{i=1}^m \mathcal{L}(\hat{c}_i, c_i)$
   \STATE Do one training step with loss $\mathcal{L}_\texttt{CE} + \omega \mathcal{\tilde{L}}_\texttt{\boda}$
   \UNTIL{iterate over all training samples at current epoch $e$}
    \STATE \texttt{/*}~~\texttt{Update}~~\texttt{feature}~~\texttt{statistics}~~\texttt{with}~~\texttt{momentum}~~\texttt{updating}~~\texttt{*/}
   \FORALL{$(d,c)\in \mathcal{M}$}
   \STATE Estimate current feature statistics $\{{\boldsymbol{\mu}}_{d,c}, \boldsymbol{{\Sigma}}_{d,c}\}$
   \STATE $\boldsymbol{\mu}_{d,c}^{(e+1)} \leftarrow \alpha\times \boldsymbol{\mu}_{d,c}^{(e)} + (1-\alpha)\times \boldsymbol{\mu}_{d,c}$
   \STATE $\boldsymbol{\Sigma}_{d,c}^{(e+1)} \leftarrow \alpha\times \boldsymbol{\Sigma}_{d,c}^{(e)} + (1-\alpha)\times \boldsymbol{\Sigma}_{d,c}$
   \ENDFOR
   \ENDFOR
\end{algorithmic}
\end{algorithm}

\section{Details of MDLT Datasets}
\label{sec-appendix:mdlt-dataset-details}

In this section, we provide the detailed information of the curated MDLT datasets we used in our experiments. Table \ref{appendix:table:dataset-details} provides an overview of the datasets.
Table \ref{appendix:table:dataset-images} provides the image examples across domains for each MDLT dataset.

\paragraph{\texttt{Digits-MLT}.}
We construct \texttt{Digits-MLT} by combining two digit datasets: (1) MNIST-M \cite{ganin2016dann}, a variant of the original MNIST handwritten digit classification dataset \cite{lecun1998mnist} with colorful background, and (2) SVHN \cite{netzer2011svhn}. The original MNIST-M dataset contains 60,000 training samples and 10,000 testing examples, and the original SVHN dataset contains 73,257 images for training and 26,032 images for testing. Both datasets have examples of dimension $(3, 32, 32)$ and 10 classes. We create \texttt{Digits-MLT} with controllable degrees of data imbalance, where we keep the maximum number of samples each $(d,c)$ to be 1,000, and manually vary the imbalance degree to adjust the number of samples for minority $(d,c)$. For validation and test set, we use the original test set of the two datasets, but keep the number of samples each $(d,c)$ to be 800.

\paragraph{\texttt{VLCS-MLT}.}
The original \texttt{VLCS} dataset \cite{fang2013vlcs} is an object recognition dataset that comprises photographic domains $d \in \{$ \text{Caltech101}, \text{LabelMe}, \text{SUN09}, \text{VOC2007} $\}$, with scenes captured from urban to rural. The dataset contains 5 classes with 10,729 examples of dimension $(3, 224, 224)$. To construct \texttt{VLCS-MLT}, for each $(d,c)$ we split out a validation set of size 15 and a test set of size 30, and leave the rest for training.

\begin{table}[!t]
\setlength{\tabcolsep}{5pt}
\caption{Detailed statistics of the curated MDLT datasets used in our experiments. For the synthetic \texttt{Digits-MLT} dataset, we manually vary the minimum $(d,c)$ size to simulate different degrees of imbalance.}
\vspace{-15pt}
\label{appendix:table:dataset-details}
\small
\begin{center}
\resizebox{\textwidth}{!}{
\begin{tabular}{lccccccc}
\toprule[1.5pt]
\textbf{Dataset} & \textbf{\texttt{\#} Domains} & \textbf{\texttt{\#} Classes} & \textbf{Max $(d,c)$ size} & \textbf{Min $(d,c)$ size} & \textbf{\texttt{\#} Training set} & \textbf{\texttt{\#} Val. set} & \textbf{\texttt{\#} Test set} \\ \midrule\midrule
\texttt{Digits-MLT} & 2 & 10 & 1,000 & 10 $\sim$ 1,000 & 20,000 $\sim$ 4,956 & 16,000 & 16,000 \\ \midrule
\texttt{VLCS-MLT} & 4 & 5 & 1,454 & 0 & 9,872 & 285 & 572 \\ \midrule
\texttt{PACS-MLT} & 4 & 7 & 741 & 5 & 7,891 & 700 & 1,400 \\ \midrule
\texttt{OfficeHome-MLT} & 4 & 65 & 84 & 0 & 11,688 & 1,300 & 2,600 \\ \midrule
\texttt{TerraInc-MLT} & 4 & 10 & 4,455 & 0 & 23,269 & 353 & 708 \\ \midrule
\texttt{DomainNet-MLT} & 6 & 345 & 778 & 0 & 468,574 & 39,240 & 78,761 \\
\bottomrule[1.5pt]
\end{tabular}}
\end{center}
\end{table}

\begin{table}[!t]
\setlength{\tabcolsep}{7pt}
\caption{Overview of images from different domains in all MDLT datasets. For each dataset, we pick a single class and show illustrative images from each domain.}
\vspace{-6pt}
\label{appendix:table:dataset-images}
\begin{center}
\begin{tabular}{lcccccc}
\toprule[1.5pt]
    \textbf{Dataset} & \multicolumn{6}{l}{\textbf{Domains}} \\
    \midrule\midrule
    & \tiny{MNIST-M} & \tiny{SVHN} & & & & \\
    \texttt{Digits-MLT} &
        \raisebox{-.5\height}{\includegraphics[width=25pt, height=25pt]{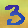}} &
        \raisebox{-.5\height}{\includegraphics[width=25pt, height=25pt]{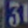}} & & & & \\
    \midrule
    & \tiny{Caltech101} & \tiny{LabelMe} & \tiny{SUN09} & \tiny{VOC2007} & & \\
    \texttt{VLCS-MLT} &
        \raisebox{-.5\height}{\includegraphics[width=25pt, height=25pt]{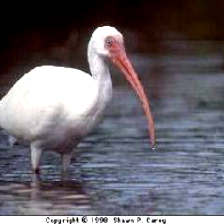}} &
        \raisebox{-.5\height}{\includegraphics[width=25pt, height=25pt]{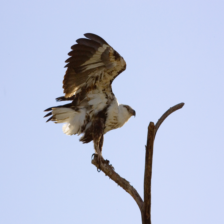}} &
        \raisebox{-.5\height}{\includegraphics[width=25pt, height=25pt]{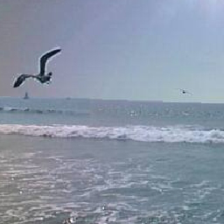}} &
        \raisebox{-.5\height}{\includegraphics[width=25pt, height=25pt]{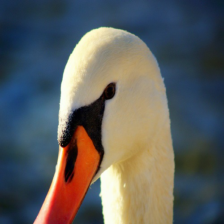}} & &
        \\
    \midrule
    & \tiny{Art} & \tiny{Cartoon} & \tiny{Photo} & \tiny{Sketch} & & \\
    \texttt{PACS-MLT} &
        \raisebox{-.5\height}{\includegraphics[width=25pt, height=25pt]{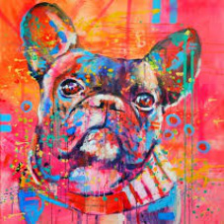}} &
        \raisebox{-.5\height}{\includegraphics[width=25pt, height=25pt]{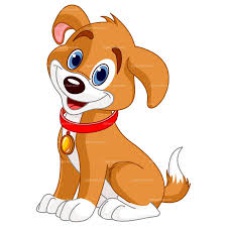}} &
        \raisebox{-.5\height}{\includegraphics[width=25pt, height=25pt]{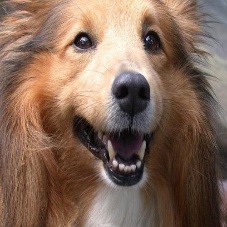}} &
        \raisebox{-.5\height}{\includegraphics[width=25pt, height=25pt]{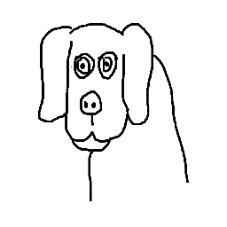}} & &
        \\
    \midrule
    & \tiny{Art} & \tiny{Clipart} & \tiny{Product} & \tiny{Photo} & & \\
    \texttt{OfficeHome-MLT} &
        \raisebox{-.5\height}{\includegraphics[width=25pt, height=25pt]{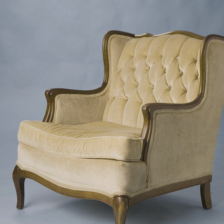}} &
        \raisebox{-.5\height}{\includegraphics[width=25pt, height=25pt]{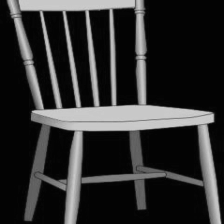}} &
        \raisebox{-.5\height}{\includegraphics[width=25pt, height=25pt]{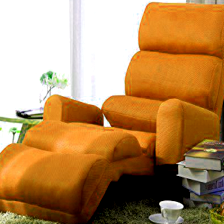}} &
        \raisebox{-.5\height}{\includegraphics[width=25pt, height=25pt]{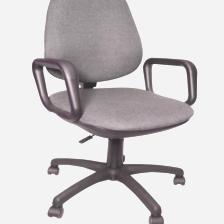}} & &
        \\
    \midrule
    & \tiny{L100} & \tiny{L38} & \tiny{L43} & \tiny{L46} & & \\
    \texttt{TerraInc-MLT} &
        \raisebox{-.5\height}{\includegraphics[width=25pt, height=25pt]{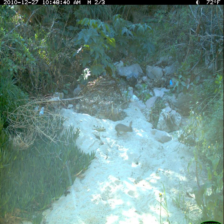}} &
        \raisebox{-.5\height}{\includegraphics[width=25pt, height=25pt]{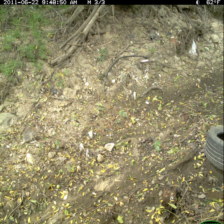}} &
        \raisebox{-.5\height}{\includegraphics[width=25pt, height=25pt]{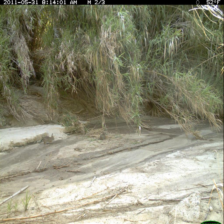}} &
        \raisebox{-.5\height}{\includegraphics[width=25pt, height=25pt]{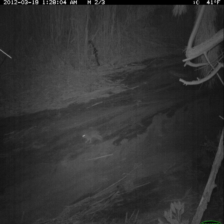}} &
        \\
    & \multicolumn{6}{l}{\tiny{\emph{(camera trap location)}}} \\
    \midrule
    & \tiny{Clipart} & \tiny{Infographic} & \tiny{Painting} & \tiny{QuickDraw} & \tiny{Photo} & \tiny{Sketch} \\
    \texttt{DomainNet-MLT} &
        \raisebox{-.5\height}{\includegraphics[width=25pt, height=25pt]{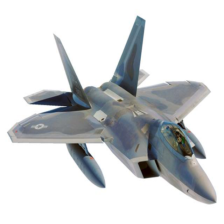}} &
        \raisebox{-.5\height}{\includegraphics[width=25pt, height=25pt]{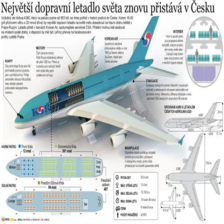}} &
        \raisebox{-.5\height}{\includegraphics[width=25pt, height=25pt]{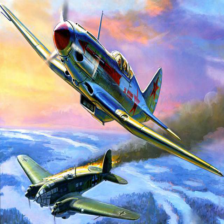}} &
        \raisebox{-.5\height}{\includegraphics[width=25pt, height=25pt]{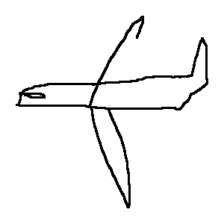}} &
        \raisebox{-.5\height}{\includegraphics[width=25pt, height=25pt]{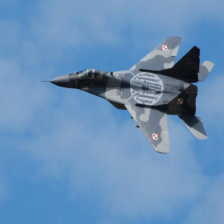}} &
        \raisebox{-.5\height}{\includegraphics[width=25pt, height=25pt]{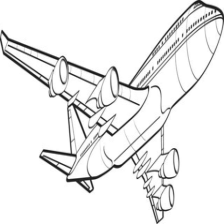}}
        \\
\bottomrule[1.5pt]
\end{tabular}
\end{center}
\end{table}

\paragraph{\texttt{PACS-MLT}.}
The original \texttt{PACS} dataset \cite{li2017pacs} is an object recognition dataset that comprises four domains $d \in \{$ \text{art}, \text{cartoons}, \text{photos}, \text{sketches} $\}$ with image style changes. It contains 7 classes with 9,991 examples of dimension $(3, 224, 224)$. We construct \texttt{PACS-MLT} in a simialr manner as \texttt{VLCS-MLT}, where we split out a validation set of size 25 and a test set of size 50 for each $(d,c)$, and leave the rest for training.

\paragraph{\texttt{OfficeHome-MLT}.}
The original \texttt{OfficeHome} dataset \cite{venkateswara2017officehome} includes domains $d \in \{$ \text{art}, \text{clipart}, \text{product}, \text{real} $\}$, containing 15,588 examples of dimension $(3, 224, 224)$ and 65 classes. We make \texttt{OfficeHome-MLT} by splitting out a validation set of size 5 and a test set of size 10 for each $(d,c)$, leaving the rest for training.

\paragraph{\texttt{TerraInc-MLT}.}
\texttt{TerraInc-MLT} is constructed from \texttt{TerraIncognita} dataset \cite{beery2018recognition}, a species classification dataset that contains photographs of wild animals taken by camera traps at locations $d \in \{ \text{L100}, \text{L38}, \text{L43}, \text{L46}\}$. The dataset contains 10 classes with 24,788 examples of dimension $(3, 224, 224)$. For each $(d,c)$, we split out a validation set of size 10 and a test set of size 20, and use all remaining samples for training.

\paragraph{\texttt{DomainNet-MLT}.}
We construct \texttt{DomainNet-MLT} using \texttt{DomainNet} dataset \cite{peng2019domainnet}, a large-scale multi-domain dataset for object recognition that consists of six domains $d \in \{$ \text{clipart}, \text{infograph}, \text{painting}, \text{quickdraw}, \text{real}, \text{sketch} $\}$, 345 classes, and 586,575 examples of size $(3, 224, 224)$. To construct \texttt{DomainNet-MLT}, for each $(d,c)$ we split out a validation set of size 20 and a test set of size 40, and leave the rest for training.

\section{Experimental Settings}
\label{sec-appendix:exp-settings}

\subsection{Implementation Details}
\label{subsec-appendix:impl-details}

For the synthetic \texttt{Digits-MLT} dataset, we fix the network architecture as a small MNIST CNN \cite{gulrajani2020domainbed} for all algorithms, and use no data augmentation.
For all other MDLT datasets, following \cite{gulrajani2020domainbed}, we use the pretrained ResNet-50 model \cite{he2016deep} as the backbone network for all algorithms, and use the same data augmentation protocol as \cite{gulrajani2020domainbed}: random crop and resize to $224\times 224$ pixels, random horizontal flips, random color jitter, grayscaling the image with 10\% probability, and normalization using the ImageNet channel statistics.
We train all models using the Adam optimizer \cite{kingma2015adam} for 5,000 steps on all MDLT datasets except \texttt{DomainNet-MLT}, on which we train longer for 15,000 steps to ensure convergence.
We fix a batch size of 64 per domain for \texttt{Digits-MLT} experiments, a batch size of 32 per domain for \texttt{DomainNet-MLT} experiments, and a batch size of 24 per domain for experiments on all other datasets.

For all MDLT datasets except \texttt{OfficeHome-MLT} and \texttt{TerraInc-MLT}, we define \emph{many-shot} $(d,c)$ pairs as with over 100 training samples, \emph{medium-shot} as with 20$\sim$100 training samples, and \emph{few-shot} as with under 20 training samples.
For \texttt{OfficeHome-MLT}, we define \emph{many-shot} as $(d,c)$ pairs with over 60 training samples, \emph{medium-shot} as with 20$\sim$60 training samples, and \emph{few-shot} as with under 20 training samples.
For \texttt{TerraInc-MLT}, we define \emph{many-shot} as $(d,c)$ pairs with over 100 training samples, \emph{medium-shot} as with 25$\sim$100 training samples, and \emph{few-shot} as with under 25 training samples.

\subsection{Competing Algorithms}
\label{subsec-appendix:all-algo-details}

We compare \boda to a large number of algorithms that span different learning strategies. We group them according to their categories, and provide detailed descriptions for each algorithm below.
\begin{Itemize}
    \item \emph{Vanilla:} The empirical risk minimization (\textbf{ERM})~\cite{vapnik1999overview} minimizes the sum of errors across all domains and samples.
    \item \emph{Distributionally robust optimization:} Group distributionally robust optimization (\textbf{GroupDRO}) \cite{sagawa2020groupdro} performs ERM while increasing the importance of domains with larger errors.
    \item \emph{Cross-domain data augmentation:} Inter-domain mixup (\textbf{Mixup})~\cite{xu2020interdomain_mixup_aaai} performs ERM on linear interpolations of examples from random pairs of domains and their labels. Style-agnostic network (\textbf{SagNet}) \cite{nam2019sagnet} disentangles style encodings from image content by randomizing and augmenting styles.
    \item \emph{Meta-learning:} Meta-learning for domain generalization (\textbf{MLDG}) \cite{li2018mldg} leverages meta-learning to learn how to generalize across domains.
    \item \emph{Domain-invariant representation learning:} Invariant risk minimization (\textbf{IRM}) \cite{arjovsky2019irm} learns a feature representation such that the optimal linear classifier on top of that representation matches across domains. Domain adversarial neural networks (\textbf{DANN}) \cite{ganin2016dann} employ an adversarial network to match feature distributions. Class-conditional DANN (\textbf{CDANN}) \cite{li2018cdann} builds upon DANN but further matches the conditional distributions across domains for all labels. Deep correlation alignment (\textbf{CORAL}) \cite{sun2016coral} matches the mean and covariance of feature distributions. Maximum mean discrepancy (\textbf{MMD}) \cite{li2018mmd} matches the MMD \cite{gretton2012kernel} of feature distributions.
    \item \emph{Transfer learning:} Marginal transfer learning (\textbf{MTL}) \cite{blanchard2021mtl_marginal_transfer_learning} estimates a mean embedding per domain, passed as a second argument to the classifier.
    \item \emph{Multi-task learning:} Gradient matching for domain generalization (\textbf{Fish}) \cite{shi2021fish} maximizes the inner product between gradients from different domains through a multi-task objective.
    \item \emph{Imbalanced learning:} Focal loss (\textbf{Focal}) \cite{lin2017focal} reduces the relative loss for well-classified samples and focuses on difficult samples. Class-balanced loss (\textbf{CBLoss}) \cite{cui2019class} proposes re-weighting by the inverse effective number of samples. The LDAM loss (\textbf{LDAM}) \cite{cao2019learning} employs a modified marginal loss that favors minority samples more. Balanced-Softmax (\textbf{BSoftmax}) \cite{ren2020bsoftmax} extends Softmax to an unbiased estimation that considers the number of samples of each class. Self-supervised pre-training (\textbf{SSP}) \cite{yang2020rethinking} uses self-supervised learning as a first-stage pre-training to alleviate the network dependence on imbalanced labels. Classifier re-training (\textbf{CRT}) \cite{kang2020decoupling} decomposes the representation and classifier learning into two stages, where it fine-tunes the classifier using class-balanced sampling with representation fixed in the second stage.
\end{Itemize}

\subsection{Hyperparameters Search Protocol}
\label{subsec-appendix:hp-details}

For a fair evaluation across different algorithms, following the training protocol in \cite{gulrajani2020domainbed}, for each algorithm we conduct a random search of 20 trials over a joint distribution of its all hyperparameters. We then use the validation set to select the best hyperparameters for each algorithm, fix them and rerun the experiments under 3 different random seeds to report the final average accuracy (and standard deviation). Such process ensures the comparison is best-versus-best, and the hyperparameters are optimized for all algorithms.

We detail the hyperparameter choices for each algorithm in Table \ref{appendix:table:hyperparameters}.

\begin{table}[!t]
\setlength{\tabcolsep}{5pt}
\caption{Hyperparameters search space for all experiments.}
\vspace{-8pt}
\label{appendix:table:hyperparameters}
\small
\begin{center}
\resizebox{0.9\textwidth}{!}{
\begin{tabular}{llll}
\toprule[1.5pt]
\textbf{Condition} & \textbf{Parameter} & \textbf{Default value} & \textbf{Random distribution} \\
\midrule\midrule
\multicolumn{4}{l}{\emph{\textbf{General:}}} \\
\midrule
\multirow{4}{*}{ResNet}     & learning rate & 0.00005 & $10^{\text{Uniform}(-5, -3.5)}$ \\
                            & dropout & 0 & $\text{RandomChoice}([0, 0.1, 0.5])$ \\
                            & generator learning rate & 0.00005 & $10^{\text{Uniform}(-5, -3.5)}$ \\
                            & discriminator learning rate & 0.00005 & $10^{\text{Uniform}(-5, -3.5)}$ \\
\midrule
\multirow{3}{*}{not ResNet} & learning rate & 0.001 & $10^{\text{Uniform}(-4.5, -3.5)}$ \\
                            & generator learning rate & 0.001 & $10^{\text{Uniform}(-4.5, -2.5)}$ \\
                            & discriminator learning rate & 0.001 & $10^{\text{Uniform}(-4.5, -2.5)}$ \\
\midrule
\multirow{2}{*}{\texttt{Digits-MLT}}        & weight decay & 0 & 0 \\
                            & generator weight decay & 0 & 0 \\
\midrule
\multirow{2}{*}{not \texttt{Digits-MLT}}    & weight decay & 0 & $10^{\text{Uniform}(-6, -2)}$ \\
                            & generator weight decay & 0 & $10^{\text{Uniform}(-6, -2)}$ \\
\midrule\midrule
\multicolumn{4}{l}{\emph{\textbf{Algorithm-specific:}}} \\
\midrule
\multirow{2}{*}{IRM}        & lambda & 100 & $10^{\text{Uniform}(-1, 5)}$ \\
                            & iterations of penalty annealing & 500 & $10^{\text{Uniform}(0, 4)}$ \\
\midrule
GroupDRO                    & eta & 0.01 & $10^{\text{Uniform}(-3, -1)}$ \\
\midrule
Mixup                       & alpha & 0.2 & $10^{\text{Uniform}(0, 4)}$ \\
\midrule
MLDG                        & beta & 1 & $10^{\text{Uniform}(-1, 1)}$ \\
\midrule
CORAL, MMD                  & gamma & 1 & $10^{\text{Uniform}(-1, 1)}$ \\
\midrule
\multirow{5}{1.5cm}{DANN, CDANN} & lambda & 1.0 & $10^{\text{Uniform}(-2, 2)}$ \\
                            & discriminator weight decay & 0 & $10^{\text{Uniform}(-6, -2)}$ \\
                            & discriminator steps        & 1 & $2^{\text{Uniform}(0, 3)}$ \\
                            & gradient penalty           & 0 & $10^{\text{Uniform}(-2, 1)}$ \\
                            & adam $\beta_1$             & 0.5 & $\text{RandomChoice}([0, 0.5])$ \\
\midrule
MTL                         & ema & 0.99 & $\text{RandomChoice}([.5, .9, .99, 1])$ \\
\midrule
SagNet                      & adversary weight & 0.1 & $10^{\text{Uniform}(-2, 1)}$ \\
\midrule
Fish                        & meta learning rate & 0.5 & $\text{RandomChoice}([.05, .1, .5])$ \\
\midrule
Focal                       & gamma & 1 & $0.5 * 10^{\text{Uniform}(0, 1)}$ \\
\midrule
CBLoss                      & beta & 0.9999 & $1 - 10^{\text{Uniform}(-5, -2)}$ \\
\midrule
\multirow{2}{*}{LDAM}       & max\_m & 0.5 & $10^{\text{Uniform}(-1, -0.1)}$ \\
                            & scale & 30 & $\text{RandomChoice}([10, 30])$ \\
\midrule
\multirow{2}{*}{BoDA}       & nu & 1 & $10^{\text{Uniform}(-0.5, 0)}$ \\
                            & \boda loss weight & 0.1 & $10^{\text{Uniform}(-2, -0.5)}$ \\
\bottomrule[1.5pt]
\end{tabular}}
\end{center}
\vspace{-0.4cm}
\end{table}

\subsection{Settings for DG Experiments}
\label{subsec-appendix:dg-setups}

For DG experiments, we strictly follow the training protocols described in \cite{gulrajani2020domainbed}. Across all benchmark DG datasets, we keep the same hyperparameter search space for \boda as in Table \ref{appendix:table:hyperparameters}.
We fix all other training parameters unchanged so that the results of \boda are directly comparable to the results in \cite{gulrajani2020domainbed}.

For model selection, we use the \emph{training-domain validation set} protocol in \cite{gulrajani2020domainbed} with $80\% - 20\%$ training-validation split, and the average out-domain test performance is reported across all runs for each domain.

\section{Complete Results for MDLT}
\label{sec-appendix:complete-results-mdlt}

We provide complete evaluation results on the five MDLT datasets.
In addition to the reported results in the main paper, for each dataset we also include the accuracy on each domain together with the averaged and the worst accuracy.

\subsection{\texttt{VLCS-MLT}}
\vspace{0.5cm}

\begin{table}[H]
\setlength{\tabcolsep}{2.5pt}
\caption{Complete evaluation results on \texttt{VLCS-MLT}.}
\vspace{-10pt}
\label{appendix:table:vlcs-mlt}
\small
\begin{center}
\adjustbox{max width=\textwidth}{
\begin{tabular}{lcccccccccc}
\toprule[1.5pt]
& \multicolumn{6}{c}{\textbf{Accuracy (by domain)}} & \multicolumn{4}{c}{\textbf{Accuracy (by shot)}} \\
\cmidrule(lr){2-7} \cmidrule(lr){8-11}
\textbf{Algorithm} & \textbf{C} & \textbf{L} & \textbf{S} & \textbf{V} & \textbf{Average} & \textbf{Worst} & \textbf{Many} & \textbf{Medium} & \textbf{Few} & \textbf{Zero} \\
\midrule
ERM & 99.3 \scriptsize$\pm0.3$ & 53.6 \scriptsize$\pm1.1$ & 65.9 \scriptsize$\pm1.2$ & 86.4 \scriptsize$\pm0.7$ & 76.3 \scriptsize$\pm0.4$ & 53.6 \scriptsize$\pm1.1$ & 84.6 \scriptsize$\pm0.5$ & 76.6 \scriptsize$\pm0.4$ & $-$ & 32.9 \scriptsize$\pm0.4$ \\
IRM & 99.1 \scriptsize$\pm0.4$ & 52.3 \scriptsize$\pm0.7$ & 68.8 \scriptsize$\pm1.4$ & 86.0 \scriptsize$\pm0.3$ & 76.5 \scriptsize$\pm0.2$ & 52.3 \scriptsize$\pm0.7$ & 85.3 \scriptsize$\pm0.6$ & 75.5 \scriptsize$\pm1.0$ & $-$ & 33.5 \scriptsize$\pm1.0$ \\
GroupDRO & 98.7 \scriptsize$\pm0.3$ & 54.1 \scriptsize$\pm1.3$ & 67.5 \scriptsize$\pm1.5$ & 86.7 \scriptsize$\pm0.3$ & 76.7 \scriptsize$\pm0.4$ & 54.1 \scriptsize$\pm1.3$ & 85.3 \scriptsize$\pm0.9$ & 76.2 \scriptsize$\pm1.0$ & $-$ & 34.5 \scriptsize$\pm2.0$ \\
Mixup & 99.3 \scriptsize$\pm0.3$ & 52.7 \scriptsize$\pm1.3$ & 66.1 \scriptsize$\pm0.0$ & 85.3 \scriptsize$\pm1.1$ & 75.9 \scriptsize$\pm0.1$ & 52.7 \scriptsize$\pm1.3$ & 84.4 \scriptsize$\pm0.2$ & 77.1 \scriptsize$\pm0.6$ & $-$ & 29.2 \scriptsize$\pm1.4$ \\
MLDG & 99.3 \scriptsize$\pm0.3$ & 53.6 \scriptsize$\pm0.5$ & 68.3 \scriptsize$\pm0.4$ & 86.4 \scriptsize$\pm0.5$ & 76.9 \scriptsize$\pm0.2$ & 53.6 \scriptsize$\pm0.5$ & 84.9 \scriptsize$\pm0.3$ & 77.5 \scriptsize$\pm1.0$ & $-$ & 34.4 \scriptsize$\pm0.9$ \\
CORAL & 99.3 \scriptsize$\pm0.3$ & 51.6 \scriptsize$\pm0.7$ & 67.5 \scriptsize$\pm1.8$ & 85.3 \scriptsize$\pm0.9$ & 75.9 \scriptsize$\pm0.5$ & 51.6 \scriptsize$\pm0.7$ & 84.3 \scriptsize$\pm0.6$ & 75.5 \scriptsize$\pm0.5$ & $-$ & 34.5 \scriptsize$\pm0.8$ \\
\end{tabular}}
\end{center}
\end{table}

\newpage

\subsection{\texttt{PACS-MLT}}
\vspace{0.5cm}

\begin{table}[H]
\setlength{\tabcolsep}{2.5pt}
\caption{Complete evaluation results on \texttt{PACS-MLT}.}
\vspace{-10pt}
\label{appendix:table:pacs-mlt}
\small
\begin{center}
\adjustbox{max width=\textwidth}{
\begin{tabular}{lcccccccccc}
\toprule[1.5pt]
& \multicolumn{6}{c}{\textbf{Accuracy (by domain)}} & \multicolumn{4}{c}{\textbf{Accuracy (by shot)}} \\
\cmidrule(lr){2-7} \cmidrule(lr){8-11}
\textbf{Algorithm} & \textbf{A} & \textbf{C} & \textbf{P} & \textbf{S} & \textbf{Average} & \textbf{Worst} & \textbf{Many} & \textbf{Medium} & \textbf{Few} & \textbf{Zero} \\
\midrule
ERM & 96.8 \scriptsize$\pm0.1$ & 97.0 \scriptsize$\pm0.3$ & 98.9 \scriptsize$\pm0.3$ & 95.8 \scriptsize$\pm0.2$ & 97.1 \scriptsize$\pm0.1$ & 95.8 \scriptsize$\pm0.2$ & 97.1 \scriptsize$\pm0.0$ & 97.0 \scriptsize$\pm0.0$ & 98.0 \scriptsize$\pm0.9$ & $-$ \\
IRM & 96.8 \scriptsize$\pm0.1$ & 96.3 \scriptsize$\pm0.7$ & 98.7 \scriptsize$\pm0.2$ & 95.2 \scriptsize$\pm0.4$ & 96.7 \scriptsize$\pm0.2$ & 95.2 \scriptsize$\pm0.4$ & 96.8 \scriptsize$\pm0.2$ & 96.7 \scriptsize$\pm0.7$ & 94.7 \scriptsize$\pm1.4$ & $-$ \\
GroupDRO & 96.9 \scriptsize$\pm0.2$ & 97.0 \scriptsize$\pm0.4$ & 99.0 \scriptsize$\pm0.1$ & 95.3 \scriptsize$\pm0.4$ & 97.0 \scriptsize$\pm0.1$ & 95.3 \scriptsize$\pm0.4$ & 97.3 \scriptsize$\pm0.1$ & 95.3 \scriptsize$\pm1.2$ & 94.7 \scriptsize$\pm3.6$ & $-$ \\
Mixup & 96.5 \scriptsize$\pm0.3$ & 96.9 \scriptsize$\pm0.7$ & 98.5 \scriptsize$\pm0.2$ & 95.1 \scriptsize$\pm0.2$ & 96.7 \scriptsize$\pm0.2$ & 95.1 \scriptsize$\pm0.2$ & 97.0 \scriptsize$\pm0.1$ & 96.7 \scriptsize$\pm0.3$ & 91.3 \scriptsize$\pm2.7$ & $-$ \\
MLDG & 96.6 \scriptsize$\pm0.2$ & 97.2 \scriptsize$\pm0.3$ & 98.5 \scriptsize$\pm0.1$ & 94.1 \scriptsize$\pm0.3$ & 96.6 \scriptsize$\pm0.1$ & 94.1 \scriptsize$\pm0.3$ & 96.8 \scriptsize$\pm0.1$ & 96.3 \scriptsize$\pm0.7$ & 92.7 \scriptsize$\pm0.5$ & $-$ \\
CORAL & 96.9 \scriptsize$\pm0.4$ & 97.0 \scriptsize$\pm0.5$ & 98.3 \scriptsize$\pm0.3$ & 94.3 \scriptsize$\pm0.7$ & 96.6 \scriptsize$\pm0.5$ & 94.3 \scriptsize$\pm0.7$ & 96.6 \scriptsize$\pm0.5$ & 97.0 \scriptsize$\pm0.8$ & 94.7 \scriptsize$\pm0.5$ & $-$ \\
\end{tabular}}
\end{center}
\end{table}

\newpage

\subsection{\texttt{OfficeHome-MLT}}
\vspace{0.5cm}

\begin{table}[H]
\setlength{\tabcolsep}{2.5pt}
\caption{Complete evaluation results on \texttt{OfficeHome-MLT}.}
\vspace{-10pt}
\label{appendix:table:officehome-mlt}
\small
\begin{center}
\adjustbox{max width=\textwidth}{
\begin{tabular}{lcccccccccc}
\toprule[1.5pt]
& \multicolumn{6}{c}{\textbf{Accuracy (by domain)}} & \multicolumn{4}{c}{\textbf{Accuracy (by shot)}} \\
\cmidrule(lr){2-7} \cmidrule(lr){8-11}
\textbf{Algorithm} & \textbf{A} & \textbf{C} & \textbf{P} & \textbf{R} & \textbf{Average} & \textbf{Worst} & \textbf{Many} & \textbf{Medium} & \textbf{Few} & \textbf{Zero} \\
\midrule
ERM & 71.3 \scriptsize$\pm0.1$ & 78.4 \scriptsize$\pm0.2$ & 89.6 \scriptsize$\pm0.3$ & 83.3 \scriptsize$\pm0.2$ & 80.7 \scriptsize$\pm0.0$ & 71.3 \scriptsize$\pm0.1$ & 87.8 \scriptsize$\pm0.2$ & 81.0 \scriptsize$\pm0.2$ & 63.1 \scriptsize$\pm0.1$ & 63.3 \scriptsize$\pm7.2$ \\
IRM & 70.7 \scriptsize$\pm0.2$ & 78.5 \scriptsize$\pm0.8$ & 89.4 \scriptsize$\pm0.5$ & 83.8 \scriptsize$\pm0.6$ & 80.6 \scriptsize$\pm0.4$ & 70.7 \scriptsize$\pm0.2$ & 87.6 \scriptsize$\pm0.4$ & 81.5 \scriptsize$\pm0.4$ & 61.1 \scriptsize$\pm0.9$ & 56.7 \scriptsize$\pm1.4$ \\
GroupDRO & 68.7 \scriptsize$\pm0.9$ & 79.0 \scriptsize$\pm0.2$ & 89.4 \scriptsize$\pm0.4$ & 83.3 \scriptsize$\pm0.5$ & 80.1 \scriptsize$\pm0.3$ & 68.7 \scriptsize$\pm0.9$ & 88.1 \scriptsize$\pm0.2$ & 80.8 \scriptsize$\pm0.4$ & 59.8 \scriptsize$\pm1.2$ & 51.7 \scriptsize$\pm3.6$ \\
Mixup & 72.3 \scriptsize$\pm0.6$ & 79.1 \scriptsize$\pm0.4$ & 89.7 \scriptsize$\pm0.1$ & 83.9 \scriptsize$\pm0.2$ & 81.2 \scriptsize$\pm0.2$ & 72.3 \scriptsize$\pm0.6$ & 87.9 \scriptsize$\pm0.4$ & 81.8 \scriptsize$\pm0.1$ & 64.1 \scriptsize$\pm0.4$ & 60.0 \scriptsize$\pm4.1$ \\
MLDG & 70.2 \scriptsize$\pm0.6$ & 78.2 \scriptsize$\pm0.5$ & 89.4 \scriptsize$\pm0.4$ & 83.7 \scriptsize$\pm0.3$ & 80.4 \scriptsize$\pm0.2$ & 70.2 \scriptsize$\pm0.6$ & 87.1 \scriptsize$\pm0.1$ & 81.3 \scriptsize$\pm0.3$ & 61.3 \scriptsize$\pm1.0$ & 61.7 \scriptsize$\pm1.4$ \\
\end{tabular}}
\end{center}
\end{table}

\newpage

\subsection{\texttt{TerraInc-MLT}}
\vspace{0.5cm}

\begin{table}[H]
\setlength{\tabcolsep}{2.5pt}
\caption{Complete evaluation results on \texttt{TerraInc-MLT}.}
\vspace{-10pt}
\label{appendix:table:terrainc-mlt}
\small
\begin{center}
\adjustbox{max width=\textwidth}{
\begin{tabular}{lcccccccccc}
\toprule[1.5pt]
& \multicolumn{6}{c}{\textbf{Accuracy (by domain)}} & \multicolumn{4}{c}{\textbf{Accuracy (by shot)}} \\
\cmidrule(lr){2-7} \cmidrule(lr){8-11}
\textbf{Algorithm} & \textbf{L100} & \textbf{L38} & \textbf{L43} & \textbf{L46} & \textbf{Average} & \textbf{Worst} & \textbf{Many} & \textbf{Medium} & \textbf{Few} & \textbf{Zero} \\
\midrule
ERM & 80.3 \scriptsize$\pm1.3$ & 71.2 \scriptsize$\pm0.7$ & 82.2 \scriptsize$\pm0.3$ & 67.4 \scriptsize$\pm0.3$ & 75.3 \scriptsize$\pm0.3$ & 67.4 \scriptsize$\pm0.3$ & 85.6 \scriptsize$\pm0.8$ & 69.6 \scriptsize$\pm3.2$ & 66.1 \scriptsize$\pm2.4$ & 14.4 \scriptsize$\pm2.8$ \\
IRM & 78.2 \scriptsize$\pm0.9$ & 69.6 \scriptsize$\pm2.0$ & 81.1 \scriptsize$\pm0.7$ & 64.3 \scriptsize$\pm1.3$ & 73.3 \scriptsize$\pm0.7$ & 64.3 \scriptsize$\pm1.3$ & 83.5 \scriptsize$\pm0.6$ & 70.0 \scriptsize$\pm1.8$ & 58.3 \scriptsize$\pm3.4$ & 20.1 \scriptsize$\pm1.4$ \\
GroupDRO & 68.3 \scriptsize$\pm1.0$ & 68.8 \scriptsize$\pm1.3$ & 82.6 \scriptsize$\pm0.2$ & 68.1 \scriptsize$\pm0.8$ & 72.0 \scriptsize$\pm0.4$ & 66.6 \scriptsize$\pm0.2$ & 84.7 \scriptsize$\pm1.1$ & 64.6 \scriptsize$\pm4.7$ & 38.9 \scriptsize$\pm1.2$ & 13.5 \scriptsize$\pm1.1$ \\
Mixup & 75.4 \scriptsize$\pm1.4$ & 70.2 \scriptsize$\pm1.3$ & 78.3 \scriptsize$\pm0.6$ & 60.4 \scriptsize$\pm1.1$ & 71.1 \scriptsize$\pm0.7$ & 60.4 \scriptsize$\pm1.1$ & 83.2 \scriptsize$\pm0.7$ & 60.0 \scriptsize$\pm0.6$ & 56.1 \scriptsize$\pm3.0$ & 12.2 \scriptsize$\pm2.1$ \\
MLDG & 82.3 \scriptsize$\pm0.9$ & 73.5 \scriptsize$\pm2.0$ & 83.8 \scriptsize$\pm1.4$ & 66.9 \scriptsize$\pm0.5$ & 76.6 \scriptsize$\pm0.2$ & 66.9 \scriptsize$\pm0.5$ & 86.1 \scriptsize$\pm0.6$ & 73.8 \scriptsize$\pm3.9$ & 70.6 \scriptsize$\pm3.7$ & 18.8 \scriptsize$\pm2.4$ \\
\end{tabular}}
\end{center}
\end{table}

\newpage

\subsection{\texttt{DomainNet-MLT}}
\vspace{0.5cm}

\begin{table}[H]
\setlength{\tabcolsep}{2.5pt}
\caption{Complete evaluation results on \texttt{DomainNet-MLT}.}
\vspace{-10pt}
\label{appendix:table:domainnet-mlt}
\small
\begin{center}
\adjustbox{max width=\textwidth}{
\begin{tabular}{lcccccccccccc}
\toprule[1.5pt]
& \multicolumn{8}{c}{\textbf{Accuracy (by domain)}} & \multicolumn{4}{c}{\textbf{Accuracy (by shot)}} \\
\cmidrule(lr){2-9} \cmidrule(lr){10-13}
\textbf{Algorithm} & \textbf{clip} & \textbf{info} & \textbf{paint} & \textbf{quick} & \textbf{real} & \textbf{sketch} & \textbf{Average} & \textbf{Worst} & \textbf{Many} & \textbf{Medium} & \textbf{Few} & \textbf{Zero} \\
\midrule
ERM & 68.6 \scriptsize$\pm0.1$ & 29.4 \scriptsize$\pm0.3$ & 57.1 \scriptsize$\pm0.2$ & 62.8 \scriptsize$\pm0.3$ & 72.1 \scriptsize$\pm0.2$ & 61.7 \scriptsize$\pm0.2$ & 58.6 \scriptsize$\pm0.2$ & 29.4 \scriptsize$\pm0.3$ & 66.0 \scriptsize$\pm0.1$ & 56.1 \scriptsize$\pm0.1$ & 35.9 \scriptsize$\pm0.5$ & 27.6 \scriptsize$\pm0.3$ \\
IRM & 66.7 \scriptsize$\pm0.2$ & 27.6 \scriptsize$\pm0.1$ & 56.0 \scriptsize$\pm0.2$ & 60.1 \scriptsize$\pm0.1$ & 72.0 \scriptsize$\pm0.0$ & 60.2 \scriptsize$\pm0.2$ & 57.1 \scriptsize$\pm0.1$ & 27.6 \scriptsize$\pm0.1$ & 64.7 \scriptsize$\pm0.1$ & 54.3 \scriptsize$\pm0.3$ & 33.5 \scriptsize$\pm0.3$ & 25.8 \scriptsize$\pm0.3$ \\
GroupDRO & 60.1 \scriptsize$\pm0.2$ & 25.9 \scriptsize$\pm0.2$ & 50.3 \scriptsize$\pm0.1$ & 63.9 \scriptsize$\pm0.2$ & 64.9 \scriptsize$\pm0.2$ & 56.7 \scriptsize$\pm0.3$ & 53.6 \scriptsize$\pm0.1$ & 25.9 \scriptsize$\pm0.2$ & 61.8 \scriptsize$\pm0.1$ & 49.1 \scriptsize$\pm0.3$ & 30.7 \scriptsize$\pm0.7$ & 22.0 \scriptsize$\pm0.1$ \\
Mixup & 67.6 \scriptsize$\pm0.2$ & 28.7 \scriptsize$\pm0.0$ & 56.4 \scriptsize$\pm0.2$ & 60.0 \scriptsize$\pm0.4$ & 72.1 \scriptsize$\pm0.1$ & 60.9 \scriptsize$\pm0.1$ & 57.6 \scriptsize$\pm0.1$ & 28.7 \scriptsize$\pm0.0$ & 64.9 \scriptsize$\pm0.2$ & 54.5 \scriptsize$\pm0.1$ & 35.6 \scriptsize$\pm0.2$ & 27.3 \scriptsize$\pm0.3$ \\
\end{tabular}}
\end{center}
\end{table}

\newpage

\section{Complete Results for DG}
\label{sec-appendix:complete-results-dg}

We provide detailed results of Table \ref{table:avg-all-dg} across five DG benchmarks \cite{gulrajani2020domainbed}.
Results for all algorithms except \boda are directly copied from \cite{gulrajani2020domainbed}.

\subsection{\texttt{VLCS}}
\vspace{0.4cm}

\begin{table}[H]
\setlength{\tabcolsep}{5pt}
\caption{Complete domain generalization results on \texttt{VLCS}.}
\vspace{-6pt}
\label{appendix:table:dg-vlcs}
\small
\begin{center}
\begin{tabular}{lcccc|c}
\toprule[1.5pt]
\textbf{Algorithm} & \textbf{C} & \textbf{L} & \textbf{S} & \textbf{V} & \textbf{Avg} \\
\midrule
ERM & 97.7 \scriptsize$\pm0.4$ & 64.3 \scriptsize$\pm0.9$ & 73.4 \scriptsize$\pm0.5$ & 74.6 \scriptsize$\pm1.3$ & 77.5 \\
IRM & 98.6 \scriptsize$\pm0.1$ & 64.9 \scriptsize$\pm0.9$ & 73.4 \scriptsize$\pm0.6$ & 77.3 \scriptsize$\pm0.9$ & 78.5 \\
GroupDRO & 97.3 \scriptsize$\pm0.3$ & 63.4 \scriptsize$\pm0.9$ & 69.5 \scriptsize$\pm0.8$ & 76.7 \scriptsize$\pm0.7$ & 76.7 \\
Mixup & 98.3 \scriptsize$\pm0.6$ & 64.8 \scriptsize$\pm1.0$ & 72.1 \scriptsize$\pm0.5$ & 74.3 \scriptsize$\pm0.8$ & 77.4 \\
MLDG & 97.4 \scriptsize$\pm0.2$ & 65.2 \scriptsize$\pm0.7$ & 71.0 \scriptsize$\pm1.4$ & 75.3 \scriptsize$\pm1.0$ & 77.2 \\
CORAL & 98.3 \scriptsize$\pm0.1$ & \textbf{66.1} \scriptsize$\pm1.2$ & 73.4 \scriptsize$\pm0.3$ & 77.5 \scriptsize$\pm1.2$ & \textbf{78.8} \\
MMD & 97.7 \scriptsize$\pm0.1$ & 64.0 \scriptsize$\pm1.1$ & 72.8 \scriptsize$\pm0.2$ & 75.3 \scriptsize$\pm3.3$ & 77.5 \\
\end{tabular}
\end{center}
\end{table}

\vspace{0.5cm}
\subsection{\texttt{PACS}}
\vspace{0.4cm}

\begin{table}[H]
\setlength{\tabcolsep}{5pt}
\caption{Complete domain generalization results on \texttt{PACS}.}
\vspace{-6pt}
\label{appendix:table:dg-pacs}
\small
\begin{center}
\begin{tabular}{lcccc|c}
\toprule[1.5pt]
\textbf{Algorithm} & \textbf{A} & \textbf{C} & \textbf{P} & \textbf{S} & \textbf{Avg} \\
\midrule
ERM & 84.7 \scriptsize$\pm0.4$ & 80.8 \scriptsize$\pm0.6$ & 97.2 \scriptsize$\pm0.3$ & 79.3 \scriptsize$\pm1.0$ & 85.5 \\
IRM & 84.8 \scriptsize$\pm1.3$ & 76.4 \scriptsize$\pm1.1$ & 96.7 \scriptsize$\pm0.6$ & 76.1 \scriptsize$\pm1.0$ & 83.5 \\
GroupDRO & 83.5 \scriptsize$\pm0.9$ & 79.1 \scriptsize$\pm0.6$ & 96.7 \scriptsize$\pm0.3$ & 78.3 \scriptsize$\pm2.0$ & 84.4 \\
Mixup & 86.1 \scriptsize$\pm0.5$ & 78.9 \scriptsize$\pm0.8$ & 97.6 \scriptsize$\pm0.1$ & 75.8 \scriptsize$\pm1.8$ & 84.6 \\
MLDG & 85.5 \scriptsize$\pm1.4$ & 80.1 \scriptsize$\pm1.7$ & 97.4 \scriptsize$\pm0.3$ & 76.6 \scriptsize$\pm1.1$ & 84.9 \\
CORAL & \textbf{88.3} \scriptsize$\pm0.2$ & 80.0 \scriptsize$\pm0.5$ & 97.5 \scriptsize$\pm0.3$ & 78.8 \scriptsize$\pm1.3$ & 86.2 \\
MMD & 86.1 \scriptsize$\pm1.4$ & 79.4 \scriptsize$\pm0.9$ & 96.6 \scriptsize$\pm0.2$ & 76.5 \scriptsize$\pm0.5$ & 84.6 \\
\end{tabular}
\end{center}
\end{table}

\newpage

\subsection{\texttt{OfficeHome}}
\vspace{0.5cm}

\begin{table}[H]
\setlength{\tabcolsep}{5pt}
\caption{Complete domain generalization results on \texttt{OfficeHome}.}
\vspace{-6pt}
\label{appendix:table:dg-officehome}
\small
\begin{center}
\begin{tabular}{lcccc|c}
\toprule[1.5pt]
\textbf{Algorithm} & \textbf{A} & \textbf{C} & \textbf{P} & \textbf{R} & \textbf{Avg} \\
\midrule
ERM & 61.3 \scriptsize$\pm0.7$ & 52.4 \scriptsize$\pm0.3$ & 75.8 \scriptsize$\pm0.1$ & 76.6 \scriptsize$\pm0.3$ & 66.5 \\
IRM & 58.9 \scriptsize$\pm2.3$ & 52.2 \scriptsize$\pm1.6$ & 72.1 \scriptsize$\pm2.9$ & 74.0 \scriptsize$\pm2.5$ & 64.3 \\
GroupDRO & 60.4 \scriptsize$\pm0.7$ & 52.7 \scriptsize$\pm1.0$ & 75.0 \scriptsize$\pm0.7$ & 76.0 \scriptsize$\pm0.7$ & 66.0 \\
Mixup & 62.4 \scriptsize$\pm0.8$ & 54.8 \scriptsize$\pm0.6$ & 76.9 \scriptsize$\pm0.3$ & 78.3 \scriptsize$\pm0.2$ & 68.1 \\
MLDG & 61.5 \scriptsize$\pm0.9$ & 53.2 \scriptsize$\pm0.6$ & 75.0 \scriptsize$\pm1.2$ & 77.5 \scriptsize$\pm0.4$ & 66.8 \\
CORAL & 65.3 \scriptsize$\pm0.4$ & 54.4 \scriptsize$\pm0.5$ & 76.5 \scriptsize$\pm0.1$ & 78.4 \scriptsize$\pm0.5$ & 68.7 \\
MMD & 60.4 \scriptsize$\pm0.2$ & 53.3 \scriptsize$\pm0.3$ & 74.3 \scriptsize$\pm0.1$ & 77.4 \scriptsize$\pm0.6$ & 66.3 \\
\end{tabular}
\end{center}
\end{table}

\vspace{0.8cm}
\subsection{\texttt{TerraInc}}
\vspace{0.5cm}

\begin{table}[H]
\setlength{\tabcolsep}{5pt}
\caption{Complete domain generalization results on \texttt{TerraInc}.}
\vspace{-6pt}
\label{appendix:table:dg-terrainc}
\small
\begin{center}
\begin{tabular}{lcccc|c}
\toprule[1.5pt]
\textbf{Algorithm} & \textbf{L100} & \textbf{L38} & \textbf{L43} & \textbf{L46} & \textbf{Avg} \\
\midrule
ERM & 49.8 \scriptsize$\pm4.4$ & 42.1 \scriptsize$\pm1.4$ & 56.9 \scriptsize$\pm1.8$ & 35.7 \scriptsize$\pm3.9$ & 46.1 \\
IRM & 54.6 \scriptsize$\pm1.3$ & 39.8 \scriptsize$\pm1.9$ & 56.2 \scriptsize$\pm1.8$ & 39.6 \scriptsize$\pm0.8$ & 47.6 \\
GroupDRO & 41.2 \scriptsize$\pm0.7$ & 38.6 \scriptsize$\pm2.1$ & 56.7 \scriptsize$\pm0.9$ & 36.4 \scriptsize$\pm2.1$ & 43.2 \\
Mixup & \textbf{59.6} \scriptsize$\pm2.0$ & 42.2 \scriptsize$\pm1.4$ & 55.9 \scriptsize$\pm0.8$ & 33.9 \scriptsize$\pm1.4$ & 47.9 \\
MLDG & 54.2 \scriptsize$\pm3.0$ & 44.3 \scriptsize$\pm1.1$ & 55.6 \scriptsize$\pm0.3$ & 36.9 \scriptsize$\pm2.2$ & 47.7 \\
CORAL & 51.6 \scriptsize$\pm2.4$ & 42.2 \scriptsize$\pm1.0$ & 57.0 \scriptsize$\pm1.0$ & 39.8 \scriptsize$\pm2.9$ & 47.6 \\
MMD & 41.9 \scriptsize$\pm3.0$ & 34.8 \scriptsize$\pm1.0$ & 57.0 \scriptsize$\pm1.9$ & 35.2 \scriptsize$\pm1.8$ & 42.2 \\
\end{tabular}
\end{center}
\end{table}

\newpage

\subsection{\texttt{DomainNet}}
\vspace{0.5cm}

\begin{table}[H]
\setlength{\tabcolsep}{5pt}
\caption{Complete domain generalization results on \texttt{DomainNet}.}
\vspace{-6pt}
\label{appendix:table:dg-domainnet}
\small
\begin{center}
\adjustbox{max width=\textwidth}{
\begin{tabular}{lcccccc|c}
\toprule[1.5pt]
\textbf{Algorithm} & \textbf{clip} & \textbf{info} & \textbf{paint} & \textbf{quick} & \textbf{real} & \textbf{sketch} & \textbf{Avg} \\
\midrule
ERM & 58.1 \scriptsize$\pm0.3$ & 18.8 \scriptsize$\pm0.3$ & 46.7 \scriptsize$\pm0.3$ & 12.2 \scriptsize$\pm0.4$ & 59.6 \scriptsize$\pm0.1$ & 49.8 \scriptsize$\pm0.4$ & 40.9 \\
IRM & 48.5 \scriptsize$\pm2.8$ & 15.0 \scriptsize$\pm1.5$ & 38.3 \scriptsize$\pm4.3$ & 10.9 \scriptsize$\pm0.5$ & 48.2 \scriptsize$\pm5.2$ & 42.3 \scriptsize$\pm3.1$ & 33.9 \\
GroupDRO & 47.2 \scriptsize$\pm0.5$ & 17.5 \scriptsize$\pm0.4$ & 33.8 \scriptsize$\pm0.5$ & 9.3 \scriptsize$\pm0.3$ & 51.6 \scriptsize$\pm0.4$ & 40.1 \scriptsize$\pm0.6$ & 33.3 \\
Mixup & 55.7 \scriptsize$\pm0.3$ & 18.5 \scriptsize$\pm0.5$ & 44.3 \scriptsize$\pm0.5$ & 12.5 \scriptsize$\pm0.4$ & 55.8 \scriptsize$\pm0.3$ & 48.2 \scriptsize$\pm0.5$ & 39.2 \\
MLDG & 59.1 \scriptsize$\pm0.2$ & 19.1 \scriptsize$\pm0.3$ & 45.8 \scriptsize$\pm0.7$ & 13.4 \scriptsize$\pm0.3$ & 59.6 \scriptsize$\pm0.2$ & 50.2 \scriptsize$\pm0.4$ & 41.2 \\
CORAL & 59.2 \scriptsize$\pm0.1$ & 19.7 \scriptsize$\pm0.2$ & 46.6 \scriptsize$\pm0.3$ & 13.4 \scriptsize$\pm0.4$ & 59.8 \scriptsize$\pm0.2$ & 50.1 \scriptsize$\pm0.6$ & 41.5 \\
MMD & 32.1 \scriptsize$\pm13.3$ & 11.0 \scriptsize$\pm4.6$ & 26.8 \scriptsize$\pm11.3$ & 8.7 \scriptsize$\pm2.1$ & 32.7 \scriptsize$\pm13.8$ & 28.9 \scriptsize$\pm11.9$ & 23.4 \\
\end{tabular}}
\end{center}
\end{table}

\vspace{0.8cm}
\subsection{Averages}
\vspace{0.5cm}

\begin{table}[H]
\setlength{\tabcolsep}{5pt}
\caption{Complete domain generalization results over all DG benchmarks.}
\vspace{-6pt}
\label{appendix:table:dg-avg}
\small
\begin{center}
\adjustbox{max width=\textwidth}{
\begin{tabular}{lccccc|c}
\toprule[1.5pt]
\textbf{Algorithm} & \texttt{VLCS} & \texttt{PACS} & \texttt{OfficeHome} & \texttt{TerraInc} & \texttt{DomainNet} & \textbf{Avg} \\
\midrule
ERM & 77.5 \scriptsize$\pm0.4$ & 85.5 \scriptsize$\pm0.2$ & 66.5 \scriptsize$\pm0.3$ & 46.1 \scriptsize$\pm1.8$ & 40.9 \scriptsize$\pm0.1$ & 63.3 \\
IRM & 78.5 \scriptsize$\pm0.5$ & 83.5 \scriptsize$\pm0.8$ & 64.3 \scriptsize$\pm2.2$ & 47.6 \scriptsize$\pm0.8$ & 33.9 \scriptsize$\pm2.8$ & 61.6 \\
GroupDRO & 76.7 \scriptsize$\pm0.6$ & 84.4 \scriptsize$\pm0.8$ & 66.0 \scriptsize$\pm0.7$ & 43.2  \scriptsize$\pm1.1$ & 33.3 \scriptsize$\pm0.2$ & 60.7 \\
Mixup & 77.4 \scriptsize$\pm0.6$ & 84.6 \scriptsize$\pm0.6$ & 68.1 \scriptsize$\pm0.3$ & 47.9 \scriptsize$\pm0.8$ & 39.2 \scriptsize$\pm0.1$ & 63.4 \\
MLDG & 77.2 \scriptsize$\pm0.4$ & 84.9 \scriptsize$\pm1.0$ & 66.8 \scriptsize$\pm0.6$ & 47.7 \scriptsize$\pm0.9$ & 41.2 \scriptsize$\pm0.1$ & 63.6 \\
CORAL & \textbf{78.8} \scriptsize$\pm0.6$ & 86.2 \scriptsize$\pm0.3$ & 68.7 \scriptsize$\pm0.3$ & 47.6 \scriptsize$\pm1.0$ & 41.5 \scriptsize$\pm0.1$ & 64.5 \\
MMD & 77.5 \scriptsize$\pm0.9$ & 84.6 \scriptsize$\pm0.5$ & 66.3 \scriptsize$\pm0.1$ & 42.2 \scriptsize$\pm1.6$ & 23.4 \scriptsize$\pm9.5$ & 58.8 \\
\end{tabular}}
\end{center}
\end{table}

\newpage

\section{Additional Analysis and Studies}
\label{sec-appendix:additional-study}

\subsection{Ablation Studies for \boda}
\label{subsec-appendix:ablation}

\paragraph{Effect of Balanced Distance.}
We study the effect of adding balanced distance in \boda compared to the vanilla \texttt{DA} loss. As Table \ref{appendix:table:ablation-bdist} demonstrates, incorporating balanced distance in \boda is essential for addressing MDLT: we observe that \boda improves over \texttt{DA} by a large margin, resulting in an averaged improvements of $2.3\%$ over all MDLT benchmarks. The improvements are especially large on datasets with severe data imbalance across domains (e.g., \texttt{TerraInc-MLT}).

\vspace{0.8cm}
\begin{table}[H]
\setlength{\tabcolsep}{5pt}
\caption{Ablation study on effect of adding balanced distance in \boda.}
\vspace{-7pt}
\label{appendix:table:ablation-bdist}
\small
\begin{center}
\adjustbox{max width=\textwidth}{
\begin{tabular}{lccccc|c}
\toprule[1.5pt]
& \textbf{\texttt{VLCS-MLT}} & \textbf{\texttt{PACS-MLT}} & \textbf{\texttt{OfficeHome-MLT}} & \textbf{\texttt{TerraInc-MLT}} & \textbf{\texttt{DomainNet-MLT}} & \textbf{Avg} \\
\midrule \midrule
\texttt{DA} & 76.6 \scriptsize$\pm0.4$ & 96.8 \scriptsize$\pm0.2$ & 80.7 \scriptsize$\pm0.3$ & 76.4 \scriptsize$\pm0.5$ & 58.9 \scriptsize$\pm0.2$ & 77.9 \\
\boda & \textbf{77.3} \scriptsize$\pm0.2$ & \textbf{97.2} \scriptsize$\pm0.1$ & \textbf{82.3} \scriptsize$\pm0.1$ & \textbf{82.3} \scriptsize$\pm0.3$ & \textbf{61.7} \scriptsize$\pm0.1$ & \textbf{80.2} \\
\midrule
Gains & \textcolor{darkgreen}{\texttt{+}\textbf{0.7}} & \textcolor{darkgreen}{\texttt{+}\textbf{0.4}} & \textcolor{darkgreen}{\texttt{+}\textbf{1.6}} & \textcolor{darkgreen}{\texttt{+}\textbf{5.9}} & \textcolor{darkgreen}{\texttt{+}\textbf{2.8}} & \textcolor{darkgreen}{\texttt{+}\textbf{2.3}} \\
\bottomrule[1.5pt]
\end{tabular}}
\end{center}
\vspace{0.5cm}
\end{table}

\paragraph{Effect of Different Distance Calibration Coefficient $\lambda^{d',c'}_{d,c}$.}
We further investigate the effect of different distance calibration coefficients in \boda. Recall that $\lambda^{d',c'}_{d,c} = \left(N_{d',c'} / N_{d,c}\right)^{\nu}$ indicates how much we would like to transfer $(d,c)$ to $(d',c')$, based on their relative sample sizes. We vary the value of $\nu$, and study its effect on \boda performance across all MDLT datasets.
Table \ref{appendix:table:ablation-cdist} reveals several interesting findings.
First, when $\nu=0$ (i.e., no calibration is used as the coefficient is always equal to 1), \boda performance is lower than those with a positive $\nu$, confirming the effectiveness of the calibrated distance.
Moreover, when we vary $\nu$ between $0.5 - 1.5$, the overall performance gains are similar across different choices, where $\nu$ around 0.9 seems to achieve the best results.
Finally, when compared to ERM, we demonstrate that \boda consistently obtains notable gains across different $\nu$.

\vspace{0.8cm}
\begin{table}[H]
\setlength{\tabcolsep}{8pt}
\caption{Ablation study on effect of distance calibration coefficient $\lambda^{d',c'}_{d,c}$ in \boda. We vary the value of $\nu$ and report the averaged results over all five MDLT datasets.}
\vspace{-6pt}
\label{appendix:table:ablation-cdist}
\small
\begin{center}
\adjustbox{max width=\textwidth}{
\begin{tabular}{lcccccccc|c}
\toprule[1.5pt]
$\nu$ & 0 & 0.5 & 0.7 & 0.9 & 1 & 1.1 & 1.2 & 1.5 & \textcolor{gray}{ERM} \\
\midrule
\boda & 78.9 & 80.1 & 80.0 & 80.2 & 80.1 & 79.8 & 79.6 & 79.2 & \textcolor{gray}{77.6} \\
\bottomrule[1.5pt]
\end{tabular}}
\end{center}
\end{table}

\subsection{Absolute Accuracy Gains on All MDLT Benchmarks}
\label{subsec-appendix:abs-gains}

\begin{figure}[ht]
\centering
\includegraphics[width=\textwidth]{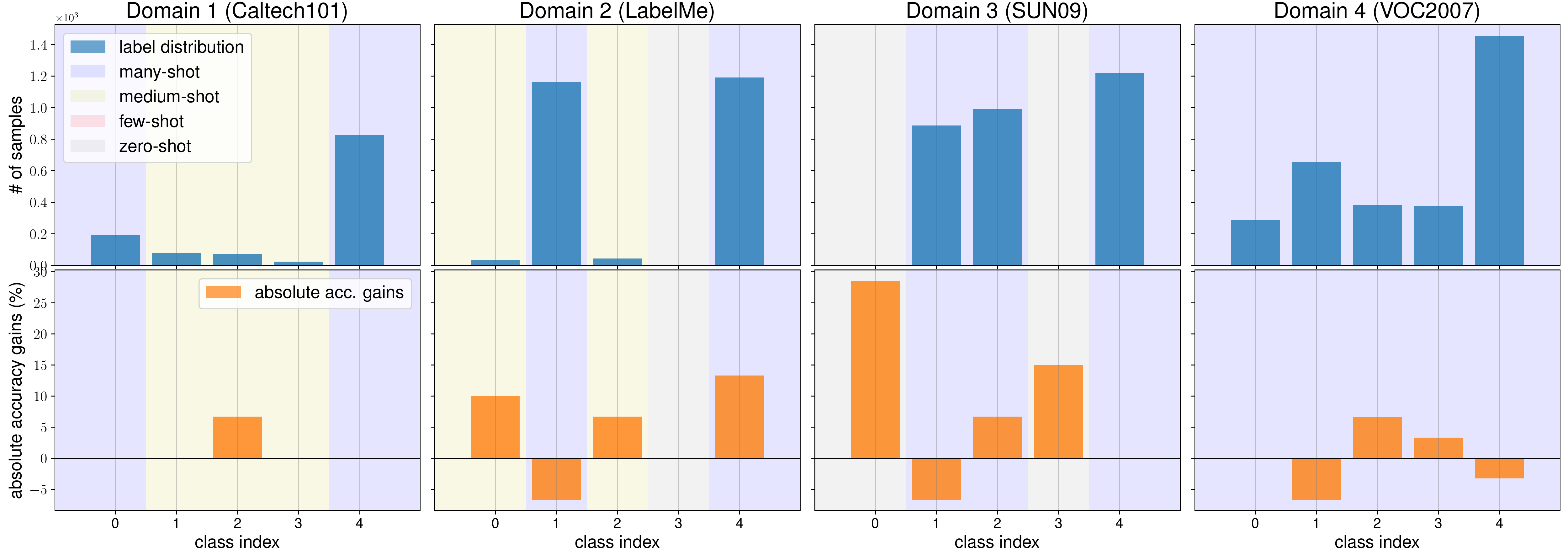}
\vspace{-0.5cm}
\caption{The absolute accuracy gains of \boda \emph{vs.} ERM over all domain-class pairs on \texttt{VLCS-MLT}.}
\label{appendix:fig:gains-vlcs}
\end{figure}

\begin{figure}[ht]
\centering
\includegraphics[width=\textwidth]{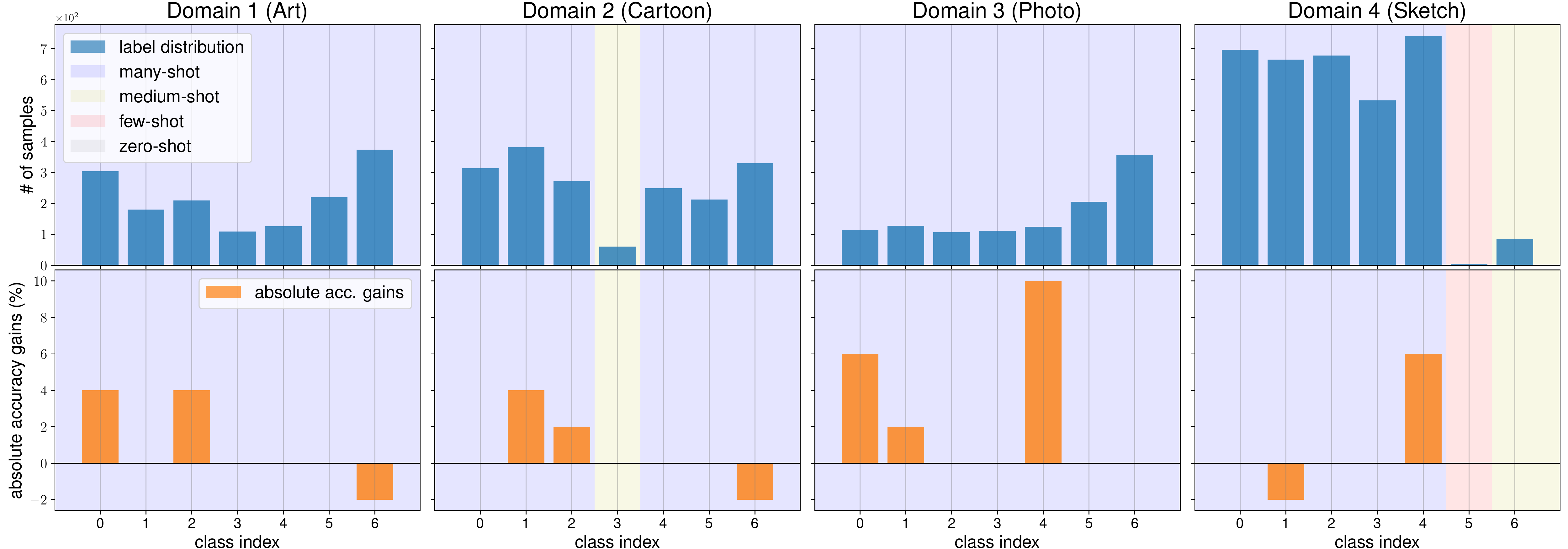}
\vspace{-0.5cm}
\caption{The absolute accuracy gains of \boda \emph{vs.} ERM over all domain-class pairs on \texttt{PACS-MLT}.}
\label{appendix:fig:gains-pacs}
\end{figure}

\begin{figure}[ht]
\centering
\includegraphics[width=\textwidth]{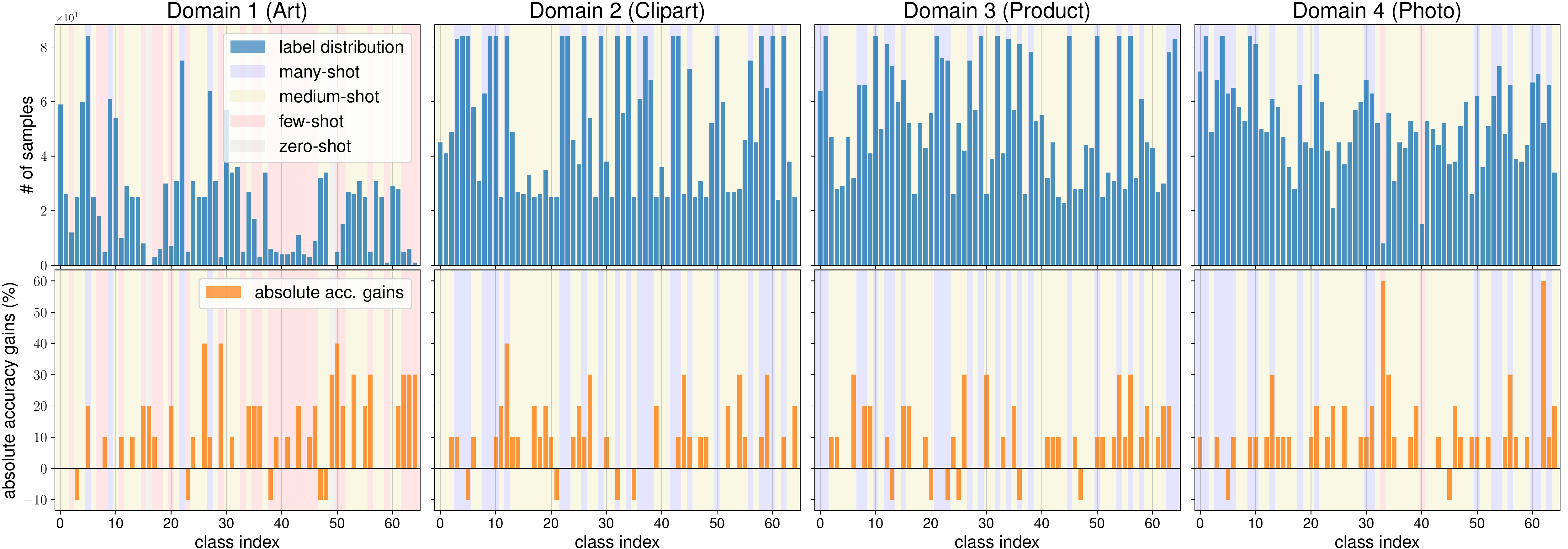}
\vspace{-0.5cm}
\caption{The absolute accuracy gains of \boda \emph{vs.} ERM over all domain-class pairs on \texttt{OfficeHome-MLT}.}
\label{appendix:fig:gains-officehome}
\end{figure}

\begin{figure}[ht]
\centering
\includegraphics[width=\textwidth]{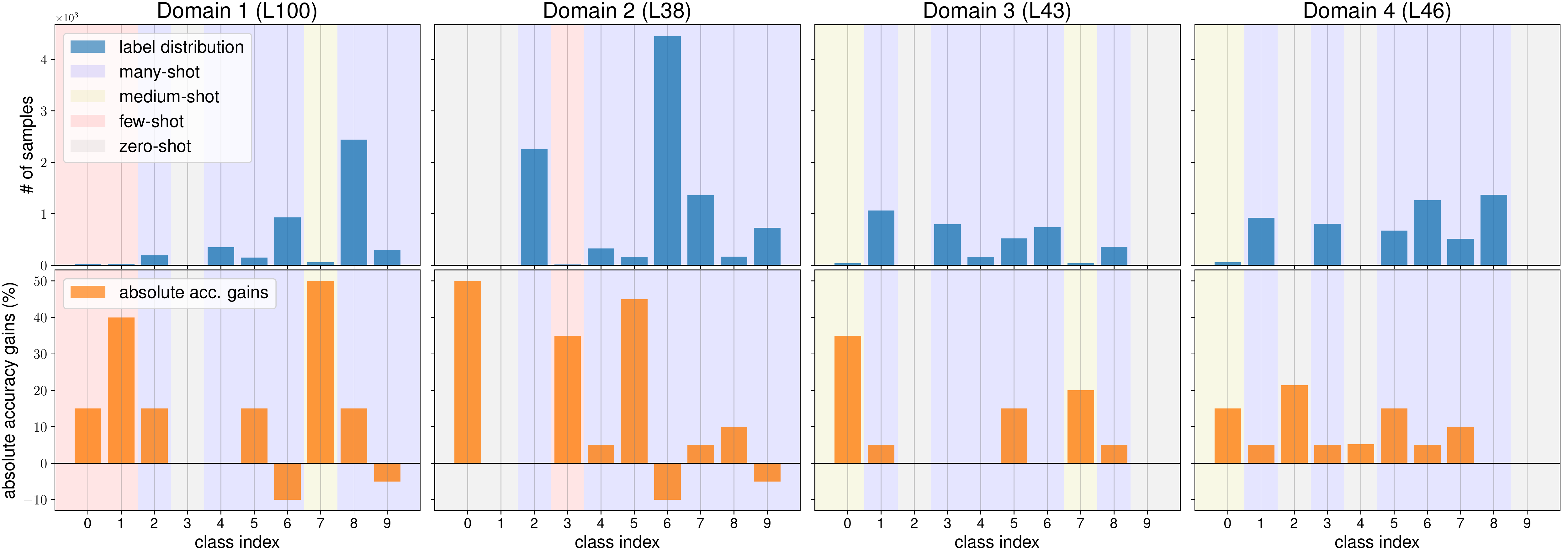}
\vspace{-0.5cm}
\caption{The absolute accuracy gains of \boda \emph{vs.} ERM over all domain-class pairs on \texttt{TerraInc-MLT}.}
\label{appendix:fig:gains-terraincognita}
\end{figure}

\begin{figure}[t]
\centering
\includegraphics[width=\textwidth]{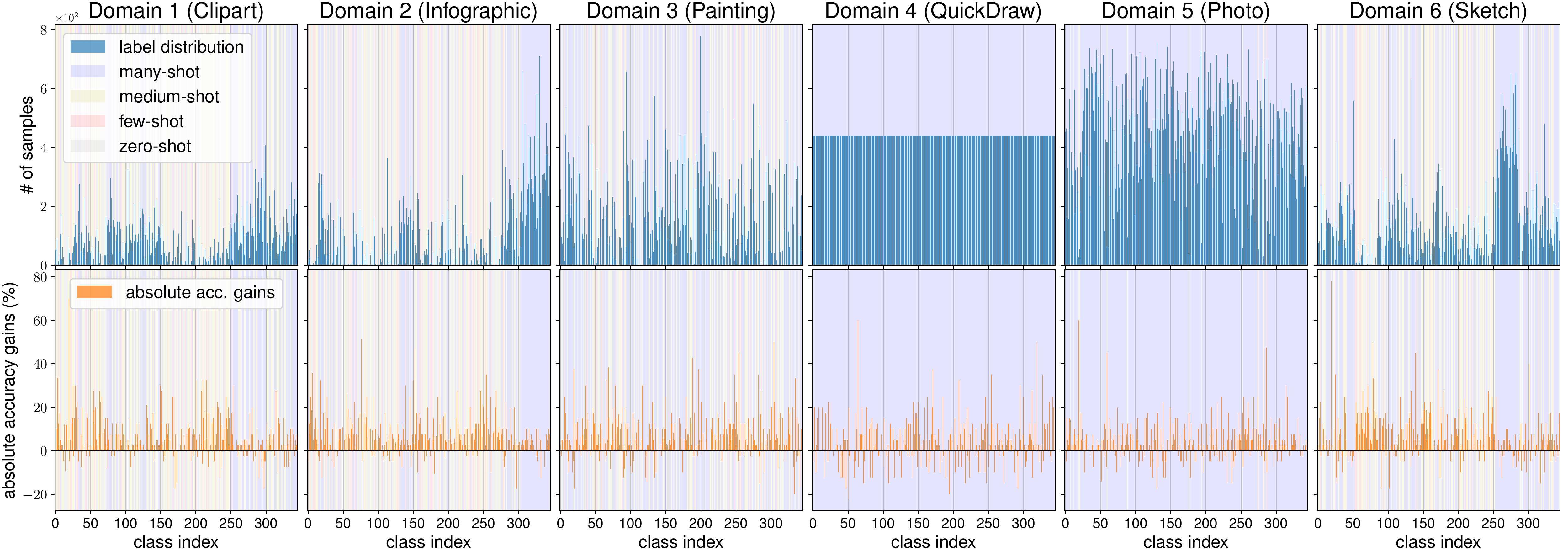}
\vspace{-0.5cm}
\caption{The absolute accuracy gains of \boda \emph{vs.} ERM over all domain-class pairs on \texttt{DomainNet-MLT}.}
\label{appendix:fig:gains-domainnet}
\end{figure}

We provide additional results for understanding how \boda performs across \emph{all} domain-class pair when cross-domain imbalance occurs.
Similar to Fig.~\ref{fig:gains-officehome} in the main text, we plot the absolute gains of \boda over ERM on all five MDLT datasets, shown in Figs. \ref{appendix:fig:gains-vlcs}, \ref{appendix:fig:gains-pacs}, \ref{appendix:fig:gains-officehome}, \ref{appendix:fig:gains-terraincognita}, and \ref{appendix:fig:gains-domainnet}.
Across all datasets, we observe that \boda establishes large improvements w.r.t. all regions, especially for the few-shot and zero-shot ones.

\newpage

\vspace{5cm}

\newpage

\vspace{5cm}

\newpage

\vspace{1cm}

\subsection{Robustness to Diverse Skewed Label Distributions}
\label{subsec-appendix:diverse-skewed-labels}

We investigate how \boda performs under arbitrary label imbalance across domains, especially when the cross-domain label distributions are both \emph{imbalance} and \emph{divergent}. We again employ the \texttt{Digits-MLT} dataset, and manually vary the label proportions for each domain.

\begin{figure}[H]
\centering
\includegraphics[width=0.8\textwidth]{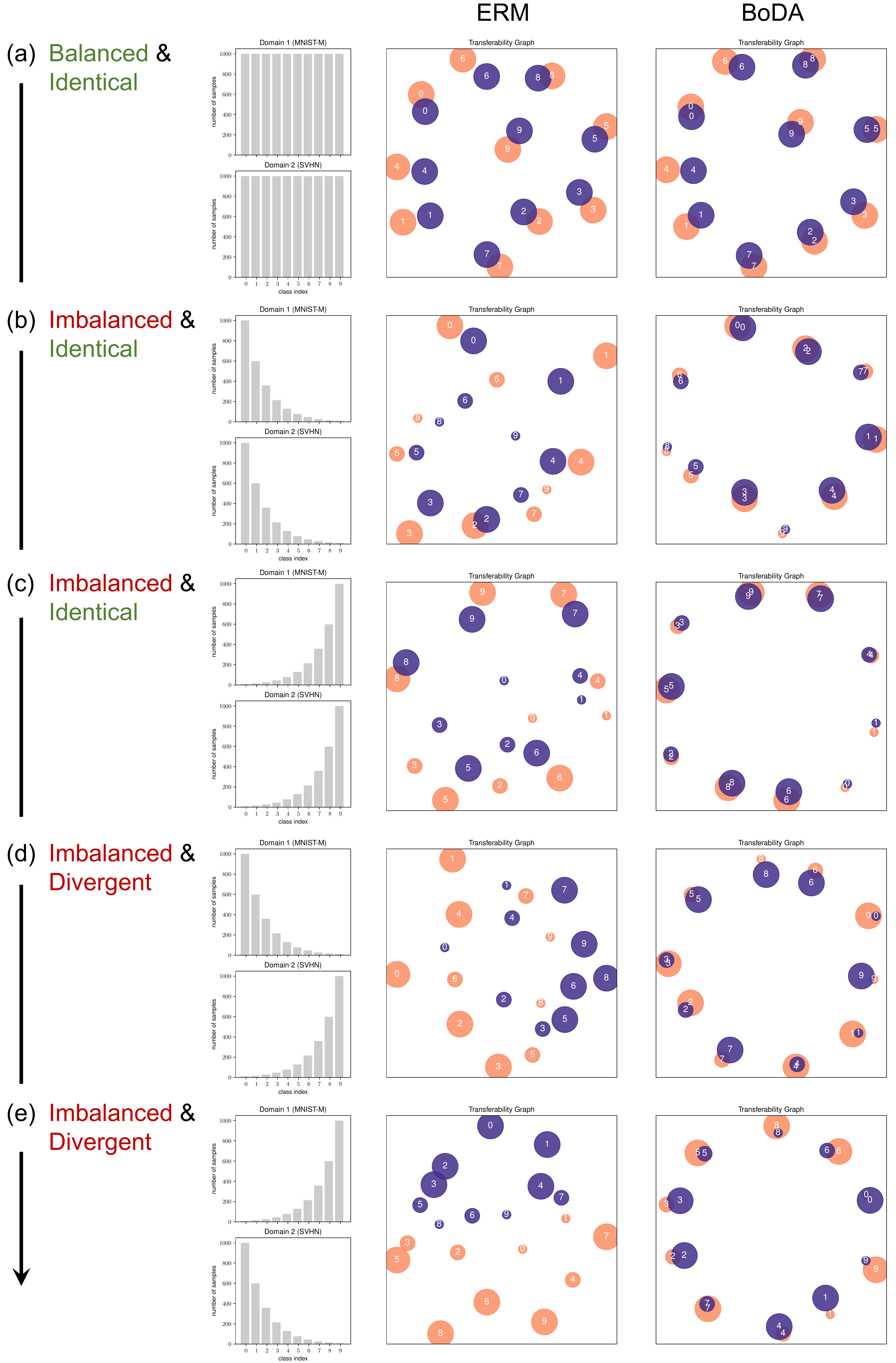}
\vspace{-0.4cm}
\caption{The evolving patterns of the transferability graph of \boda \emph{vs.} ERM across different label configurations on \texttt{Digits-MLT}. Label distributions for two domains are \textbf{(a)} balanced and identical; \textbf{(b)(c)} imbalanced and identical; \textbf{(d)(e)} imbalanced and divergent. \boda maintains consistent and transferable representations across all label configurations, and leads to much better test accuracy.}
\label{appendix:fig:compare-mds}
\end{figure}

As Fig.~\ref{appendix:fig:compare-mds} demonstrates, when the label distributions for two domains are balanced and identical, both ERM and \boda maintains discriminative representations.
If the label distributions become imbalanced but still identical across domains, ERM is still able to align similar classes in the two domains, but with majority classes being closer in terms of transferability than minority classes. In contrast, \boda maintains consistent transferability regardless of number of samples within each class.
Finally, as the label distributions become further mismatched across domains, ERM is not able to align the domains and produces a clear gap; by contrast, \boda maintains consistent and transferable representations even under severe data imbalance.
As a result, \boda substantially boosts the performance upon ERM, with an average gains of $6.4\%$ across all label configurations.

\subsection{Transferability \emph{vs.} Generalization on More Datasets}
\label{subsec-appendix:stats-corr-real-datasets}

\begin{figure}[!t]
\centering
\includegraphics[width=\textwidth]{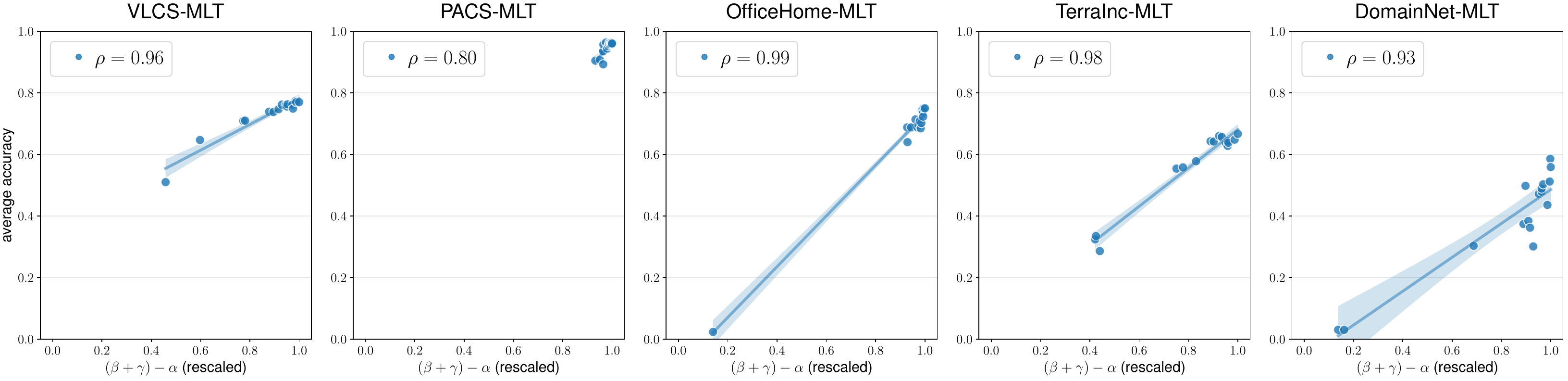}
\vspace{-0.4cm}
\caption{Correspondence between $(\beta+\gamma) - \alpha$ quantity and test accuracy across different MDLT datasets. Each point within each plot corresponds to a model trained with ERM using different hyperparameters.}
\label{appendix:fig:motivate-stats-real}
\end{figure}

We provide further results on transferability statistics \emph{vs.} generalization on real MDLT datasets, in addition to results on \texttt{Digits-MLT} as we showed in the main text.

Specifically, on all five MDLT datasets, we train 20 ERM models with varying hyperparameters, calculate the $(\alpha,\beta,\gamma)$ statistics for each model, and plot its classification accuracy against $(\beta+\gamma) - \alpha$.
Fig.~\ref{appendix:fig:motivate-stats-real} reveals similar and consistent findings, that the $(\alpha,\beta,\gamma)$ statistics characterize model performance in MDLT. Across all datasets, the $(\beta+\gamma) - \alpha$ quantity displays a very strong correlation with test performance across the entire range, suggesting that the $(\alpha,\beta,\gamma)$ statistics govern the success of learning in MDLT.

\subsection{Additional Visualization of Feature Discrepancy}
\label{subsec-appendix:feat-stats-discrepancy}

\begin{figure}[!t]
\centering
\includegraphics[width=0.85\textwidth]{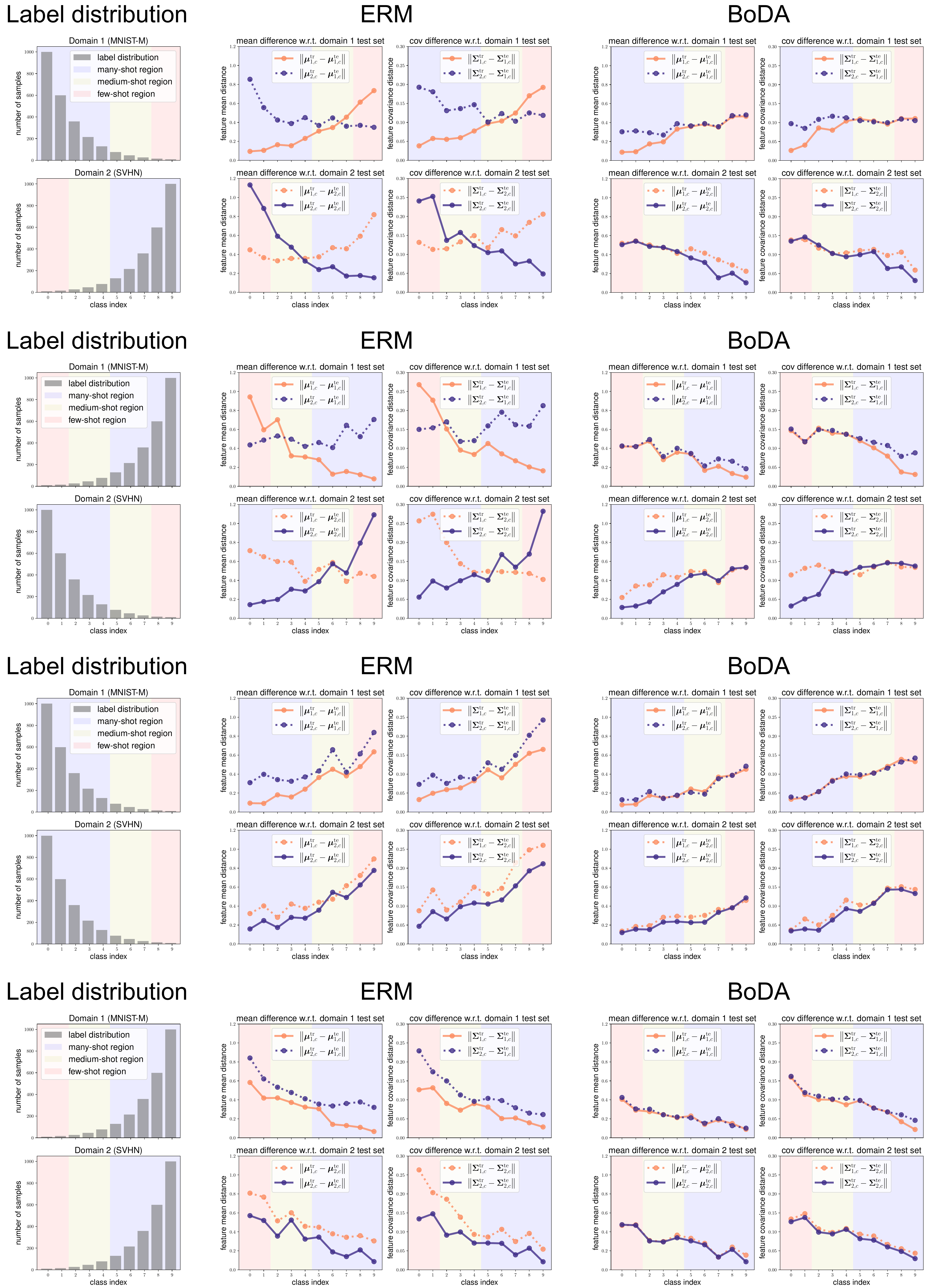}
\vspace{-0.3cm}
\caption{Feature discrepancy of \boda \emph{vs.} ERM across different label configurations on \texttt{Digits-MLT}. Each row plots a per-domain label distribution, and the feature mean / covariance distance between training and test data on each domain for both ERM and \boda. \boda enables better learned tail $(d,c)$ with smaller feature discrepancy.}
\label{appendix:fig:compare-feat-stats}
\end{figure}

We provide additional results for understanding \boda, i.e., how \boda calibrates the feature statistics. Fig.~\ref{appendix:fig:compare-feat-stats} shows the feature discrepancy of \boda \emph{vs.} ERM across different label configurations on \texttt{Digits-MLT}.
In addition to the mean distance we showed in the main text, we show also the feature covariance distance between training and test data, and plot them for both domains.
Similarly, solid lines plot the distance between training and test data from the same domain-class pairs. Dashed lines plot the distance between test data from a particular domain-class pair and the training data with which it shares the same class but differs in the domain.
The figure also shows regions with different data densities using colors \textcolor{manyshot}{blue}, \textcolor{medshot}{yellow}, \textcolor{fewshot}{red}.

As the figure confirms, across different label distributions, \boda consistently learns better representations especially for the tail data (i.e., the red regions), where the feature mean/covariance distance between training and test data becomes smaller and more aligned across domains.
Comparing \boda with ERM further demonstrates that \boda maintains consistent and transferable representations with smaller feature discrepancy.

